\documentclass{article}

\usepackage{arxiv}

\usepackage{amsmath,amssymb,amsfonts}
\usepackage{algorithmic}
\usepackage{graphicx}
\usepackage{textcomp}
\usepackage{hyperref}
\usepackage{pifont}
\usepackage{xcolor}
\usepackage{bm}
\usepackage{pdflscape}
\usepackage{multirow}
\usepackage{subfigure}
\usepackage{subcaption}
\usepackage{tabularx}

\newcommand{\cmark}{\textcolor{green!60!black}{\ding{51}}}  
\newcommand{\xmark}{\textcolor{red}{\ding{55}}}             

\title{AIFloodSense: A Global Aerial Imagery Dataset for Semantic Segmentation and Understanding of Flooded Environments}
\newif\ifuniqueAffiliation
\uniqueAffiliationtrue

\ifuniqueAffiliation 
\author{
Georgios~Simantiris\thanks{Corresponding author. \texttt{ddk230@edu.hmu.gr}} \\
Department of Management Science and Technology\\
Hellenic Mediterranean University\\
Agios Nikolaos, Crete 72100, Greece
\And
Konstantinos~Bacharidis \\
Department of Management Science and Technology\\
Hellenic Mediterranean University\\
Agios Nikolaos, Crete 72100, Greece\\
Institute of Computer Science, FORTH\\
Heraklion, Crete 70013, Greece
\And
Apostolos~Papanikolaou \\
Department of Management Science and Technology\\
Hellenic Mediterranean University\\
Agios Nikolaos, Crete 72100, Greece
\And
Petros~Giannakakis \\
Department of Management Science and Technology\\
Hellenic Mediterranean University\\
Agios Nikolaos, Crete 72100, Greece
\And
Costas~Panagiotakis \\
Department of Management Science and Technology\\
Hellenic Mediterranean University\\
Agios Nikolaos, Crete 72100, Greece\\
Institute of Computer Science, FORTH\\
Heraklion, Crete 70013, Greece
}
\else
\fi

\renewcommand{\shorttitle}{\textit{arXiv} Template}

\begin{document}
\maketitle

\begin{abstract}
Accurate flood detection from visual data is a critical step toward improving disaster response and risk assessment, yet datasets for flood segmentation remain scarce due to the challenges of collecting and annotating large-scale imagery. Existing resources are often limited in geographic scope and annotation detail, hindering the development of robust, generalized computer vision methods. To bridge this gap, we introduce AIFloodSense, a comprehensive, publicly available aerial imagery dataset comprising 470 high-resolution images from 230 distinct flood events across 64 countries and six continents. Unlike prior benchmarks, AIFloodSense ensures global diversity and temporal relevance (2022-2024), supporting three complementary tasks: (i) Image Classification with novel sub-tasks for environment type, camera angle, and continent recognition; (ii) Semantic Segmentation providing precise pixel-level masks for flood, sky, and buildings; and (iii) Visual Question Answering (VQA) to enable natural language reasoning for disaster assessment. We establish baseline benchmarks for all tasks using state-of-the-art architectures, demonstrating the dataset's complexity and its value in advancing domain-generalized AI tools for climate resilience.  Our dataset is available at \url{https://sites.google.com/site/costaspanagiotakis/research/aifloodsense}.

\end{abstract}





\section{Introduction}
\label{sec:intro}

The increasing frequency and severity of flood events worldwide, driven by climate change and urban expansion \cite{Ritchie2025}, highlight the urgent need for advanced tools to assess and mitigate their impact \cite{Kondratyev2006natural}. Aerial imagery, captured through Unmanned Aerial Vehicles (UAVs), satellites, or other remote sensing technologies, has proven invaluable for flood monitoring and management. These images offer detailed spatial insights, enabling accurate segmentation and analysis of flooded areas. However, the effectiveness of flood segmentation relies heavily on the availability of high-quality, annotated datasets specifically tailored to this purpose.

Despite the critical importance of such datasets, there remains a significant gap in the availability of comprehensive, publicly accessible datasets dedicated to flood segmentation. Existing datasets often suffer from limitations such as insufficient resolution, lack of diverse environmental conditions, or inadequate annotations. These challenges hinder the development and benchmarking of robust machine learning and deep learning algorithms capable of handling the complexities of real-world flood scenarios.

To address this gap, we present a new dataset of aerial images specifically designed for flood segmentation tasks. This dataset includes high-resolution imagery from diverse geographic locations covering all inhabited continents, encompassing various flood scenarios, terrains, and environmental conditions. Accompanied by precise pixel-level annotations for three classes, namely flood, sky, and building, as well as a multi-faceted classification scheme characterizing the environmental setting, sky visibility, and geographic provenance, it also supports visual question answering (VQA) tasks, such as determining the number of flooded buildings or identifying whether the flood occurs in an urban or rural environment. Additionally, with baselines provided for all tasks, this dataset aims to serve as a valuable resource for researchers and practitioners in the fields of remote sensing, computer vision, and disaster management.

This work introduces a novel dataset and a comprehensive analysis framework for flood segmentation, classification, and VQA, addressing critical gaps in current resources and methodologies. The key contributions of this paper are summarized as follows:
\begin{itemize}
    \item \textit{A New Dataset for Flood Segmentation and Beyond:} We present a recent, high-resolution dataset specifically designed for flood segmentation tasks. It includes pixel-level annotations for three primary classes: \textbf{flood}, \textbf{buildings}, and \textbf{sky}, enabling detailed segmentation of flooded regions while providing ancillary information on the built environment and atmospheric visibility.
    \item \textit{Environmental Classification Annotations:} The dataset further incorporates classification labels that describe the surrounding environment, distinguishing between \textbf{rural} and \textbf{urban/peri-urban} areas. Additionally, each image is annotated to indicate the \textbf{absence} or \textbf{presence of sky}, facilitating nuanced analysis of visual and atmospheric conditions, and with the geographic provenance (continent) while also providing latitude and longitude.
    \item \textit{Global and Recent Coverage:} The dataset comprises aerial images captured during flood events in 2022 and 2024, ensuring temporal relevance. It includes diverse geographic locations, spanning all inhabited continents, to provide a broad range of environmental, climatic, and cultural contexts for segmentation and classification.
    \item \textit{Baseline Segmentation and Classification Results:} To establish benchmarks, we implement and evaluate baseline methods for segmentation across the three annotated classes (flood, sky, building). Additionally, we provide baseline classification results for two key classification tasks differentiating between rural and urban/peri-urban environments, and identifying images with or without visible sky, while introducing a novel benchmark for predicting the continent of origin.
    \item \textit{Visual Question Answering (VQA) Task and Benchmarks:} As a novel extension, we introduce a visual question answering (VQA) task tailored to the dataset, enabling the development of models capable of answering flood-related questions from aerial imagery. 
    We provide baseline VQA results using state-of-the-art methods, setting benchmarks for future research to integrate flood segmentation, classification, and VQA tasks.
\end{itemize}

By introducing this dataset and its multi-faceted tasks, we aim to advance the state of research in flood monitoring, segmentation, classification, and visual question answering. This resource is expected to facilitate the development of robust, context-aware algorithms for disaster analysis and decision-making, enhancing the capabilities of artificial intelligence in flood response and mitigation efforts.

The remainder of this paper is organized as follows: Section~\ref{sec:relwork} provides an overview of the related research. Section~\ref{sec:aifsds} describes our proposed AIFloodSense dataset covering characteristics, annotation protocols, and tasks. Details on the implementation and evaluation configurations for the baseline methods are presented in Section~\ref{sec:impl}, while the experimental results are discussed in Section~\ref{sec:resdis}. Finally, future work and conclusions are provided in Sections~\ref{sec:future} and \ref{sec:conclusion}, respectively.

\section{Related Work}
\label{sec:relwork}

The development of comprehensive datasets has been instrumental in advancing research in computer vision and remote sensing, particularly for tasks such as image segmentation, classification, and visual question answering. Although extensive datasets exist for general segmentation and classification, there remains a critical gap in datasets specifically tailored to natural disasters, especially flood-related scenarios, that include pixel-level annotations for multiple classes across diverse environmental conditions.

In this section, we review existing datasets relevant to natural disaster analysis and discuss methodologies employed for various tasks, with a particular focus on flood assessment using aerial imagery.

\subsection{Datasets}
\label{subsec:ds}

A category of datasets utilized in disaster analysis comprises textual, speech-based, and social media-derived resources, as well as multimodal data integrations. The significance of non-imaging datasets for flood detection and damage assessment from internet sources was recognized early in \cite{Reuter2018}. Jamaludin et al. \cite{Jamaludin2024} leveraged crowdsourced data from social networking sites (SNS) to evaluate a single flood event, highlighting SNS as a crucial source of near real-time information that supports emergency response efforts. Similarly, Farah \cite{Farah2024} introduced M-CATNAT, a multimodal dataset of French-language tweets related to natural disasters, incorporating annotations for textual, visual, and combined modalities. This dataset not only expands the scope of disaster-related resources beyond English-language data, but also enhances multimodal disaster analysis capabilities.

Comparative studies on machine learning models applied to disaster-related tweets \cite{Boston2024} and flood-specific German-language tweets \cite{Blomeier2024} emphasize the potential of social media data for real-time disaster response and management. Soomro et al. \cite{Soomro2024} analyzed social media content related to the 2022 urban flood in Karachi, Pakistan, to assess public opinion and its role in post-disaster evaluation. Salley et al. \cite{Salley2024} conducted a case study on a flood event in Michigan, analyzing Twitter/X message content and sentiment to inform more effective crisis communication strategies and bridge the gap between official information and community-driven narratives.

Beyond textual and social media sources, several studies integrated multimodal data for enhanced flood perception. Giri et al. \cite{Giri2024}, Yan et al. \cite{Yan2024}, and Akhtar et al. \cite{Akhtar2024} employed fusion techniques that combined textual, visual, remote sensing, and geospatial data at the decision level, facilitating a more comprehensive understanding of flood events.

In addition to non-imaging datasets, imaging datasets derived from remote sensing, aerial photography, and satellite imagery have been increasingly leveraged for flood detection and analysis. These datasets offer critical visual information for tasks such as flood extent estimation, water body segmentation, and the assessment of infrastructure damage. Despite significant advancements in remote sensing technology, the availability of large-scale, flood-specific datasets with high-resolution, pixel-wise annotations remains scarce. Current datasets predominantly focus on general disaster detection rather than flood-specific segmentation, thereby limiting the development of specialized deep learning models for flood response applications. 

In the following paragraphs, we present a comprehensive review of imaging datasets employed in flood analysis, emphasizing their scope, annotation quality, and relevance to various machine learning tasks. Specifically, the AIDER \cite{AIDER2019} and Incidents1M \cite{Incidents1M2023} datasets classify disaster events, including floods, from aerial and ground-level imagery, with varying resolutions and diverse conditions of illumination and viewpoint. In particular, the latter includes approximately 1 million images and categorizes events into 43 distinct classes with multiple labels. 

The detection of damaged buildings across different damage levels is the focus of the xBD \cite{xBD2019} and ISBDA \cite{ISBDA2020}, \cite{ISBDA2021} datasets. ISBDA utilizes 1030 frames from 10 videos taken after hurricane events, while xBD comprises 22068 pre- and post-disaster optical and multiband satellite images from 8 different disasters, including hurricanes and floods, with annotations for approximately 700000 buildings. Pre- and post-satellite imagery are also used in Multi3Net \cite{Multi3Net2019} to segment flooded buildings following Hurricane Harvey in Texas, while SpaceNet8 \cite{SpaceNet82022} involves the segmentation of flooded roads in Louisiana and Dernau, Germany.

FloodNet \cite{FloodNet2021} and RescueNet \cite{RescueNet2023} stand out as two of the most extensive datasets for multi-class segmentation, featuring 2343 and 4494 aerial images respectively, with resolutions of 4000x3000 pixels, though they are geographically constrained to post-event analysis in Texas, Louisiana, and Florida. These datasets provide pixel-wise annotations for 10 and 11 classes, respectively, including water, roads, flooded and non-flooded buildings, with the latter also classifying building damage into three damage severity levels.

Satellite SAR data from Sentinel-1 and Sentinel-2 are frequently used in flood water and permanent water segmentation tasks. Ombrianet \cite{OmbriaNet2022} and WaterBodies \cite{WaterBodies2023} employ pre- and post-disaster imagery in 256x256 pixel tiles for this purpose, while the FloodExtend \cite{FloodExtend2022} dataset uses only post-disaster images. Sen1Floods11 \cite{Sen1Floods112020} consists of 4831 512x512 chips from 11 flood events across 6 continents, aimed at distinguishing between permanent and flood water. Sen12-FLOOD \cite{Sen12Flood2020} leverages pre-disaster knowledge to segment flood water from 412 time series. Finally, WorldFloods \cite{WorldFloods2021} and its extended version \cite{WorldFloodsv22023}, employ flood extent maps from Sentinel 2 satellite images to obtain 256x256 image patches including 119 and 180 flood events worldwide, respectively.

Aerial imagery for flood segmentation is finally also employed in BlessemFlood \cite{BlessemFlood212024}, which focuses on a flood event in Erftstadt-Blessem, Germany, and contains 4623 images with 512x512 resolution. Lastly, the FAD \cite{FAD2022} and FSSD \cite{FSSD2023} datasets from Kaggle offer flood segmentation with 290 (800x600) and 663 (512x512 with zero-padding) images, respectively. 

A comparison of these imaging datasets, along with their attributes and characteristics, is provided in Table~\ref{tab:dspresentation}. A notable gap is the lack of recent flood event datasets that cover the entire globe. Moreover, while some datasets support multiple tasks, those that do so often lack dedicated flood segmentation. As a final remark we should note that while several datasets tackle flood segmentation and related remote sensing tasks, they remain constrained in both scale and diversity, with limited coverage of scene topologies, viewpoints, and resolutions. These limitations restrict the robustness and transferability of models to new environments. To overcome this, recent efforts have proposed data augmentation through generative approaches \cite{simantiris2025closing}, as well as automatic or weakly supervised annotation pipelines \cite{simantiris2024unsupervised}, \cite{paul2021flood}, which alleviate the reliance on costly manual labeling. However, in our analysis, we do not consider such strategies, as our goal is to establish a fair and direct comparison between the proposed dataset and existing real-world datasets without introducing confounding factors from synthetic data or pseudo-annotations. By doing so, we focus on assessing the intrinsic value of our dataset in consistent and realistic evaluation settings.

\subsection{Methodologies}
\label{subsec:meth}

\subsubsection{Classification}
\label{ssubsec:cl}

In recent years, image-based post-disaster damage detection has been dominated by deep learning-based methods (\cite{Bejiga2017}, \cite{Kyrkou2020},\cite{Nguyen2017}, \cite{Sharma2017}, \cite{Zhao2018}). Early approaches largely adapted traditional computer vision backbones for disaster-specific tasks. For instance, Bejiga et al. \cite{Bejiga2017} utilized a Support Vector Machine on top of a Convolutional Neural Network (CNN) followed by a Hidden Markov Model post-processing to detect avalanches. Similarly, Sharma et al. \cite{Sharma2017} evaluated for fire detection top-performing CNNs, namely VGGNet \cite{Simonyan2015VGGNet} and ResNet \cite{he2016deep}, while Kyrkou and Theocharides \cite{Kyrkou2020} extensively applied CNNs for emergency response scenarios involving fire, flood, collapsed buildings, and crashed cars. Addressing the need for aerial analysis, Zhao et al. \cite{Zhao2018} developed a saliency detection algorithm focused on fast location and segmentation of the core fire area with a 15-layered self-learning DCNN architecture. 

Furthermore, Da et al. \cite{Da2022SDAFormer} in their SDAFormer framework employ a symmetric hierarchical Transformer within a Siamese U-Net architecture to assess building damage from pre- and post-disaster satellite imagery, achieving state-of-the-art performance on the xBD dataset. WaterDetectionNet \cite{Huang2024WaterDetectionNet} explored a hybrid CNN-Transformer architecture enriched with attention modules to accurately segment floodwaters in complex SAR imagery. Most recently, the field has moved toward Foundation Models and Generalizable Transformers to address label scarcity in new disaster events. For instance, DAVI \cite{Ahn2025Davi} leverages the Segment Anything Model (SAM) \cite{kirillov2023sam} to generate pseudo-labels for unsupervised domain adaptation, enabling accurate damage detection in unseen regions.

\begin{landscape}
\begin{table}[]
\centering
\caption{Overview of existing flood-related imaging datasets, sorted by the year of publication. "Imagery" specifies the acquisition platform, where satellite entries cover optical, SAR, and multispectral data. Ration in the "Flood imgs / total imgs" column account for datasets containing also non flood-related imagery. Parenthetical values in the Semantic Segmentation and Classification columns indicate the total number of classes. A dash -- is used to indicate that data is missing or a category doesn't apply.}
\label{tab:dspresentation}

\setlength{\tabcolsep}{3pt} 

\resizebox{\linewidth}{!}{%
\begin{tabular}{|l|c|c|c|c|c|c|c|c|c|c|c|c|}
\hline
\multicolumn{1}{|c|}{\textbf{Dataset}} &
  \textbf{Publ.} &
  \begin{tabular}[c]{@{}c@{}}\textbf{Flood}\\\textbf{dates}\end{tabular} &
  \begin{tabular}[c]{@{}c@{}}\textbf{Conti-}\\\textbf{nents}\end{tabular} &
  \textbf{Countries} &
  \begin{tabular}[c]{@{}c@{}}\textbf{Flood}\\ \textbf{events}\end{tabular} &
  \begin{tabular}[c]{@{}c@{}}\textbf{Pre/post}\\ \textbf{disaster}\end{tabular} &
  \textbf{Imagery} &
  \begin{tabular}[c]{@{}c@{}}\textbf{Flood imgs /}\\ \textbf{total imgs}\end{tabular} &
  \textbf{Resolution} &
  \begin{tabular}[c]{@{}c@{}}\textbf{Semantic}\\\textbf{segmen-}\\\textbf{tation}\\\textbf{(classes)}\end{tabular} &
  \begin{tabular}[c]{@{}c@{}}\textbf{Classi-}\\ \textbf{fication}\\\textbf{(classes)}\end{tabular} &
  \textbf{VQA} \\ \hline\hline
  
Multi3Net \cite{Multi3Net2019} &
  \textbf{2019} &
  2017 &
  1 &
  1 &
  1 &
  both &
  Satellite &
  -- / -- &
  -- &
  \cmark~(2)&
  \xmark~(--)&
  \xmark \\ \hline
  
AIDER \cite{AIDER2019} &
  \textbf{2019} &
  $\leq$ 2019 &
  -- &
  -- &
  -- &
  post &
  \textbf{Aerial} &
  \begin{tabular}[c]{@{}c@{}}370 / 2545\end{tabular} &
  varies &
  \xmark~(--)&
  \cmark~(5)&
  \xmark \\ \hline
  
xBD \cite{xBD2019} &
  \textbf{2019} &
  2016--2019 &
  4 &
  8 &
  8 &
  both &
  Satellite &
  -- / 22068 &
  1024x1024 &
  \xmark~(--)&
  \cmark~(4)&
  \xmark \\ \hline
  
Sen1Floods11 \cite{Sen1Floods112020} &
  \textbf{2020} &
  2016--2019 &
  \textbf{6} &
  -- &
  11 &
  post &
  Satellite &
  4831 / 4831 &
  512x512 &
  \cmark~(3)&
  \xmark~(--)&
  \xmark \\ \hline
  
ISBDA \cite{ISBDA2020} &
  \textbf{2020} &
  2017--2019 &
  1 &
  1 &
  6 &
  post &
  \textbf{Aerial} &
  1030 / 1030 &
  varies &
  \xmark~(--)&
  \cmark~(3) &
  \xmark \\ \hline
  
Sen12-FLOOD \cite{Sen12Flood2020} &
  \textbf{2020} &
  2018--2019 &
  3 &
  10 &
  -- &
  both &
  Satellite &
  $\sim$4194 / $\sim$9476 &
  512x512 &
  \xmark~(--)&
  \cmark~(2) &
  \xmark \\ \hline
  
FloodNet \cite{FloodNet2021} &
  \textbf{2021} &
  2017 &
  1 &
  1 &
  1 &
  post &
  \textbf{Aerial} &
  -- / 2343 &
  \textbf{4000x3000} &
  \textbf{\cmark}~(10) &
  \textbf{\cmark}~(2) &
  \textbf{\cmark} \\ \hline
  
WorldFloods \cite{WorldFloods2021} &
  \textbf{2021} &
  2015-2019 &
  \textbf{6} &
  -- &
  119 &
  post &
  Satellite &
  -- / 185574 &
  256x256 &
  \cmark~(5) &
  \xmark~(--) &
  \xmark \\ \hline
  
Spacenet-8 \cite{SpaceNet82022} &
  \textbf{2022} &
  2021 &
  2 &
  2 &
  2 &
  both &
  Satellite &
  -- / 1369 &
  varies &
  \cmark~(4) &
  \xmark~(--) &
  \xmark \\ \hline
  
\begin{tabular}[l]{@{}c@{}}Flood Extent\\Detection~\cite{FloodExtend2022}\end{tabular} &
  \textbf{2022} &
  $\leq$ 2021 &
  3 &
  3 &
  5 &
  post &
  Satellite &
  -- / $\sim$30000 &
  256x256 &
  \cmark~(3) &
  \xmark~(--)&
  \xmark \\ \hline
  
FAD \cite{FAD2022} &
  \textbf{2022} &
  $\leq$ 2022 &
  -- &
  -- &
  -- &
  post &
  \textbf{Aerial} &
  290 / 290 &
  800x600 &
  \cmark~(2) &
  \xmark~(--) &
  \xmark \\ \hline
  
Ombria \cite{OmbriaNet2022} &
  \textbf{2022} &
  2017--2021 &
  5 &
  17 &
  23 &
  both &
  Satellite &
  -- / 3376 &
  256x256 &
  \cmark~(2) &
  \xmark~(--) &
  \xmark \\ \hline
  
RescueNet \cite{RescueNet2023} &
  \textbf{2023} &
  2018 &
  1 &
  1 &
  1 &
  post &
  \textbf{Aerial} &
  -- / 4494 &
  \textbf{3000x4000} &
  \cmark~(11) &
  \cmark~(3) &
  \xmark \\ \hline
  
Incidents1M \cite{Incidents1M2023} &
  \textbf{2023} &
  $\leq$ 2020 &
  \textbf{6} &
  -- &
  -- &
  post &
  \begin{tabular}[c]{@{}c@{}}Aerial \&\\ Ground\end{tabular} &
  710 / 977088 &
  varies &
  \xmark~(--) &
  \cmark~(43) &
  \xmark \\ \hline
  
FSSD \cite{FSSD2023} &
  \textbf{2023} &
  $\leq$ 2023 &
  -- &
  -- &
  -- &
  post &
  \textbf{Aerial} &
  663 / 663 &
  512x512 &
  \cmark~(2) &
  \xmark~(--) &
  \xmark \\ \hline
  
WaterBodies \cite{WaterBodies2023} &
  \textbf{2023} &
  2000 - 2021 &
  \textbf{2} &
  3 &
  4 &
  both &
  \begin{tabular}[c]{@{}c@{}}Satellite\\ \& Aerial\end{tabular} &
  15000 / 105000 &
  256x256 &
  \cmark~(3) &
  \cmark~(2) &
  \xmark \\ \hline
  
WorldFloods v2 \cite{WorldFloodsv22023} &
  \textbf{2023} &
  2015-2023 &
  \textbf{6} &
  -- &
  144 & 
  both &
  Satellite &
  -- / 73809 &
  256x256 &
  \cmark~(3) &
  \xmark~(--) &
  \xmark \\ \hline
  
BlessemFlood21 \cite{BlessemFlood212024} &
  \textbf{2024} &
  2021 &
  1 &
  1 &
  1 &
  post &
  \textbf{Aerial} &
  -- / 4623 &
  512x512 &
  \cmark~(2) &
  \xmark~(--) &
  \xmark \\ \hline
  \hline
  
\textbf{AIFloodSense} (ours) &
   &
  \textbf{2022--2024} &
  \textbf{6} &
  \textbf{64} &
  \textbf{230} &
  post &
  \textbf{Aerial} &
  470 / 470&
  1024x768 &
  \textbf{\cmark}~(4) &
  \textbf{\cmark}~(10) &
  \textbf{\cmark} \\ \hline
\end{tabular}
} 
\end{table}
\end{landscape}

To establish robust baselines for binary flood classification (flood vs. non-flood), recent literature has advanced from simple CNNs to rigorous benchmarking and hybrid frameworks. Jackson et al. \cite{jackson2023flood} addressed label ambiguity in semi-supervised data via uncertainty offset analysis, while Yasi et al. \cite{yasi2024flood} enhanced generalization through deep ensemble learning. Innovation in feature optimization is exemplified by Dubey and Katarya \cite{dubey2025flood}, who integrated Swin Transformers with bio-inspired meta-heuristics. Extending this focus on reliability, Manaf et al. \cite{manaf2025aerial} recently proposed a framework combining robust deep learning with Explainable AI (XAI), ensuring that high-performance automated assessments remain interpretable and trustworthy in critical post-flood scenarios.

Beyond flood-specific tasks, the scope of aerial analysis has expanded to broader environmental monitoring \cite{zhu2025advancements} and Land Use/Land Cover classification \cite{banoth2024soil}, \cite{fayaz2024land}. To address the complexity of high-resolution imagery, recent literature has shifted from standard CNNs toward advanced spectral-spatial architectures, including hierarchical Vision Transformers \cite{khan2024transformer}, \cite{pradhan2024swinsight}, and novel State Space Models (Mamba) \cite{gao2025msfmamba}, which offer superior multi-scale feature fusion. Finally, addressing sensor orientation challenges, Hernández-López et al. \cite{hernandez2024sky} validated transfer learning for sky image classification, underscoring the importance of filtering non-nadir views to improve the robustness of automated monitoring pipelines.

\subsubsection{Semantic Segmentation}
\label{ssubsec:ss}

Deep Learning (DL) has progressively replaced traditional numerical and multicriteria analysis for flood mapping, enabling a more accurate generation of susceptibility and hazard maps \cite{bentivoglio2022deep}. Comprehensive reviews indicate that although DL significantly enhances flood prediction and management, challenges remain in data availability and generalization \cite{kumar2023state}.

In the domain of satellite imagery, Convolutional Neural Networks (CNNs) have demonstrated robust capabilities in delineating flood extents. Methods utilizing temporal differences in Synthetic Aperture Radar (SAR) and multispectral data effectively distinguish permanent water from inundated areas \cite{dong2023mapping}, \cite{OmbriaNet2022}. To address inherent uncertainties in SAR-based segmentation, Bayesian CNNs have been proposed to estimate the posterior distribution of model parameters, quantifying the prediction variance \cite{hertel2023probabilistic}. Specialized architectures, such as the multiscale Attentive Decoder Network (ADNet), have outperformed threshold-based baselines on Sentinel-1 data by leveraging co-polarization and cross-polarization inputs \cite{chouhan2023attentive}. Furthermore, U-Net variants that incorporate geomorphic features and MobileNet-V3 backbones have proven to be effective for three-class classification (flood, permanent water, background) in Sentinel-1 imagery \cite{li2023u}, \cite{WaterBodies2023}. Transformer-based architectures have also emerged as powerful alternatives, such as, for example, the Swin Transformer with a densely connected feature aggregation module (DCFAM) improves context capture \cite{wang2022novel}, while Bitemporal Image Transformers (BiT) excel in change detection tasks \cite{dong2023mapping}. Large-scale assessments using constellations like Sentinel-1/2 confirm that multi-satellite integration significantly expands coverage, though accuracy varies with catchment size \cite{tarpanelli2022effectiveness}.

For aerial and UAV imagery, research is focused on high-resolution segmentation to support rapid emergency response. The FloodNet dataset and the benchmark have validated models such as XceptionNet and ENet to distinguish floodwaters and inundated infrastructure \cite{FloodNet2021}. To optimize for real-time onboard processing, lightweight architectures such as enhanced ENet with atrous separable convolutions \cite{inthizami2022flood} and FASegNet \cite{csener2024novel} have been developed to maximize receptive fields while minimizing computational load. Similarly, the SpaceNet 8 challenge highlighted that simple U-Net architectures, when augmented with pre-training and robust data strategies, offer an optimal balance of accuracy and efficiency for detecting flooded buildings and roads \cite{SpaceNet82022}, \cite{hansch2023spacenet}. Novel approaches also include interactive segmentation models combining Mamba and convolution operations with prompt encoders to reduce labeling costs \cite{shi2024interactive}, as well as edge-computing pipelines facilitating real-time inference directly on UAV platforms \cite{hernandez2022flood}.

Despite the dominance of supervised DL, unsupervised, weakly supervised, and traditional methods remain vital, particularly where annotated data is scarce. Traditional object-based clustering (e.g., K-means, Region Growing) on SAR and optical data has shown resilience in complex environments \cite{ibrahim2021application}, \cite{landuyt2020flood}, while contextual filtering approaches provide automated mapping for non-urban extents \cite{mccormack2022methodology}. Recent innovations include the TFCSD weakly supervised framework, which decouples positive/negative sample acquisition to expedite mapping \cite{he2024efficient}, and unsupervised graph-based methods utilizing Markov Random Fields (MRF) for energy minimization \cite{trombini2023goal}. In the context of UAVs, the UFS-HT-REM method introduced a parameter-free, fully unsupervised pipeline using color and edge information to generate pseudo-labels \cite{simantiris2024ufs}, a strategy we recently extended to train deep learning models without manual ground truth \cite{simantiris2024unsupervised}. Similarly, automated datacube-based algorithms \cite{bauer2022satellite} and semi-supervised teacher-student models \cite{savitha2025} continue to advance operational capabilities by reducing reliance on extensive manual annotation.

\subsubsection{Visual Question Answering (VQA)}
\label{ssubsec:vqa}
  
The landscape of Remote Sensing VQA (RS-VQA) datasets and methods has expanded considerably, beginning with geographically and disaster event-constrained benchmarks such as RSVQA~\cite{RSVQA} and EarthVQA~\cite{EarthVQA}, and more recently with DisasterM3~\cite{DisasterM3}, which captures significantly richer geographic and semantic diversity across multiple hazard types. Early RS-VQA methods relied primarily on global visual representations. To better align visual and textual modalities, Zheng et al.~\cite{ZhengVQA} introduced a mutual-attention mechanism in which image features guide question-word attention and vice versa, with bilinear pooling used for fine-grained fusion. Felix et al.~\cite{FelixVQA} advanced this direction through an object-centric transformer that extracts local disaster-relevant features and combines them with BERT-encoded questions via a cross-modal transformer, leading to improved counting performance. Tosato et al.~\cite{tosato2024} incorporated segmentation-guided attention, injecting segmentation masks into the reasoning process to prioritize spatially relevant regions. More recently, the field has shifted toward Large Language Models (LLMs) and Multimodal LLMs (MLLMs) to support open-ended and grounded reasoning. GeoChat~\cite{Kuckreja2024} adapts LLaVA for remote sensing dialog, enabling region-aware grounded interaction, while RSGPT~\cite{RSGPT} provides a vision language model fine-tuned with high-quality remote sensing captions to improve detailed scene understanding. 

In the domain of flood disaster response, FloodNet \cite{FloodNet2021} pioneered the use of high-resolution UAV imagery for post-flood VQA, introducing questions related to condition recognition (e.g., "Is the road flooded?"), counting, and scene-level assessment. A subsequent work, SAM-VQA \cite{SAMVQA}, extended this by introducing a supervised attention framework that utilizes ground-truth segmentation masks to guide the model's focus, while also expanding the question set to include risk assessment and density estimation. More recently, ThiFAN-VQA~\cite{Ahn2025Davi} introduces a vision-language framework that follows a Chain-of-Thought prompting to reduce hallucinations and enhance reasoning accuracy in post-flood damage assessment. Our proposed dataset follows the VQA structure introduced by FloodNet but overcomes its geographic limitations by extending from a single-event setting to a multi-continent benchmark. In addition, we refine the evaluation protocols for the counting of questions by incorporating continuous error metrics, allowing a more faithful measurement of the severity of the predicted damage.  Finally, following FloodNet's evaluation scheme, in our baseline evaluation we do not incorporate disaster-specialized models. Instead, we explore general-purpose models chosen for their broad adoption, well-established architectures, and reproducibility.

\section{AIFloodSense Dataset}
\label{sec:aifsds}

\subsection{Dataset Characteristics}
\label{subsec:dschar}

\begin{figure*}[t!]
    \centering
    \includegraphics[width=\textwidth]{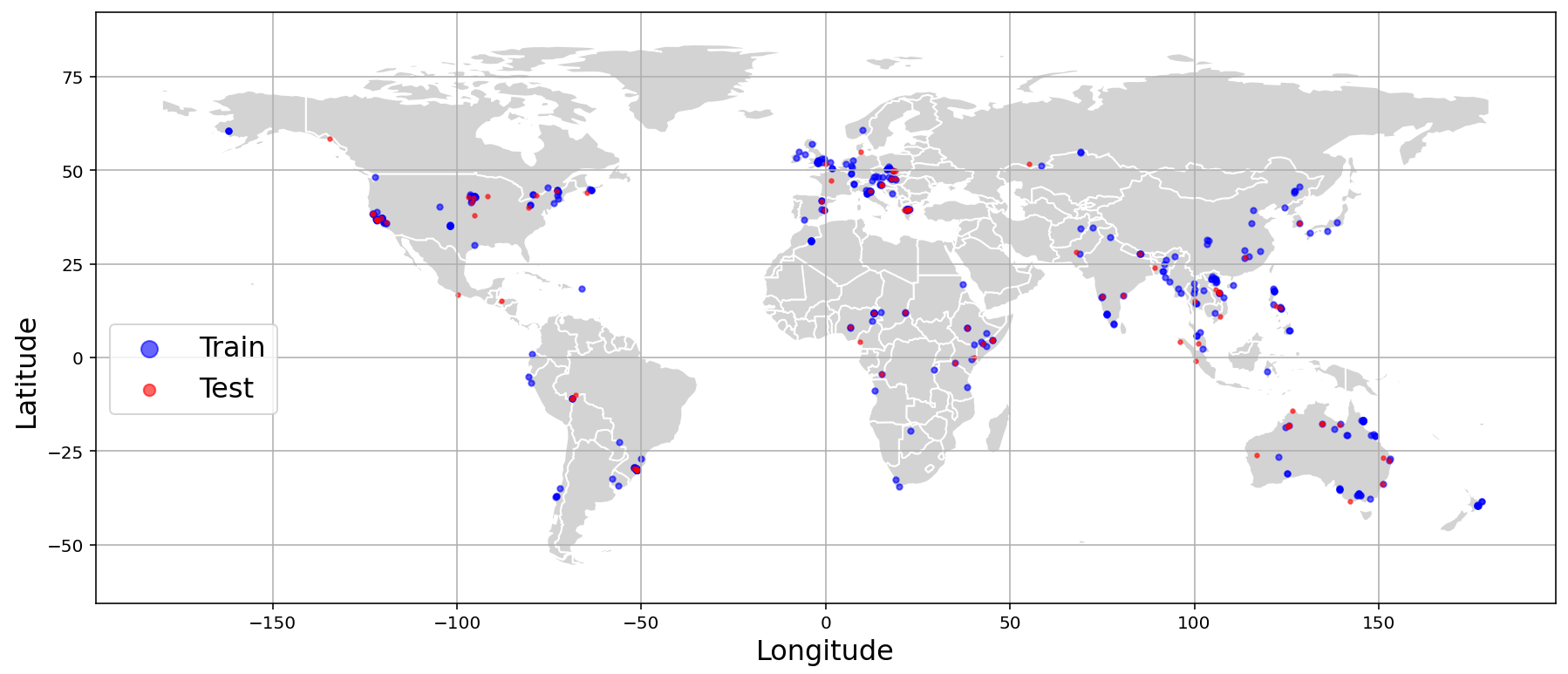}
    \caption{Geographic distribution of flood events in the proposed dataset. Training samples are indicated in blue and test samples in red, plotted on a global map to illustrate spatial coverage.}
    \label{fig:floodeventsworld}
\end{figure*}

\begin{figure*}[t!]
    \centering
    \includegraphics[width=\textwidth]{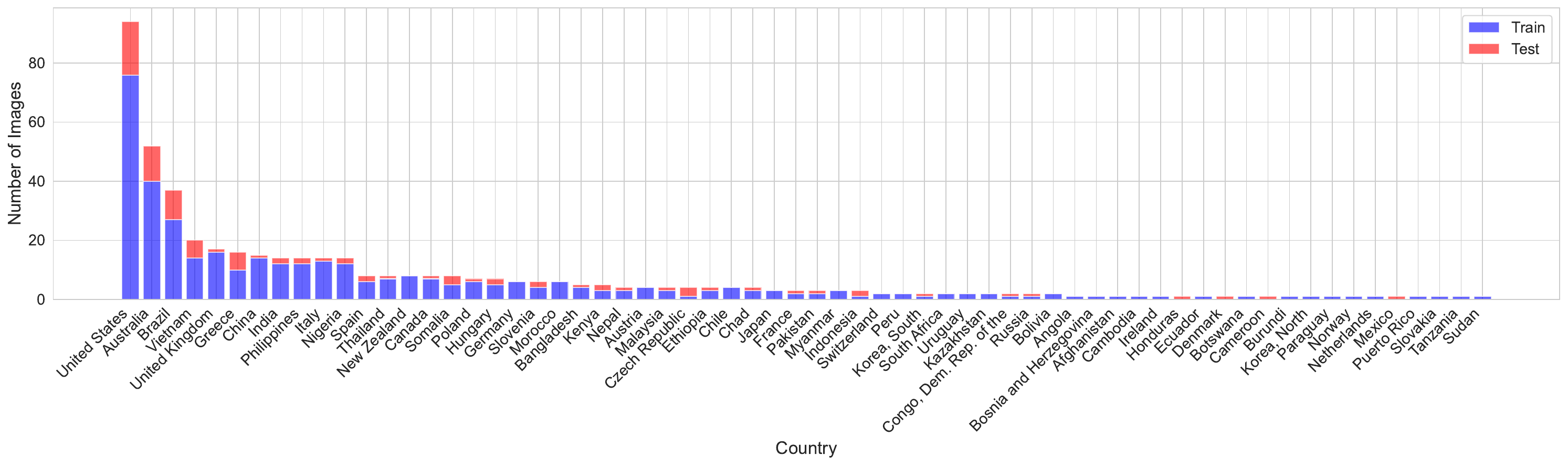}\\
    \caption{Distribution of flood event images across 64 countries in the proposed dataset, sorted in descending order by total image count. Training samples are shown in blue and test samples in red, highlighting the relative contribution of each country to the dataset.}
    \label{fig:countrydistr}
\end{figure*}

AIFloodSense is a newly curated dataset designed to advance semantic segmentation and analysis of flood events through high-resolution aerial imagery. The dataset comprises high-fidelity images sourced from the World Wide Web, capturing flood scenarios predominantly from \textbf{2023} and \textbf{2024}, with a select small subset from 2022 to ensure temporal relevance to contemporary flood disasters. The collection spans \textbf{64} countries on the six inhabited continents, Africa, Asia, Europe, North America, Oceania, and South America, as illustrated in Figures~\ref{fig:floodeventsworld} and~\ref{fig:countrydistr}. Antarctica was excluded due to its specific geomorphology, lack of permanent habitation, and minimal infrastructure, rendering flood impact assessment less relevant to the dataset's focus on human and environmental impact.

A rigorous selection protocol was employed to ensure data quality and environmental diversity. The dataset focuses exclusively on aerial imagery, primarily acquired via Unmanned Aerial Vehicles (UAVs), to provide bird's-eye perspectives suitable for large-scale flood assessment. To maintain high fidelity for computational analysis, blurred or low-quality samples were systematically excluded, resulting in a final corpus of 470 images with an average resolution of approximately 1.9 megapixels (1,908,570 pixels). These images document \textbf{230} distinct flood events, ensuring that the dataset captures a broad spectrum of contemporary flood disasters, rather than isolated events.

\textit{AIFloodSense} is characterized by significant environmental heterogeneity, encompassing both urban/peri-urban and rural landscapes. This diversity facilitates the evaluation of models across varying infrastructural densities. Furthermore, the dataset includes variations in sensor orientation, indicated by the presence or absence of sky regions, which introduces challenging variations in illumination and color contrast. Dataset statistics are summarized in Table~\ref{tab:dsstatsnr}. The comprehensive geographical distribution, combined with high-quality pixel-level annotations for four semantic classes,  Flood, Sky, Building, Background, establishes \textit{AIFloodSense} as a robust benchmark for developing and evaluating computer vision models for flood detection, segmentation, and damage assessment. This dataset holds promise for improving early warning systems, disaster response planning, and resilience strategies in flood-prone areas worldwide.

\begin{table*}[t]
\centering
\caption{Summary of the dataset distribution. Rows correspond to each continent, and columns report the total number of images, the number of images collected per year (2022-2024), the number of images per environment type (Urban/Peri-urban, Rural), and the number of images per camera angle (Sky presence, Sky absence).}
\label{tab:dsstatsnr}
\setlength{\tabcolsep}{3pt}

\resizebox{\linewidth}{!}{%
\begin{tabular}{|l||c||ccc||cc||cc|}
\hline
\multirow{2}{*}{\textbf{Continent}} & 
\multirow{2}{*}{\textbf{Imgs}} & 
\multicolumn{3}{c||}{\textbf{Imgs per Year}} & 
\multicolumn{2}{c||}{\textbf{Imgs per Environment}} & 
\multicolumn{2}{c|}{\textbf{Imgs per Camera Angle}} \\ 
\cline{3-9} 
 & 
 &
\multicolumn{1}{c|}{\textbf{2022}} & 
\multicolumn{1}{c|}{\textbf{2023}} & 
\textbf{2024} & \multicolumn{1}{c|} {\textbf{Urban/Peri-urban}} & 
\textbf{Rural} & 
\multicolumn{1}{c|}{\textbf{Sky presence}} & 
\textbf{Sky absence} \\ 
\hline
Africa                     & 51                                         & \multicolumn{1}{c|}{1}             & \multicolumn{1}{c|}{12}            & 38            & \multicolumn{1}{c|}{27}                        & 24             & \multicolumn{1}{c|}{25}                & 26              \\ \hline
Asia                       & 105                                        & \multicolumn{1}{c|}{3}             & \multicolumn{1}{c|}{30}            & 72            & \multicolumn{1}{c|}{80}                        & 25             & \multicolumn{1}{c|}{48}                & 57              \\ \hline
Europe                     & 100                                        & \multicolumn{1}{c|}{1}             & \multicolumn{1}{c|}{47}            & 52            & \multicolumn{1}{c|}{72}                        & 28             & \multicolumn{1}{c|}{39}                & 61              \\ \hline
North America              & 104                                        & \multicolumn{1}{c|}{0}             & \multicolumn{1}{c|}{62}            & 42            & \multicolumn{1}{c|}{62}                        & 42             & \multicolumn{1}{c|}{50}                & 54              \\ \hline
Oceania                    & 60                                         & \multicolumn{1}{c|}{0}             & \multicolumn{1}{c|}{44}            & 16            & \multicolumn{1}{c|}{28}                        & 32             & \multicolumn{1}{c|}{29}                & 31              \\ \hline
South America              & 50                                         & \multicolumn{1}{c|}{0}             & \multicolumn{1}{c|}{14}            & 36            & \multicolumn{1}{c|}{39}                        & 11             & \multicolumn{1}{c|}{14}                & 36              \\ \hline
\textbf{Total Imgs}                 & \textbf{470}                                        & \multicolumn{1}{c|}{\textbf{5}}    & \multicolumn{1}{c|}{\textbf{209}}  & \textbf{256}  & \multicolumn{1}{c|}{\textbf{308}}              & \textbf{162}   & \multicolumn{1}{c|}{\textbf{205}}      & \textbf{265}    \\ \hline
\end{tabular}
}
\end{table*}

\begin{figure}[t!]
    \centering
    \subfigure[Sample Distribution per Environment and Camera Angle.]{\includegraphics[width=0.49\textwidth]{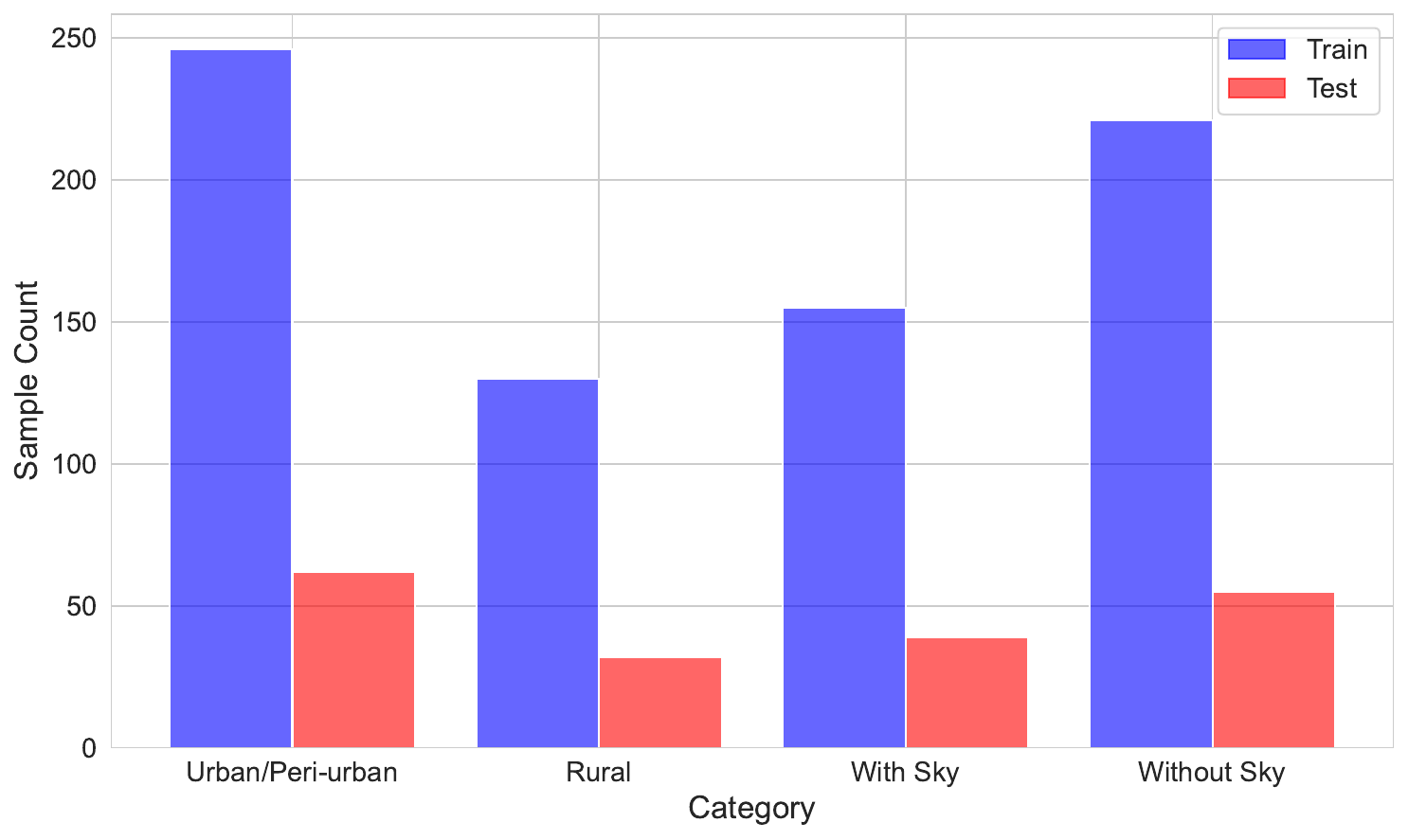}}
    \subfigure[Sample Distribution per Continent.]{\includegraphics[width=0.49\textwidth]{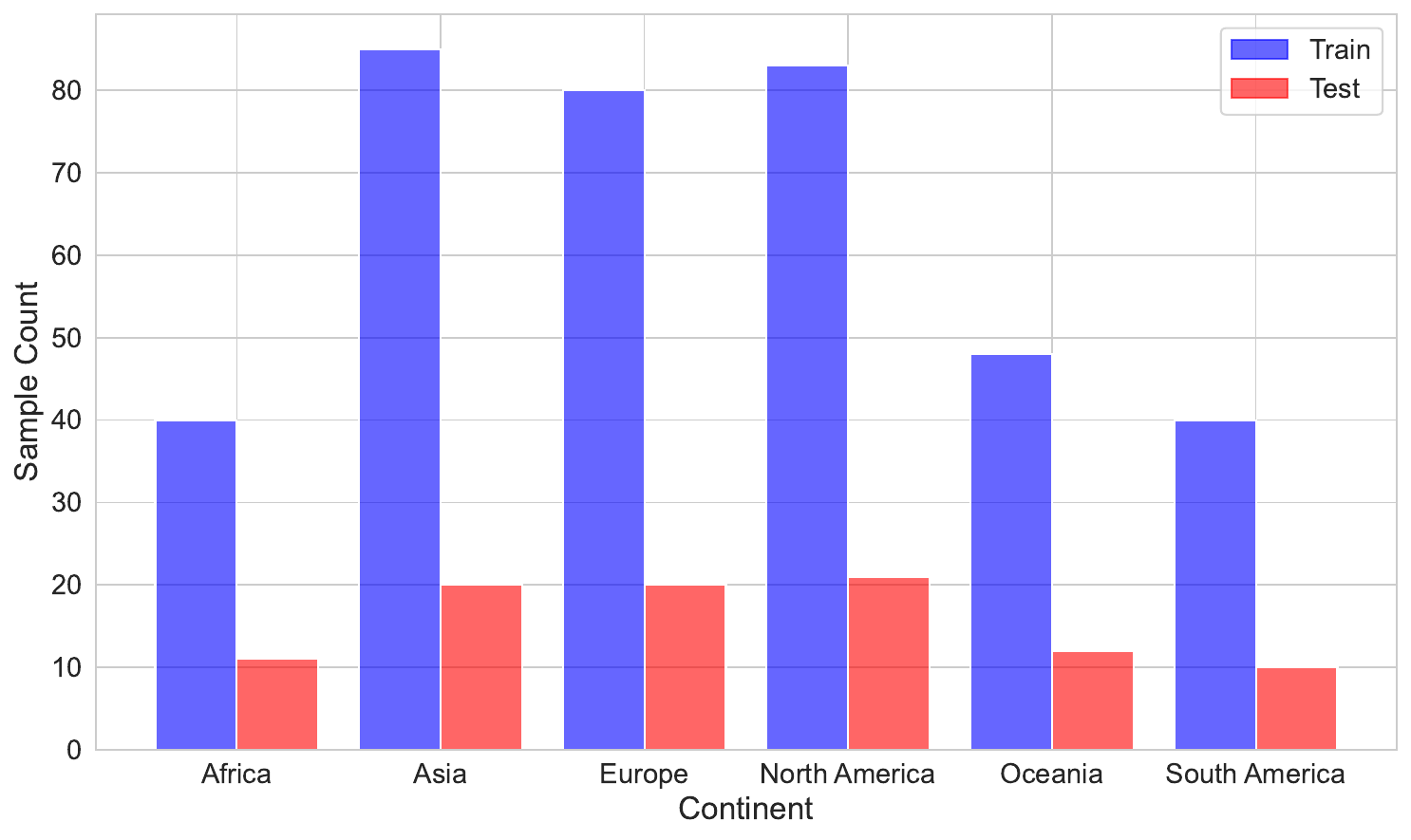}}
    \caption{Sample distributions in the dataset. (a) Number of images per environment type (Urban/Peri-urban, Rural) and camera angle (With Sky, Without Sky) for the training and test sets. (b) Number of images per continent for the training and test sets.}
    \label{fig:sampleditr}
\end{figure}

\begin{figure}[t!]
    \centering
    \includegraphics[width=0.5\textwidth]{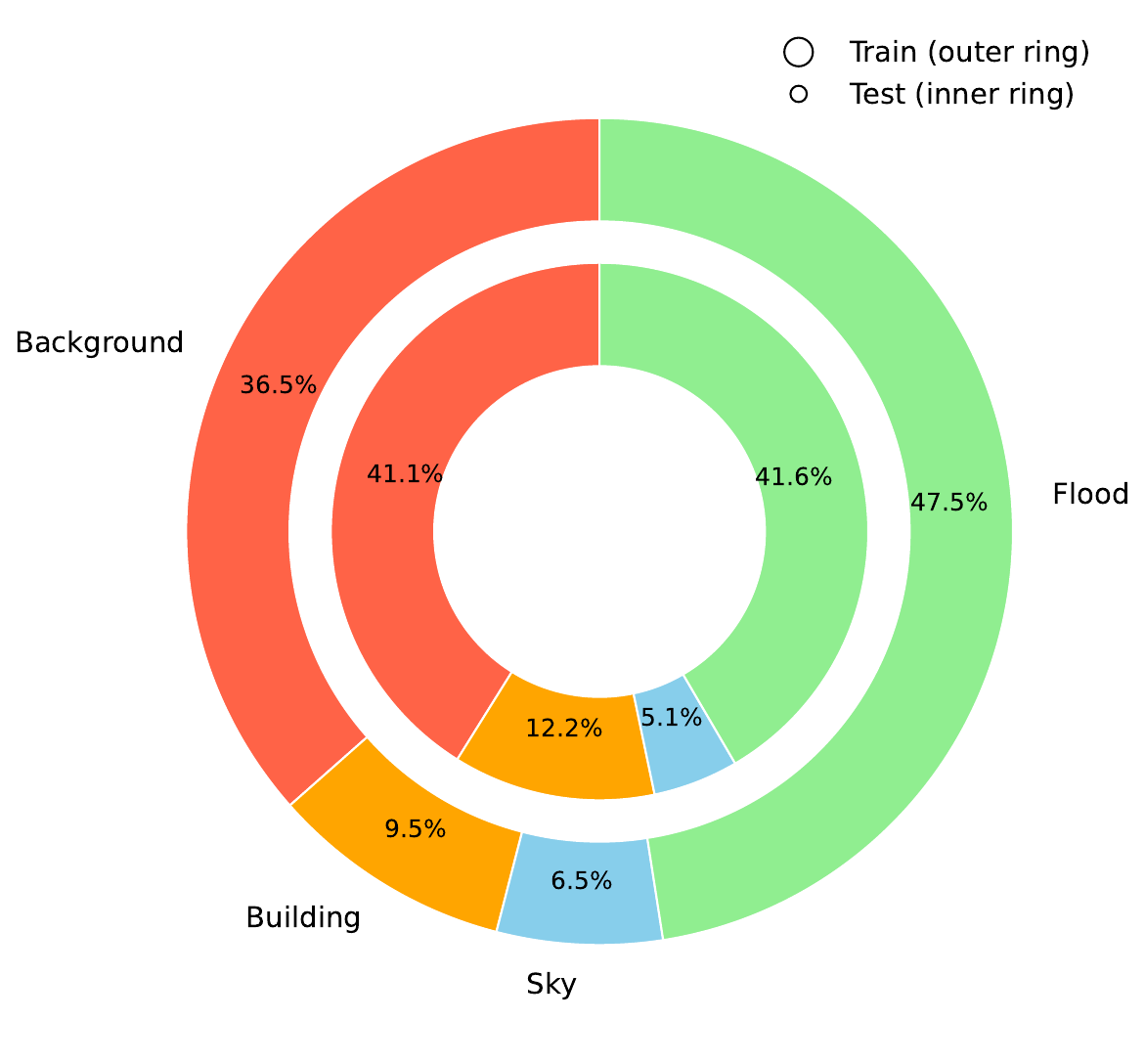}\\
    \caption{Pixel-wise class distribution in the dataset. Outer and inner rings represent the training and test sets, respectively, showing the proportion of pixels for each class (Background, Building, Sky, Flood).}
    \label{fig:classditr}
\end{figure}

The dataset supports three primary computer vision tasks: (1) \textbf{Image Classification}, featuring subtasks for environmental type (urban/peri-urban vs. rural), camera view angle (sky presence vs. absence) and continent recognition; (2) \textbf{Semantic Segmentation}, with pixel-level annotations for flooded regions, sky, buildings, and background, which support both multi-class and binary segmentation setups; and (3) \textbf{Visual Question Answering (VQA)}. To ensure a rigorous evaluation, the data were partitioned into training and testing sets using an 80/20 stratified split. This stratification algorithmically balances the distribution of environmental contexts, camera angles (Fig.~\ref{fig:sampleditr} (a)), geographical regions (Fig.~\ref{fig:sampleditr} (b)) and semantic class pixels (Fig.~\ref{fig:classditr}) between both subsets. All images were standardized to dimensions of 1024×768 pixels to facilitate deep learning model training. The complete dataset, including imagery, ground truth masks and metadata (event date, geolocation, environmental labels, and source URLs), is  publicly available at 
\url{https://sites.google.com/site/costaspanagiotakis/research/aifloodsense}.

\subsection{Annotation Protocol}
\label{subsec:annoprot}

\subsubsection{Classification \& Semantic Segmentation}
Pixel-wise annotations were created for the four semantic classes: flooded regions, buildings, sky, and background. Additionally, each image was annotated with the name of the continent in which the flood event occurred. The annotation process followed a two-stage protocol designed to ensure accuracy and consistency. Three trained annotators produced initial annotations using open-source data-labeling software Label Studio~\cite{LabelStudio}, following a set of detailed guidelines that specified class definitions, edge cases (e.g., reflections, shadows, horizon ambiguities), and labeling conventions. Each annotated image was then reviewed by two senior supervisors, who evaluated the masks, flagged inconsistencies, and provided targeted feedback. Annotators iteratively refined their work until supervisor approval was reached. Only supervisor-approved masks and classification labels were included in the final dataset.

\begin{figure*}[t!]
    \centering
    \includegraphics[width=0.95\textwidth]{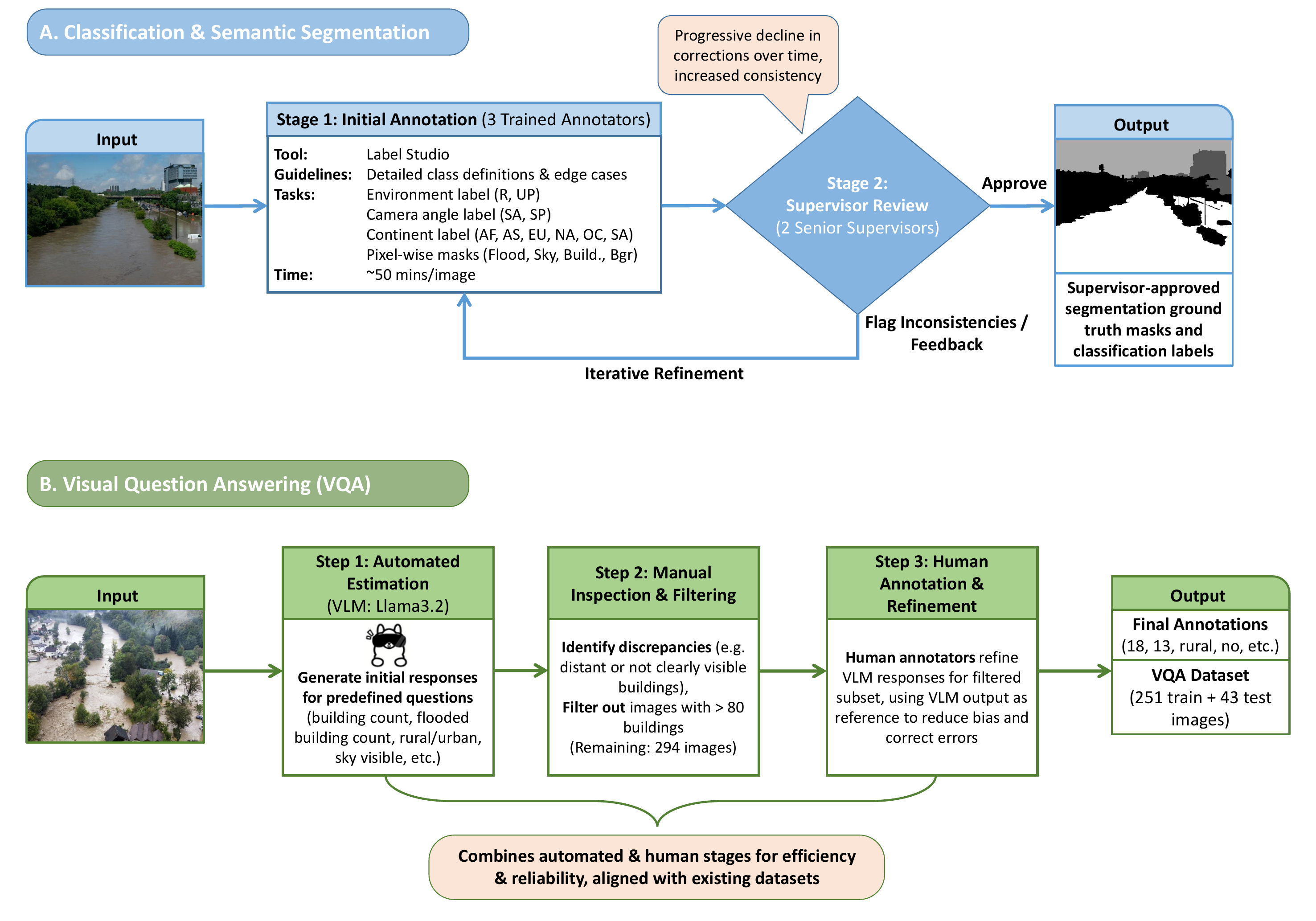}
    \caption{Schematic overview of the the annotation protocol for A. Classification and Semantic segmentation, and B. Visual Question Answering (VQA). Abbreviations used for classification labels and segmentation classes: R (rural), UP (urban/peri-urban), SA (sky absence), SP (sky presence), AF (Africa), AS (Asia), EU (Europe), NA (North America), OC (Oceania), SA (South America), Build (Building), Bgr (Background).}
    \label{fig:schemoverv}
\end{figure*}

\begin{figure}[h]
    \centering
    \includegraphics[width=0.5\textwidth]{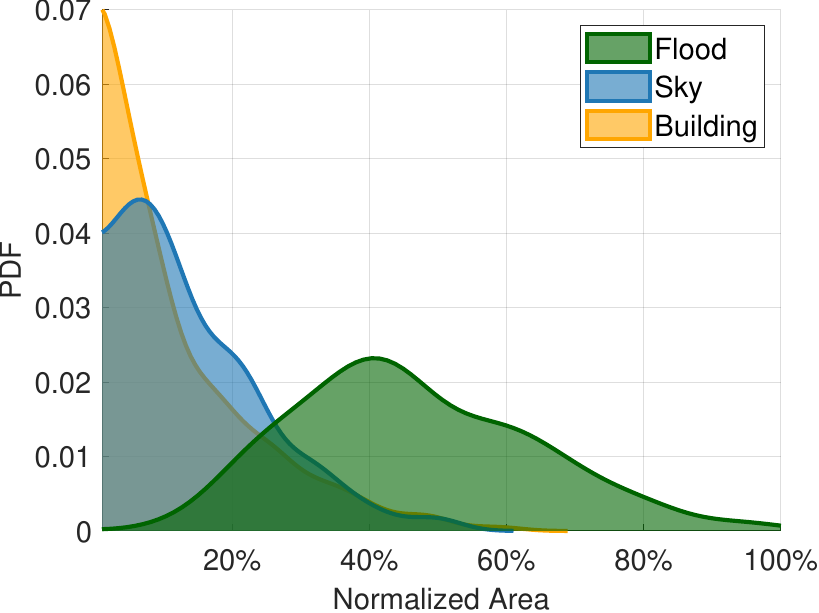}
    \caption{Estimated probability density functions (PDF) of normalized area proportions for the Flood, Sky, and Building classes, representing pixel coverage across the dataset.}
    \label{fig:kdesegmlabels}
\end{figure}

The complete annotation protocol is depicted in Figure~\ref{fig:schemoverv} A. On average, the initial annotation of a single image required approximately \textbf{50} minutes on average. Supervisor review time varied with image complexity, but we observed a progressive decline in corrections over time. Early in the annotation process, supervisors frequently identified mislabeled or ambiguous regions, but as guidelines were refined and annotators became more familiar with the task, the rate of disagreement between annotators and supervisors decreased sharply, resulting in faster approval cycles and greater consistency. This iterative, feedback-driven workflow ensured that the final dataset achieves both high annotation quality and protocol reproducibility. The distribution of pixel proportions for the segmented classes (Flood, Sky, and Building) is characterized via kernel density estimation in Figure~\ref{fig:kdesegmlabels}. The PDFs effectively capture the distribution peaks, particularly for the Sky and Building classes which are concentrated at lower coverage levels, while highlighting the distinct variability in spatial distribution for the Flood class.

\subsubsection{Visual Question Answering (VQA)}
For the VQA task we formulated five key questions that are representative of the essential semantic attributes required for flood-scene understanding: (a) \textit{how many buildings are present in the image}, (b) \textit{whether the image depicts a rural or urban/peri-urban region}, (c) \textit{whether the image contains visible sky}, (d) \textit{whether the image is flooded}, and (e) \textit{how many buildings are flooded}. Our annotation process was designed to leverage both automated estimations from a Vision-Language Model (VLM) and human validation to ensure accuracy and consistency.

Initially, we posed these five questions to a VLM (Llama3.2~\cite{grattafiori2024llama}), which generated responses for each image. 
Upon manual inspection, we observed substantial discrepancies in cases where buildings appeared at varying depths in the scene, particularly when structures were present both in the foreground and near the horizon. The latter case often led to systematic under- or over-counting, as the VLM struggled to distinguish distant buildings from background elements or occlusions.

To assess the extent of this issue, we selected a subset of images exhibiting such ambiguities and asked two independent human annotators to count the buildings. The results revealed significant inter-annotator disagreement, highlighting the inherent difficulty of accurately estimating building counts in such cases. Based on this observation, we decided to exclude highly ambiguous cases from the dataset to mitigate annotation noise. Specifically, we set a threshold of 80  buildings: images where the estimated count exceeded this limit were filtered out. This resulted in a set of 294  images. 

Following this filtering step, we conducted a second round of human annotation, refining the process by providing annotators with the initial VLM-generated responses for the filtered subset as a reference. This approach was intended to reduce subjective biases and encourage consistency across annotations while maintaining human oversight to correct systematic errors from the VLM. By incorporating both automated and human-driven annotation stages, we aimed to balance efficiency with reliability, ensuring a high-quality dataset for VQA-based flood scene understanding.

After this refinement process, we obtained a final subset of \textbf{251} train and \textbf{43} test images that support the VQA task. The selection of the five VQA questions was motivated by existing datasets \cite{FloodNet2021} addressing similar problems, ensuring alignment with prior work in the domain of flood impact assessment and disaster response. The annotation pipeline for the VQA task is illustrated in Figure~\ref{fig:schemoverv} B.

\subsection{Details on Supported Tasks}
\label{subsec:tasks}

The proposed dataset, \textit{AIFloodSence}, is designed to support three complementary computer vision tasks that together facilitate a comprehensive understanding of UAV-based flood imagery: (a) image-level classification, (b) semantic segmentation, and (c) visual question answering (VQA). The classification task encompasses several scene-level categories, such as rural vs. urban environments, sky presence, and geographic location, each offering valuable contextual cues to inform and enhance flood detection performance. The semantic segmentation task provides fine-grained pixel-level annotations that delineate flooded versus non-flooded regions, enabling precise quantification of flood extent, which is crucial for emergency response and resource allocation. The VQA component introduces a higher-level reasoning challenge, requiring models to interpret complex visual information in response to natural language questions related to flood conditions, scene elements, and potential hazards. Together, these tasks make \textit{AIFloodSense} a versatile and richly annotated benchmark for advancing multi-task learning, scene understanding, and decision support in real-world disaster management

\subsubsection{Classification Task}
\label{ssubsec:cltask}
 \noindent \textbf{Rural versus Urban/peri-urban Classification:} The rural vs. urban/peri-urban classification task provides essential insight into the land cover and built environment characteristics of the scene. Urban areas often exhibit complex textures, dense structures, and distinct water drainage patterns compared to rural or agricultural regions. As a result, models trained to segment flooded areas can benefit from this prior knowledge to better adapt their predictions based on the surrounding context, particularly under domain shifts or limited supervision.  For instance, floods in urban regions may require different mitigation strategies and present different visual characteristics compared to rural areas. As such, improving the classification of land type can contribute to more precise and context-aware flood segmentation models.\\

\noindent \textbf{Sky absence versus Sky presence Classification:} The sky absence vs. sky presence classification task is included to account for viewpoint and framing variability in UAV imagery. The presence of the sky typically implies a lower ground coverage ratio and may affect lighting conditions, contrast, and visibility of flooded regions. Identifying such scenes can inform pre-processing strategies, filtering mechanisms, or the design of models that adapt differently to wide-angle versus nadir shots.\\

\noindent \textbf{Continent Classification:} The proposed dataset introduces the novel task of continent classification. This task can be considered as a coarse form of geographic domain adaptation for UAV flood image understanding methods. The appearance of floods and the features of the landscape vary significantly between continents due to differences in architecture, vegetation, soil, and climate. By explicitly modeling the geographic domain, learning-based methods can leverage regional priors and improve generalization in diverse deployment settings. Collectively, these classification tasks enrich the dataset's utility by enabling multi-task learning, contextual scene understanding, and enhanced segmentation performance under real-world conditions.

\subsubsection{Semantic Segmentation Task}
\label{ssubsec:sstask}
Semantic segmentation of UAV images plays a pivotal role in remote sensing and disaster response applications by enabling pixel-level scene understanding from aerial perspectives. Inspired by the associated segmentation task in recent datasets (see Table~\ref{tab:dspresentation}), in the proposed \textit{AIFloodSence} dataset we support the semantic segmentation of three critical classes: (a) \textit{flood region}, (b) \textit{sky}, and (c) \textit{building}. Each UAV image in our dataset is densely annotated at the pixel level to support these categories. The segmentation task is formally defined as follows: given an RGB aerial image captured by UAV, the objective is to assign to each pixel one of the three semantic labels, \textit{flood region, sky, or building}, or designate it as background if none apply. This formulation supports use cases in post-disaster monitoring, infrastructure mapping, and environmental assessment.

The inclusion of the \textit{sky} class, in particular, helps disambiguate sky regions often prone to be misclassified as flood or building in datasets lacking this distinction. The \textit{flood region} class emphasizes submerged areas critical for damage assessment, whereas the \textit{building} class focuses on urban infrastructure potentially affected by flooding. By incorporating these three classes, our dataset facilitates targeted segmentation tasks relevant to both environmental monitoring and post-disaster response.

\subsubsection{VQA Task}
\label{ssubsec:vqatask}
The VQA annotations are designed to promote scene-level understanding by addressing complementary aspects of the data beyond pixel-wise predictions. Specifically, they span topics such as flood presence assessment, structural instance (building) counting and condition estimation (flooded or non-flooded), environmental characterization of the scene (urban or rural), and the identification of atmospheric or contextual elements (sky presence). The selected questions where presented in Section~\ref{subsec:dschar}.

The inclusion of VQA provides an additional dimension of analysis, requiring models not only to recognize objects and regions but also to reason about the scene and link visual evidence to semantic concepts. This task is particularly valuable as it bridges low-level perception with high-level reasoning, enabling the assessment of whether models can move beyond toward interpretable and actionable outputs. 

\section{Implementation}
\label{sec:impl}
To establish a controlled benchmark for evaluating baseline models, All models were implemented within the PyTorch framework, and trained under identical experimental settings to ensure a fair evaluation of their performance for each task on the proposed dataset, while for the pre-trained models the default weights were utilized. All experiments were carried out on a system equipped with an Intel i7 CPU (2.3 GHz), 40 GB of RAM, and two NVIDIA Quadro RTX 4000 GPUs. Depending on the task and model configuration, training times ranged from less than one hour to several hours in all experiments. 
For all tasks, a consistent 80\% training and 20\% testing split was applied. In the following subsections, we detail the task-specific configurations adopted for training and evaluation of the baseline models.

\subsection{Classification Task Configurations}
\label{subsec:classific_configs}
\noindent \textbf{Baseline Models:} For the classification tasks, we trained a set of five well-established image classification architectures, comprising three convolutional neural networks and and two transformer-based models. CNN selection included the residual networks ResNet18 and ResNet50~\cite{he2016deep}, alongside the resource-efficient EfficientNet-B0~\cite{EfficientNet}, which uses compound scaling. Transformer-based models consisted of the standard Vision Transformer (ViT)~\cite{dosovitskiy2020image}, specifically the vit\_base\_patch16\_224, and the hierarchical Swin-T~\cite{liu2021swin}, specifically the swin\_tiny\_patch4\_window7\_224 variant, which utilizes shifted window attention to enhance efficiency.\\

\noindent \textbf{Train/Test Configurations:} Experiments were conducted using both pre-trained weights and training from scratch. To accommodate the distinct architectural requirements of the evaluated models, we employed architecture-specific resizing strategies. For Transformer-based models, the images were resized to a square resolution of $224 \times 224$ pixels. This dimension aligns with the standard input requirements for patch-based embedding layers and ImageNet pre-training, while simultaneously optimizing computational efficiency during training. For the Convolutional Neural Networks (CNNs), we prioritized the preservation of the original scene geometry to avoid distortion artifacts. Consequently, inputs for these models were resized to $224 \times 168$ pixels. This resolution maintains the native aspect ratio of the dataset while aligning the spatial width with the input of the transformer to ensure experimental comparability.

To enhance model generalization, we employed a robust data augmentation pipeline consisting of both geometric and photometric transformations. Geometric augmentations included random horizontal flips, rotations within the range of \([-15^\circ, 15^\circ]\), and random affine transformations with a scaling factor between [0.85, 1.15]. Photometric distortions were applied via color jitter with brightness, contrast, and saturation factors of 0.2 and a hue factor of 0.1. All input images were normalized using standard ImageNet mean ($\mu = [0.485, 0.456, 0.406]$) and standard deviation ($\sigma = [0.229, 0.224, 0.225]$) statistics. The models were optimized using the Adam algorithm (weight decay = \(1\times10^{-4}\)) to minimize Cross-Entropy loss. We utilized a step-based learning rate scheduler that decays the learning rate by a factor of 0.1 every 10 epochs (initial learning rate = \(1\times10^{-4}\)). Training was conducted with a batch size of 16 for a total of 100 epochs, with the optimal model weights selected based on the highest recorded training accuracy.

\subsection{Semantic Segmentation Task Configurations}
\label{subsec:sem_seg_configs}

\noindent \textbf{Baseline Models:} For segmentation tasks, we evaluated a set of five representative architectures, including three models based on convolutional neural networks. (DeepLabV3~\cite{DeepLab}, FCN-ResNet50[~\cite{FCN}, and U-Net~\cite{Unet}) and two transformer-based models (SegFormer-B0~\cite{Segformer} and Swin-T with UPerNet (Swin-T)~\cite{liu2021swin}). CNN-based models were selected for their strong performance in dense prediction tasks and wide adoption in remote sensing applications, while transformer-based models were included to assess the benefits of long-range contextual modeling. All models were trained and tested under identical experimental conditions to enable a fair comparison of their performance on the proposed dataset. \\

\noindent \textbf{Train/Test Configurations:}
 The input images for the semantic segmentation task were standardized to a resolution of \(512 \times 384\) pixels and normalized using ImageNet mean and standard deviation statistics. To improve generalization, we employ a data augmentation pipeline comprising random horizontal flipping, rotation within the range of \([-15^\circ, 15^\circ]\), and photometric color jitter (brightness, contrast, saturation factors of 0.2 and a hue factor of 0.1). The models were trained for 100 epochs with a batch size of 8, minimizing the Dice Loss function. The optimization strategy was tailored to the architecture: the Adam optimizer was utilized for Convolutional Neural Networks (CNNs), while AdamW was employed for Transformer-based models.

For CNN models trained from scratch, the weights were initialized with Kaiming normal initialization, optimized with Adam (weight decay = \(1\times10^{-4}\)), and a step learning rate scheduler (initial learning rate = \(1\times10^{-4}\), step size = 10, decay factor = 0.5). For pre-trained CNNs, the head was reinitialized using Xavier normal initialization and optimized with Adam (weight decay = \(1\times10^{-4}\)), while the learning rate was scheduled with cosine annealing (\(T_{\text{max}} = 150\), \(\eta_{\min} = 1\times10^{-6}\)). Pre-trained weights were initialized from ImageNet-1K. For Transformer-based models, SegFormer-B0 weights were randomly initialized when trained from scratch, while Swin-T weights were initialized with Kaiming normal. Optimization was performed with AdamW (weight decay = \(1\times10^{-2}\)) and a cosine learning rate schedule with warmup (5,640 warmup steps out of 56,400 total). Pre-trained backbones included \texttt{swin\_tiny\_patch4\_window7\_224} (ImageNet-1K) and \texttt{segformer-b0-finetuned-ade-512-512} \\ (ImageNet-1K and ADE20K). Best-performing weights were selected based on maximum training accuracy.  

\subsection{VQA Task Configurations}
\label{subsec:vqa_configs}
\noindent \textbf{Baseline Models:} 
We evaluated several representative Vision-and-Language models to establish strong baselines for the VQA task. Specifically, we employed (a) \textit{ViLT}~\cite{kim2021vilt}, a lightweight transformer architecture that performs early fusion of vision and language features; and (b) \textit{BLIP-2}~\cite{li2023blip}, both in its frozen (zero-shot) form and in a fully fine-tuned configuration tailored to our dataset. In addition, we assessed the performance of larger, general-purpose multimodal LLMs, namely \textit{Gemini 2.5 Flash}~\cite{comanici2025gemini} and \textit{LLaMA 3.2 Vision}~\cite{grattafiori2024llama}, to (a) evaluate the performance of modern foundation models on the specific task supported by our dataset, and (b) to compare their performance against domain-specific VLMs for the specific task.\\

\noindent \textbf{Train/Test Configurations:} 
For the ViLT model, we used the official \texttt{vilt-b32-mlm} checkpoint as the initialization point. Training was conducted on our domain-specific VQA dataset using a batch size of 72, a learning rate of $5\times10^{-6}$, and a total of 60 epochs. We followed the standard ViLT training protocol, employing the provided multimodal data-processing pipeline without architectural modifications.

For BLIP-2, we adopted a parameter-efficient fine-tuning strategy using LoRA. Specifically, we applied rank-$8$ LoRA adapters with $\alpha=16$ and a dropout rate of 0.05 on the \texttt{q\_proj} and \texttt{v\_proj} modules of the OPT language model. 
The model was trained for 15 epochs with a learning rate of $1\times10^{-4}$ while keeping the vision encoder frozen. 
We trained the fine-tuned BLIP-2 model using our custom VQA dataset, following the same train/test split used for ViLT.

Gemini 2.5 Flash and LLaMA 3.2 Vision were evaluated via API-based prompting without any fine-tuning. For each model, we report performance across the same test split and with identical evaluation metrics.\\

\section{Experimental Results}
\label{sec:resdis}

In this section, we present the comprehensive results of the classification, segmentation, and VQA experiments. These are reported as quantitative performance scores and are supplemented by qualitative visualizations and an error analysis to provide a complete evaluation of the baseline methods. All results are reported in the test set.

\subsection{Classification Task}
\label{subsec:classific}

For quantitative evaluation, we employed standard performance metrics to assess the quality of the annotated labels. Specifically, for  binary classification tasks distinguishing urban/peri-urban versus rural regions, and sky absence versus sky presence, we report Accuracy (Acc.) alongside Precision (Pr.), Recall (Rec.), and F$_1$ score (F$_1$) to provide a comprehensive assessment of classification performance. Regarding the continent selection task, which is more challenging due to its multi-class nature and the inherent imbalance across classes, we report macro-averaged Precision, Recall, and F$_1$ scores to provide a clearer, more objective assessment of model performance across all classes, regardless of class size. Macro F$_1$ scores were computed as the unweighted mean of class-wise F$_1$ scores. 

Beyond standard performance metrics, we employ Entropy as a measure of predictive uncertainty, quantifying the dispersion of the softmax probability distribution. Low entropy indicates high model confidence, characterized by a probability distribution peaked around a single class, whereas high entropy reflects uncertainty, corresponding to a more uniform distribution of probabilities across classes. To ensure comparability between tasks with varying numbers of classes, we utilize the normalized Entropy ($H_{norm}$). By scaling the entropy relative to the maximum possible uncertainty, $H_{norm}$ provides a task-agnostic metric that, when analyzed in conjunction with Confidence (the maximum predicted probability), allows for a robust assessment of model calibration and reliability.

\noindent \textbf{Rural/Urban and peri-urban Classification:} As stressed in the previous section, recognizing the type of environment provides essential contextual information that can guide downstream tasks such as flood extent segmentation, resource allocation, damage assessment and the planning of rescue operations, where situational awareness and prioritization of affected areas are critical.

Table~\ref{tab:urbanruralclf} presents the results of baseline models for the task of urban/peri-urban versus rural environment classification. The results decisively validate the effectiveness of transfer learning across all evaluated architectures. Every model, regardless of its type (CNN or Transformer), exhibited a significant improvement when initialized with pre-trained weights. Specifically, the ViT model benefited the most, showing the largest absolute gain in the F$_1$ score, moving from 76.82\% (scratch) to a leading 85.07\% (pre-trained). This 8.25 percentage point gain underscores the necessity of leveraging generalized features learned from larger datasets for models with weak inductive biases. In contrast, while the Swin-T model also benefits, its scratch performance (F$_1$ = 81.08\%) is comparatively high, suggesting that its hierarchical architecture makes it the most data-efficient and robust for small-scale training, achieving a smaller, but still notable, gain of 1.53 percentage points.

Among pre-trained models, ViT achieved the highest F$_1$ score (85.07\%) and tied for the highest Accuracy (78.72\%), confirming its superior capability in capturing global dependencies relevant to the classification task of distinguishing urban/peri-urban vs. rural environments. The EfficientNet-B0 closely followed, achieving an F$_1$ score of 84.62\% and also tying for the top Accuracy, demonstrating the efficacy of its compound scaling approach. Notably, the Recall metric is consistently high across all models, which suggests a low rate of False Negatives, meaning that the models rarely miss a true positive instance. The primary performance variation stems from Precision, indicating that the models primarily struggle with False Positives, likely due to feature ambiguity in the dataset. The presence of substantial visual heterogeneity due to variations in terrain types, built infrastructure, floodwater coverage, and environmental conditions across images indicates that the proposed dataset is not uniform and complex, and diverse patterns challenge model generalization.

\begin{table}[h!]
\centering
\caption{Baseline method performance on rural vs urban/peri-urban classification on the \textit{AIFloodSense} test set. Results are reported as percentages (\%), with and without ImageNet pretraining.}
\begin{tabular}{|l||c|c|c|c|c|}
\hline
\multirow{2}{*}{\textbf{Model}} & \multirow{2}{*}{\begin{tabular}[c]{@{}c@{}}\textbf{Pre-}\\\textbf{trained}\end{tabular}} & \multirow{2}{*}{\textbf{Acc.}} & \multirow{2}{*}{\textbf{Pr.}} & \multirow{2}{*}{\textbf{Rec.}} & \multirow{2}{*}{\textbf{F$_1$}} \\ 
 & & & & & \\
\hline
\hline
EfficientNet-B0   & \checkmark & \textbf{78.72} & \textbf{80.88} & 88.71 & 84.62 \\ \hline
ResNet18 & \checkmark & 77.66 & 78.87 & 90.32 & 84.21 \\ \hline
ResNet50 & \checkmark & 76.60 & 76.32 & 93.55 & 84.06 \\ \hline
Swin-T     & \checkmark & 74.47 & 75.00 & 91.94 & 82.61 \\ \hline
ViT      & \checkmark & \textbf{78.72} & 79.17 & 91.94 & \textbf{85.07} \\ \hline
\hline
EfficientNet-B0   & \ding{55}  & 70.21 & 70.24 & 95.16 & 80.82 \\ \hline
ResNet18 & \ding{55}  & 68.09 & 69.05 & 93.55 & 79.45 \\ \hline
ResNet50 & \ding{55}  & 65.96 & 67.44 & 93.55 & 78.38 \\ \hline
Swin-T     & \ding{55}  & 70.21 & 69.77 & \textbf{96.77} & 81.08 \\ \hline
ViT      & \ding{55}  & 62.77 & 65.17 & 93.55 & 76.82 \\ \hline
\end{tabular}
\label{tab:urbanruralclf}
\end{table}

\begin{figure}[t!]
    \centering
    \subfigure[Sorted Normalized Entropy, Confidence, and Error Locations.]{    \includegraphics[width=0.49\textwidth]{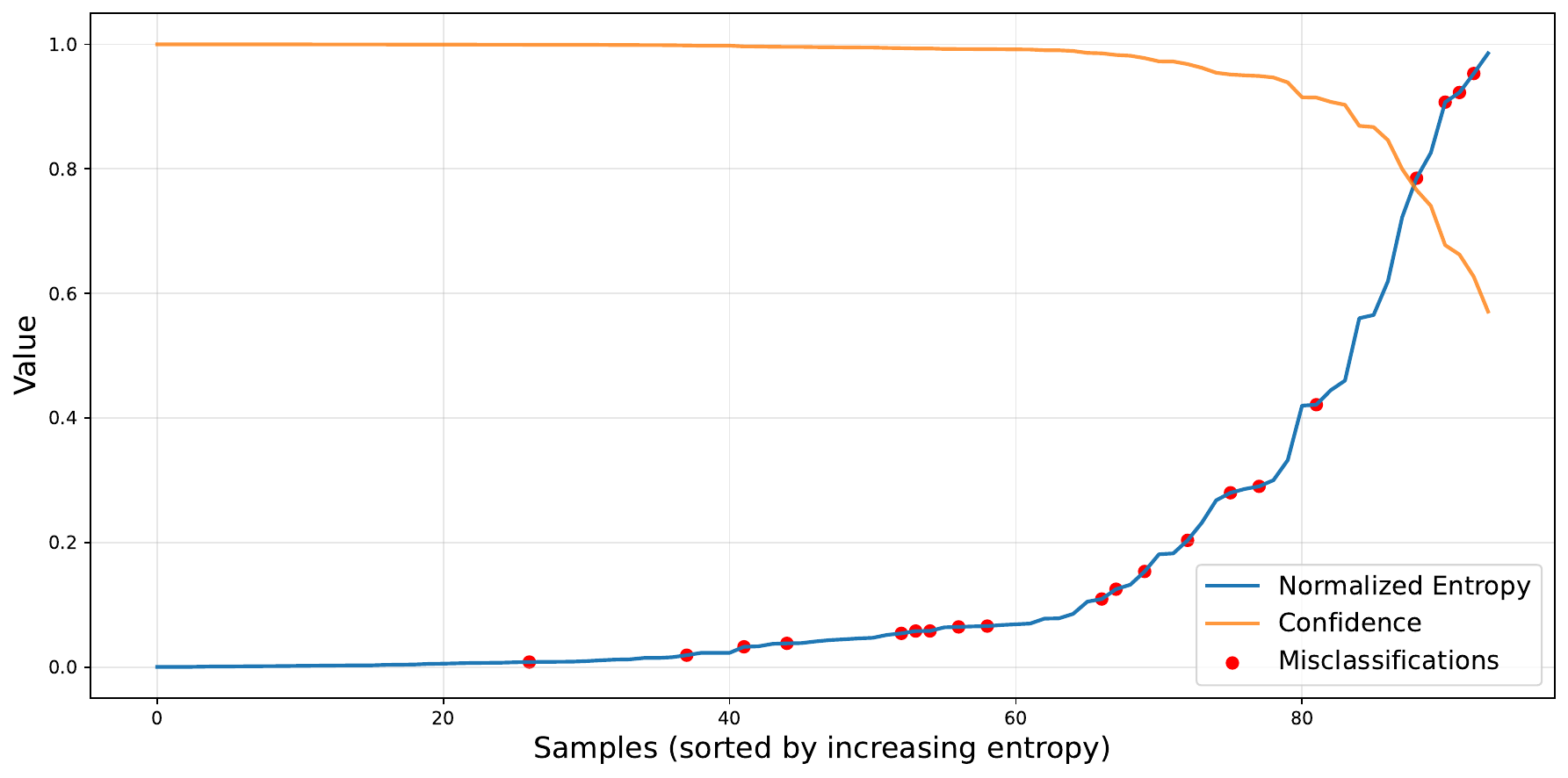}}
    \subfigure[Normalized entropy histogram.]{\includegraphics[width=0.49\textwidth]{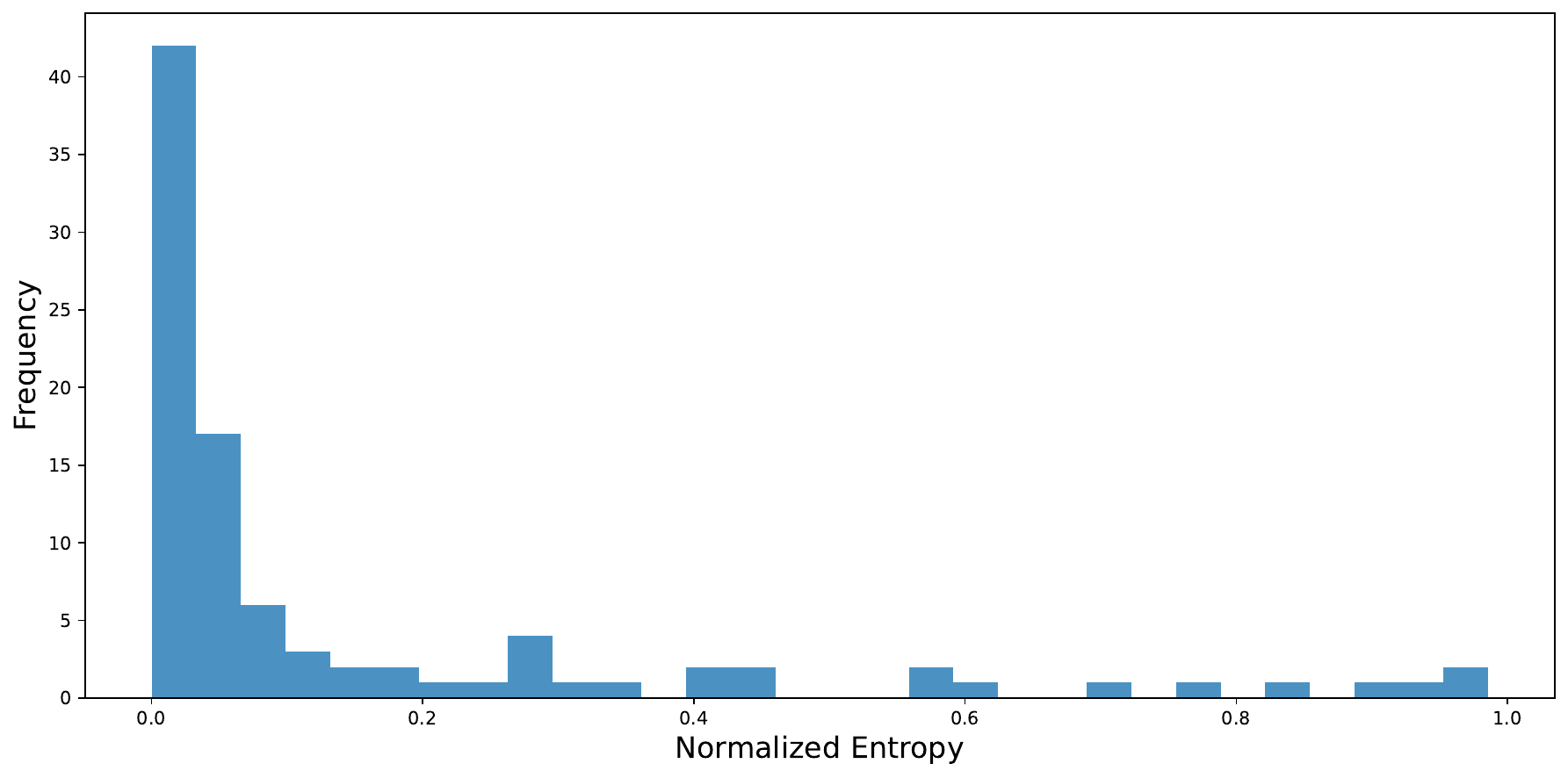}}\\
    \caption{Normalized entropy against prediction confidence with misclassifications and normalized entropy histogram for the \textit{AIFloodSense} test set of the best performing rural vs urban/peri-urban classifier, ViT pre-trained.}
    \label{fig:1_entr_conf_hist}
\end{figure}

To rigorously evaluate the reliability of the rural versus urban/peri-urban classifier, we use the berst performing model, ViT pre-trained, and analyze the relationship between predictive confidence and uncertainty using normalized entropy. Figure~\ref{fig:1_entr_conf_hist} (a) plots the normalized entropy against the prediction confidence for the test set, with samples sorted by increasing entropy. Misclassifications are overlaid as red markers. The sorted sample curve demonstrates a strong inverse correlation between confidence (orange line) and normalized entropy (blue line), validating the model's calibration. The plot reveals a subset of overconfident errors, visible as scattered red dots in the low-entropy region (normalized entropy < 0.2), suggesting that certain ambiguous samples, likely containing misleading features such as paved roads or sparse buildings in rural settings or vast vegetation in peri-urban areas, can trigger high-confidence false predictions. The accompanying histogram in Figure~\ref{fig:1_entr_conf_hist} (b) further characterizes the dataset's difficulty profile. The distribution is heavily skewed towards zero, with the vast majority of samples falling into the lowest entropy bin indicating that for a large portion of test set, the model extracts distinct, discriminative features that allow for decision-making with near-certainty, however there is a significant number of samples which present significant ambiguity.

\begin{figure*}[]
    \centering
    
    \subfigure[$H_{norm} = 0.0002$,\newline Pred = UP, GT = UP] {\includegraphics[width=0.24\textwidth]{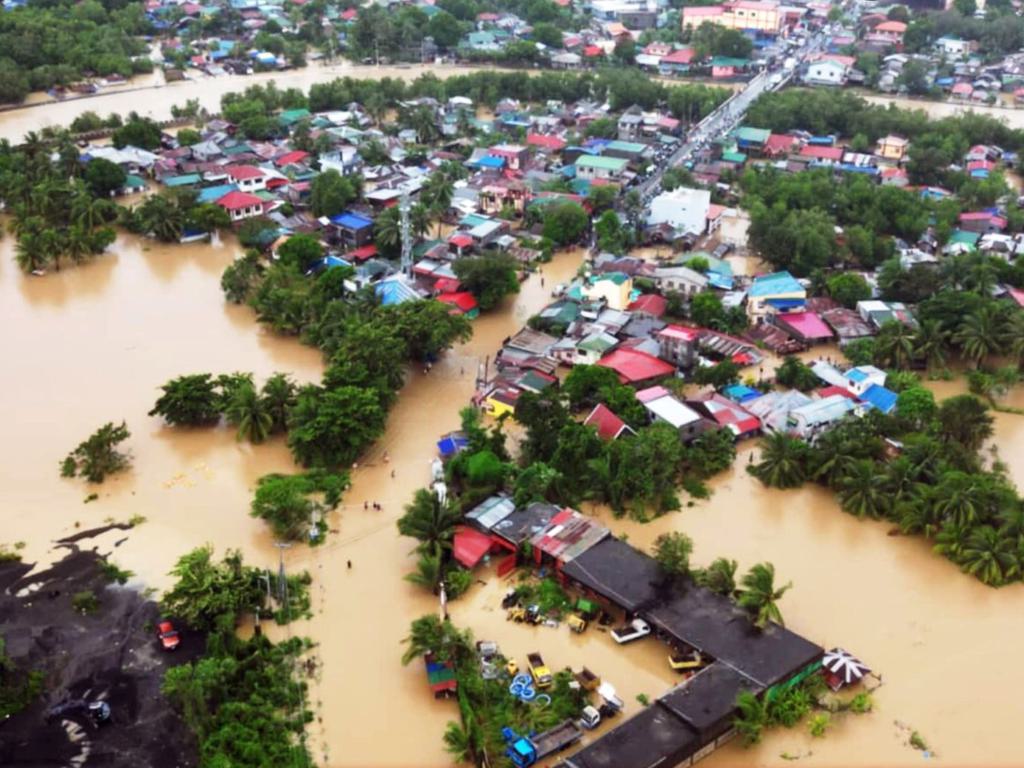}}
    \subfigure[$H_{norm} = 0.0002$,\newline Pred = UP, GT = UP]{\includegraphics[width=0.24\textwidth]{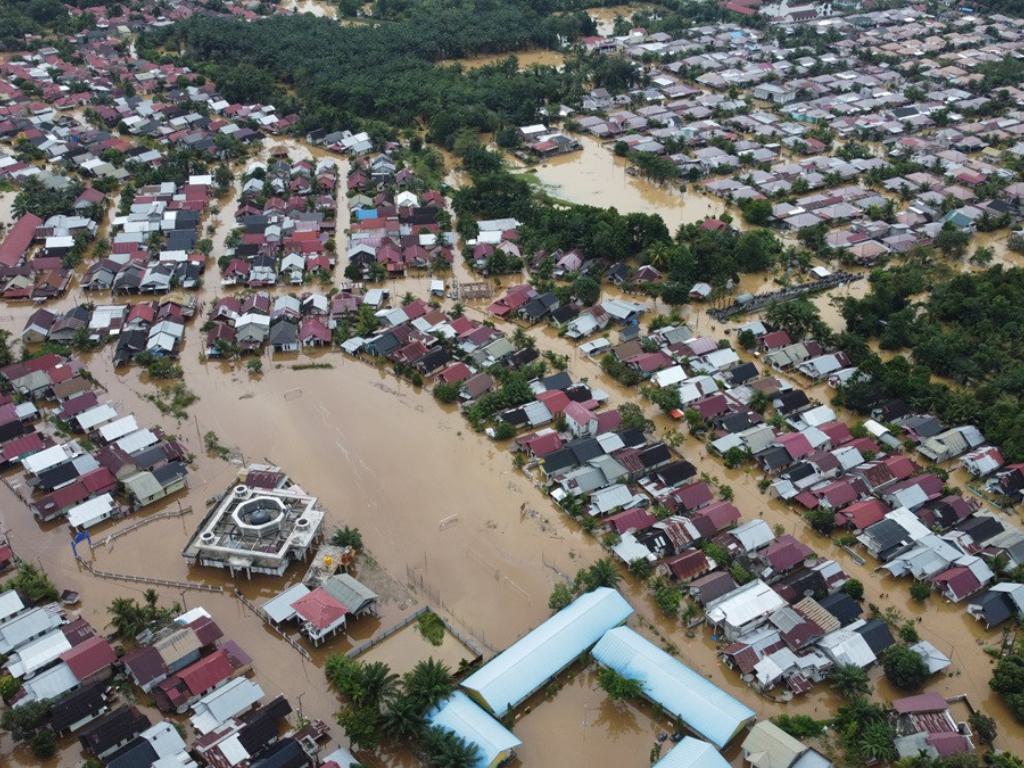}}
    \subfigure[$H_{norm} = 0.0002$,\newline Pred = UP, GT = UP] {\includegraphics[width=0.24\textwidth]{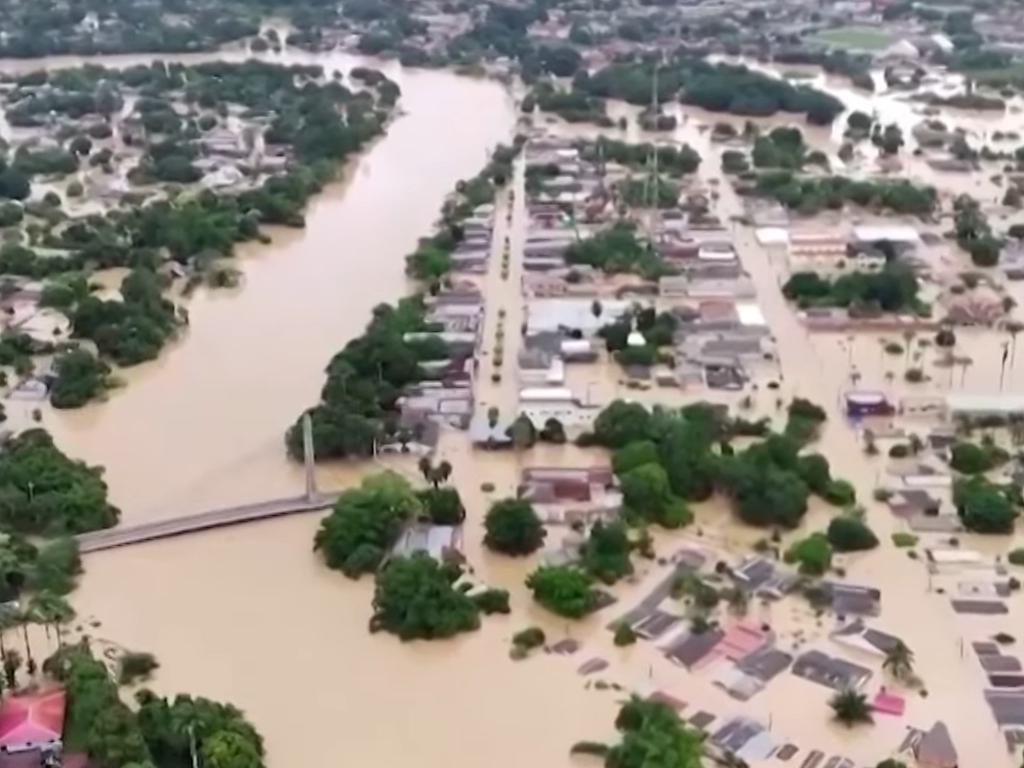}}
    \subfigure[$H_{norm} = 0.0006$,\newline Pred = UP, GT = UP] {\includegraphics[width=0.24\textwidth]{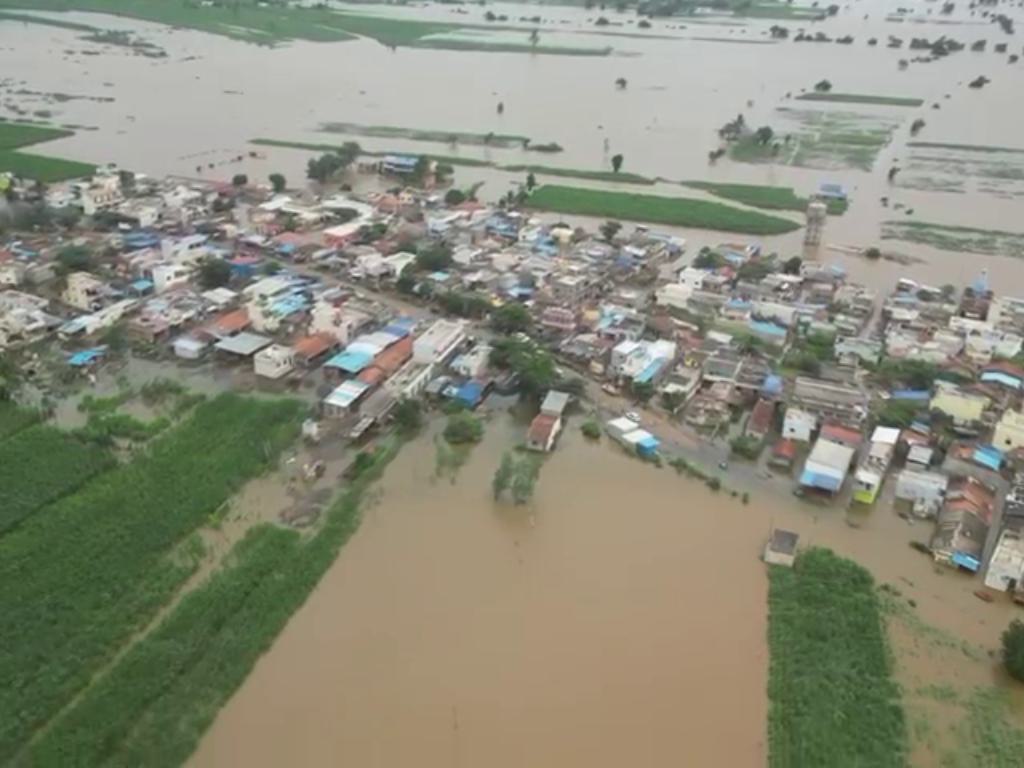}}
    \\

    \subfigure[$H_{norm} = 0.9071$,\newline Pred = UP, GT = R] {\includegraphics[width=0.24\textwidth]{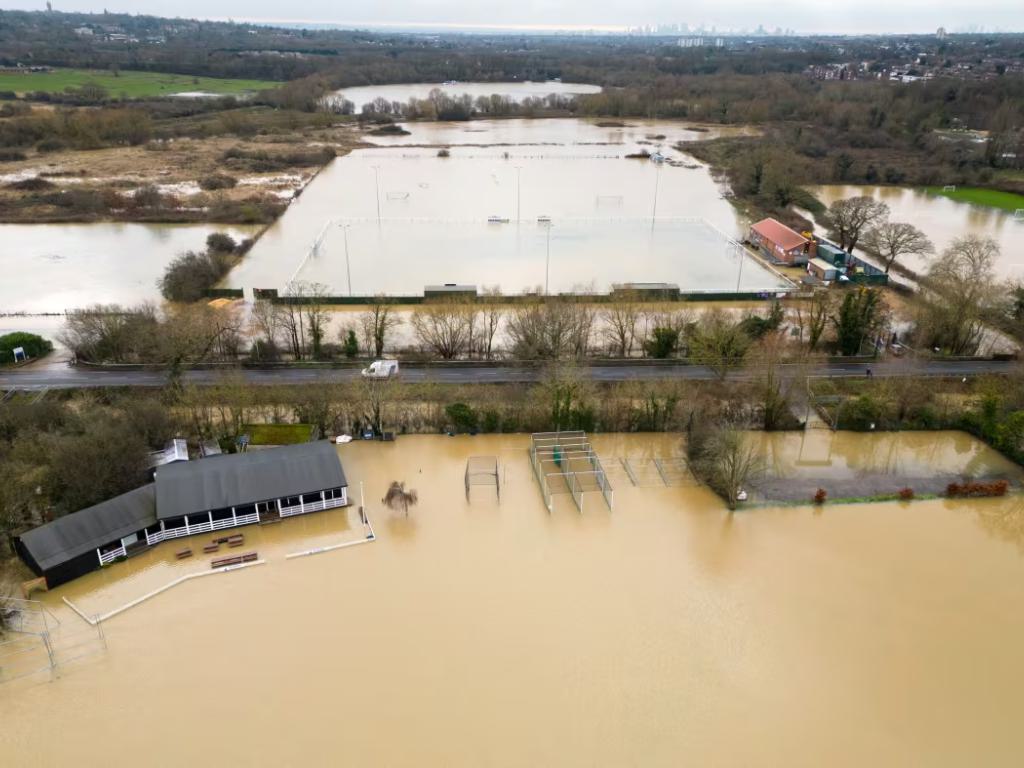}}
    \subfigure[$H_{norm} = 0.9226$,\newline Pred = UP, GT = R] {\includegraphics[width=0.24\textwidth]{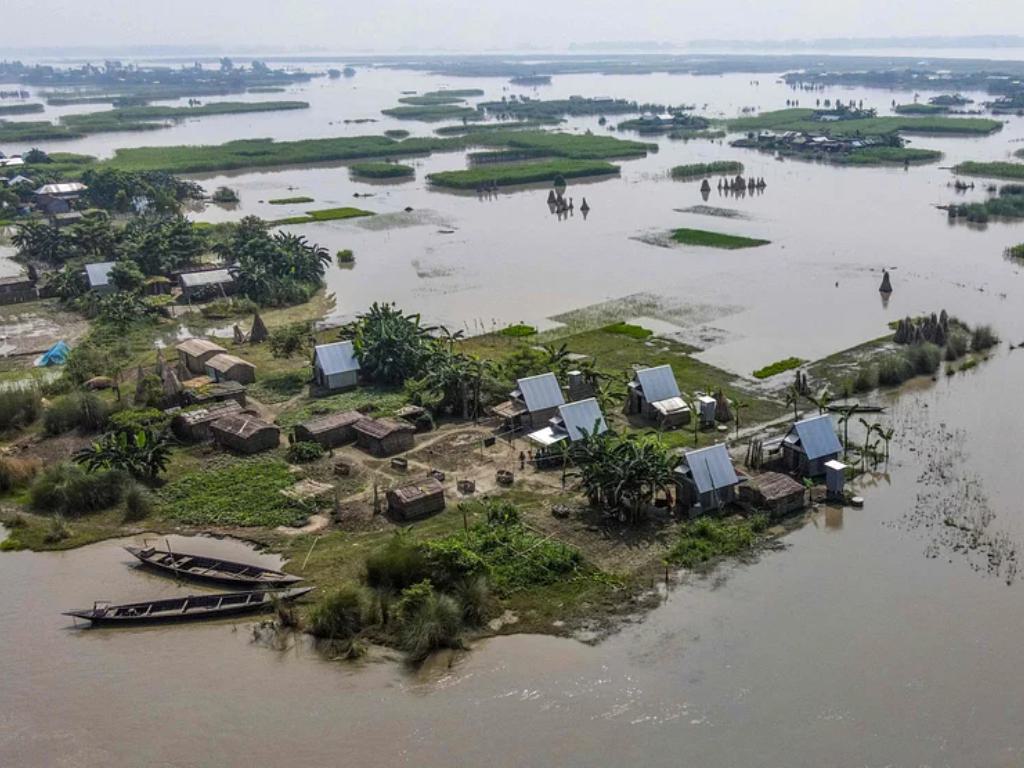}}
    \subfigure[$H_{norm} = 0.9531$,\newline Pred = UP, GT = R] {\includegraphics[width=0.24\textwidth]{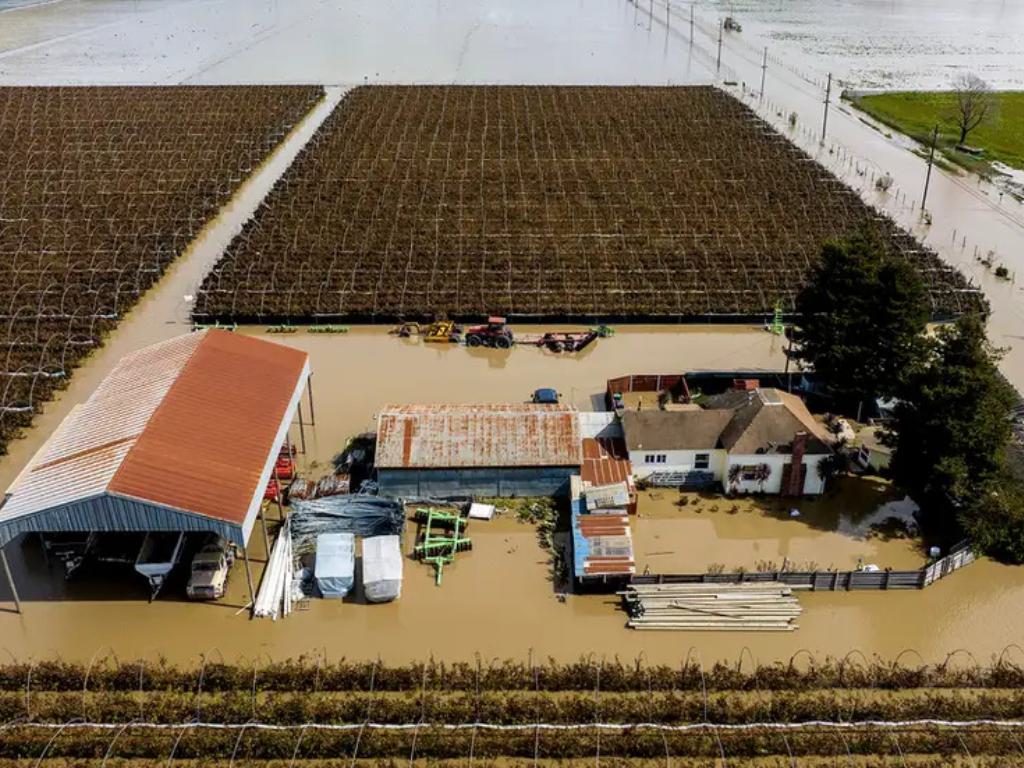}}
    \subfigure[$H_{norm} = 0.9855$,\newline Pred = UP, GT = UP] {\includegraphics[width=0.24\textwidth]{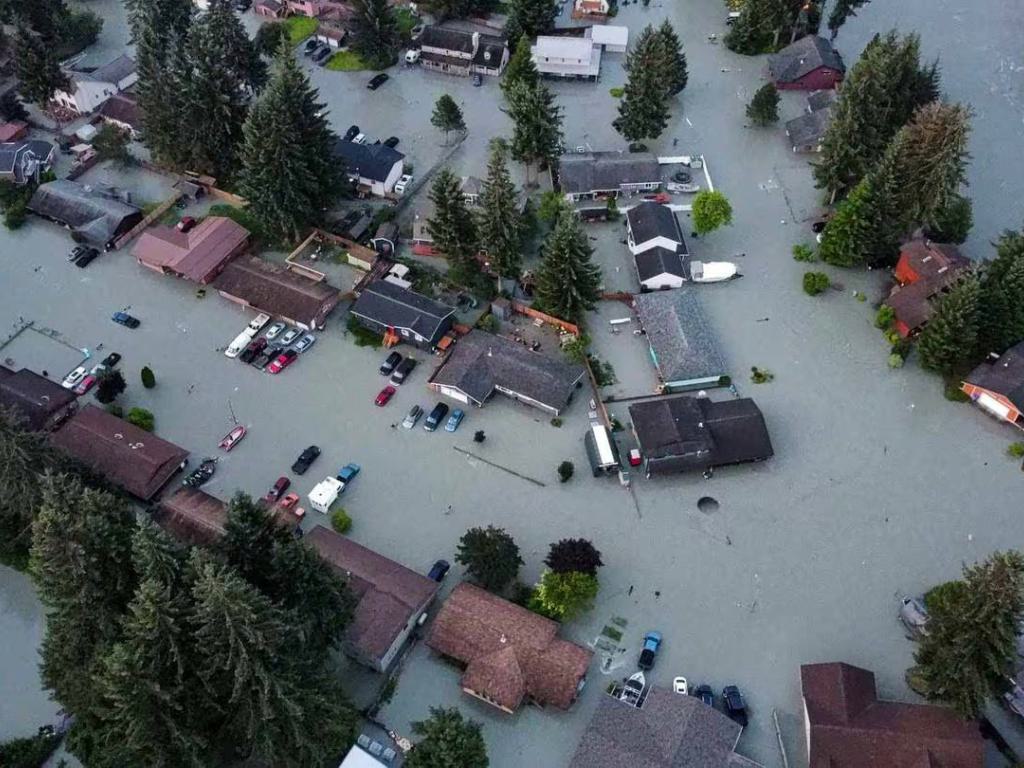}}
    \\
    
    \caption{Top 4(1st row) and bottom 4 (2nd row) samples with the lowest/highest normalized entropy ($H_{norm}$) for the \textit{AIFloodSense} test set of the best performing rural (R) vs urban/peri-urban (UP) classifier, ViT pre-trained. Pred=prediction, GT=ground truth.}
    \label{fig:1_top_bottom}
\end{figure*}

Qualitative insight is given in Figure~\ref{fig:1_top_bottom}, which displays the four samples with the lowest normalized entropy (most confident predictions) in the first row, contrasted with the four samples exhibiting the highest normalized entropy (most uncertain predictions) in the second row. The top 4 samples are characterized by high concentration of buildings, which presents a distinct environmental feature leading to their correct classification as urban/peri-urban (a)--(d). High-entropy samples typically represent ambiguous transition zones or visual occlusions where the distinction between rural and peri-urban environments is inherently less defined, as in the cases (e)--(g), where rural environments are misclassified as urban/peri-urban likely due to the sparse but visible housings.

\vspace{0.2cm}
\noindent \textbf{Sky absence/presence Classification:} The novel sky absence/presence classification task, serves as a valuable auxiliary cue for flood segmentation from UAV imagery, as the presence or absence of the sky can provide contextual information about scene geometry, viewpoint, and potentially occluded flooded areas. Accurate sky detection can enhance the robustness of models targeting the flood segmentation task by enabling improved spatial priors.

Table~\ref{tab:skynoskyclf} presents the evaluation of the baseline architectures under the two training regimes. The most prominent trend observed in the results is the substantial benefit of utilizing transfer learning for this specific task. Across all architectures, models initialized with pre-trained weights consistently outperform their counterparts trained from scratch. Specifically, there are gains in the observed F$_1$ score ranging from 3.06\% to 8.46\%, with especially transformer-based models profiting from transfer learning, allowing them to more effectively capture the broader contextual patterns present in aerial imagery.

A deeper analysis of individual metrics reveals a distinct characteristic of this classification task: the propensity for high recall. In the pre-trained condition, all five models achieved perfect Recall (100.00\%), indicating that none of the models failed to detect the presence of sky when it actually existed in the test images. Consequently, the differentiating factor in overall model performance is Precision, the ability to correctly reject negatives (absence of sky). The pre-trained ResNet50 architecture demonstrated the strongest capability in this regard, achieving the highest Precision (85.42\%) and, consequently, the highest overall Accuracy (92.55\%) and F$_1$ score (92.13\%). This suggests that while all pre-trained models were excellent at identifying positive instances, the deeper CNN was most effective at minimizing false positives.

Comparing architectures, the traditional CNN-based ResNet50 outperformed both the lightweight CNNs and the transformer-based models (Swin-T and ViT) in the pre-trained setting. It is notable that, when training from scratch, the performance hierarchy shifts slightly, with EfficientNet-B0 achieving the highest accuracy (85.11\%) among the non-pre-trained group. Furthermore, the transformer-based architectures, ViT and Swin-T, exhibited significant drops in performance without pre-training. This aligns with existing literature suggesting that vision transformers often require larger datasets or pre-training inductive biases to perform competitively with CNNs on smaller, specific tasks. Overall, the results on this task suggest that the proposed dataset presents a meaningful challenge for models trained from scratch indicating the presence of diverse visual content in the included UAV images. At the same time, the performance boost observed with pretrained architectures suggest that the dataset is rich enough to benefit from high-capacity models when transfer learning is appropriately applied.

\begin{table}[h]
\centering
\caption{Baseline method performance on sky absence/presence classification on the \textit{AIFloodSense} test set. Results are reported as percentages (\%), with and without ImageNet pretraining.}
\begin{tabular}{|l||c|c|c|c|c|}
\hline
\multirow{2}{*}{\textbf{Model}} & \multirow{2}{*}{\begin{tabular}[c]{@{}c@{}}\textbf{Pre-}\\\textbf{trained}\end{tabular}} & \multirow{2}{*}{\textbf{Acc.}} & \multirow{2}{*}{\textbf{Pr.}} & \multirow{2}{*}{\textbf{Rec.}} & \multirow{2}{*}{\textbf{F$_1$}} \\ 
 & & & & & \\
\hline
\hline
EfficientNet-B0 & \checkmark & 88.30 & 78.85 & \textbf{100.00} & 88.17 \\ \hline
ResNet18        & \checkmark & 87.23 & 77.36 & \textbf{100.00} & 87.23 \\ \hline
ResNet50        & \checkmark & \textbf{92.55} & \textbf{85.42} & \textbf{100.00} & \textbf{92.13} \\ \hline
Swin-T          & \checkmark & 90.43 & 82.00 & \textbf{100.00} & 90.11 \\ \hline
ViT             & \checkmark & 88.30 & 78.85 & \textbf{100.00} & 88.17 \\ \hline
\hline
EfficientNet-B0 & \ding{55}  & 85.11 & 75.47 &  97.56 & 85.11 \\ \hline
ResNet18        & \ding{55}  & 81.91 & 70.69 & \textbf{100.00} & 82.83 \\ \hline
ResNet50        & \ding{55}  & 82.98 & 71.93 & \textbf{100.00} & 83.67 \\ \hline
Swin-T          & \ding{55}  & 82.98 & 73.58 &  95.12 & 82.98 \\ \hline
ViT             & \ding{55}  & 81.91 & 70.69 & \textbf{100.00} & 82.83 \\ \hline
\end{tabular}
\label{tab:skynoskyclf}
\end{table}

\begin{figure}[t!]
    \centering
    \subfigure[Sorted Normalized Entropy, Confidence, and Error Locations.]{    \includegraphics[width=0.49\textwidth]{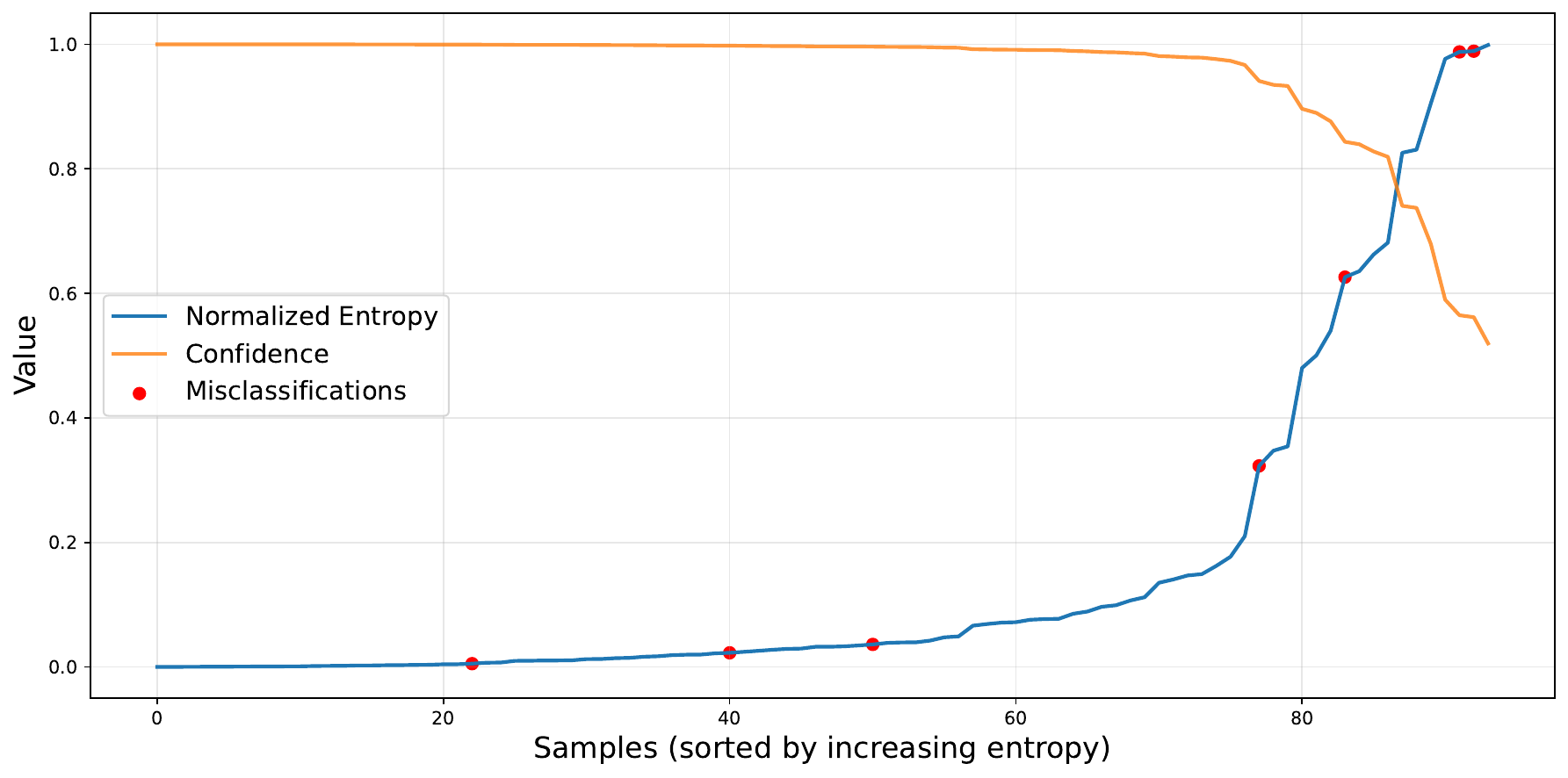}}
    \subfigure[Normalized entropy histogram.]{\includegraphics[width=0.49\textwidth]{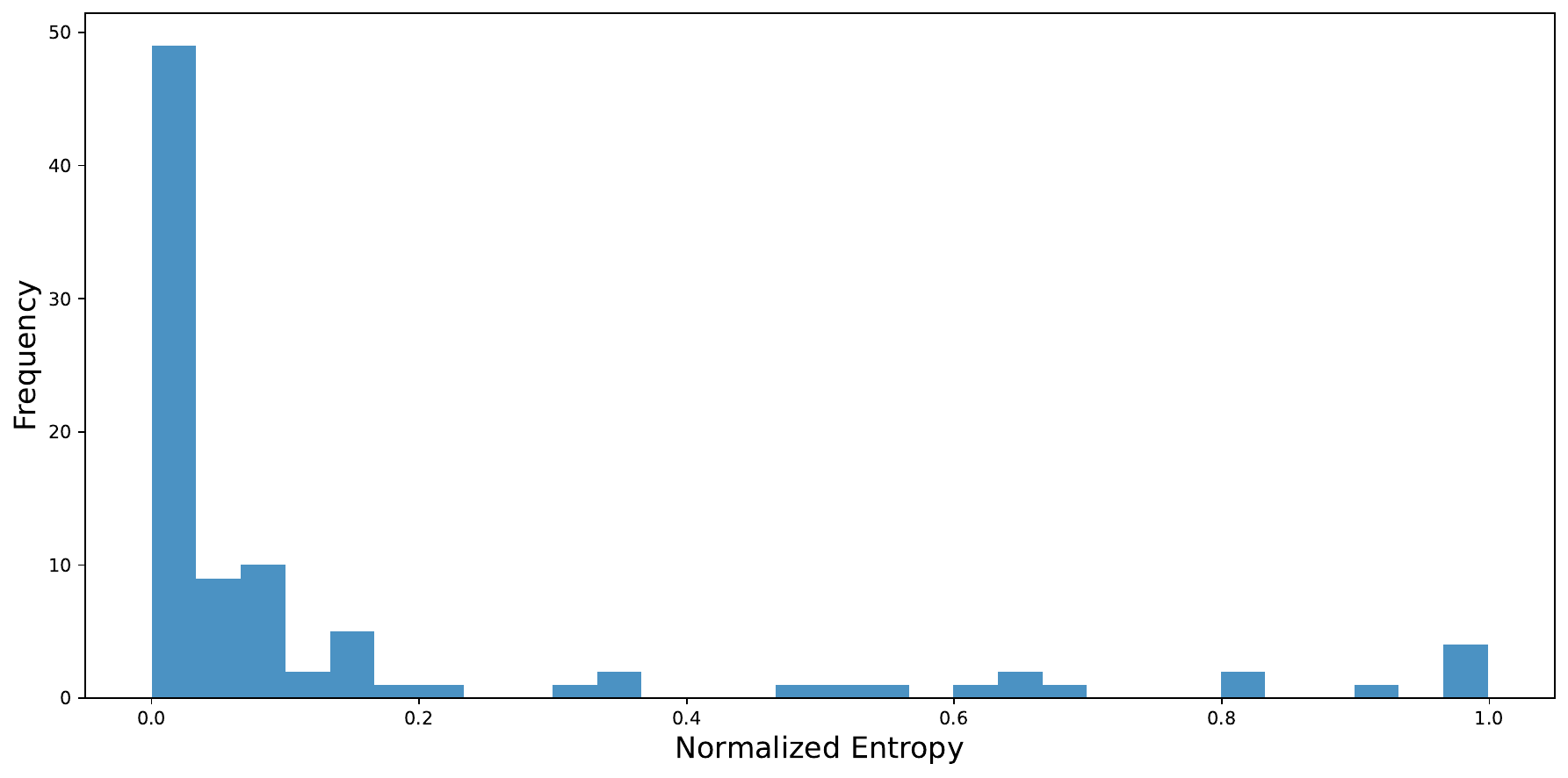}}\\
    \caption{Normalized entropy against prediction confidence with misclassifications and normalized entropy histogram for the \textit{AIFloodSense} test set of the best performing sky absence vs sky presence classifier, ResNet50 pre-trained.}
    \label{fig:2_entr_conf_hist}
\end{figure}

For the sky detection task, we further assess model calibration by examining the distribution of predictive uncertainty of the best performing model, ResNet50 pre-trained. Figure~\ref{fig:2_entr_conf_hist} (a) reveals a highly distinct confidence profile compared to the other tasks. The sorted sample curve (blue) shows that the model maintains near-zero normalized entropy and near-perfect confidence for approximately 75\% of the test set, reflecting the high accuracy metrics reported earlier in Table~\ref{tab:skynoskyclf}. The relationship between confidence (orange) and normalized entropy (blue) follows the expected inverse correlation. However, the distribution of misclassifications (red markers) exposes a critical behavior: errors are not confined solely to the high-uncertainty regime. While several misclassifications occur at the tail end where entropy is high (> 0.6), there is a notable presence of misclassifications occurring in the low-entropy region where the model expressed high confidence. The accompanying histogram (b) corroborates the model's generally decisive nature, with a massive concentration of samples in the lowest entropy bins and a sparse distribution thereafter. This suggests that while the model is highly effective at extracting sky features for the vast majority of images, the rare failure cases are often characterized by overconfidence rather than ambiguity.

\begin{figure*}[]
    \centering
    
    \subfigure[$H_{norm} = 0.0001$,\newline Pred = SP, GT = SP] {\includegraphics[width=0.24\textwidth]{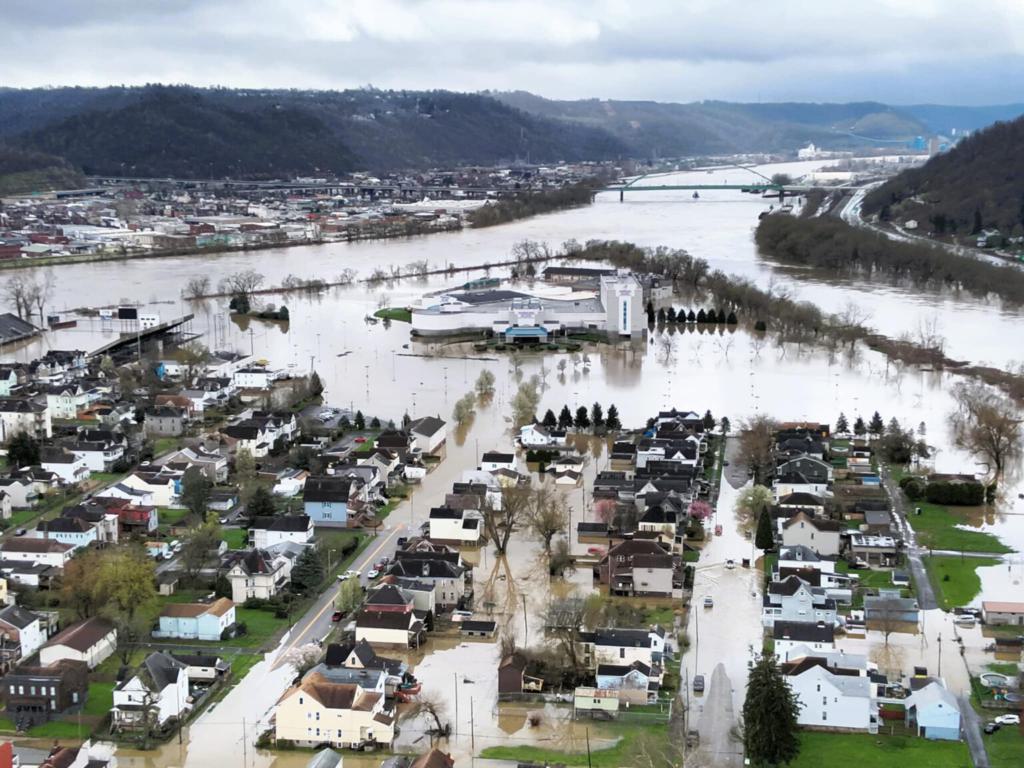}}
    \subfigure[$H_{norm} = 0.0002$,\newline Pred = SP, GT = SP] {\includegraphics[width=0.24\textwidth]{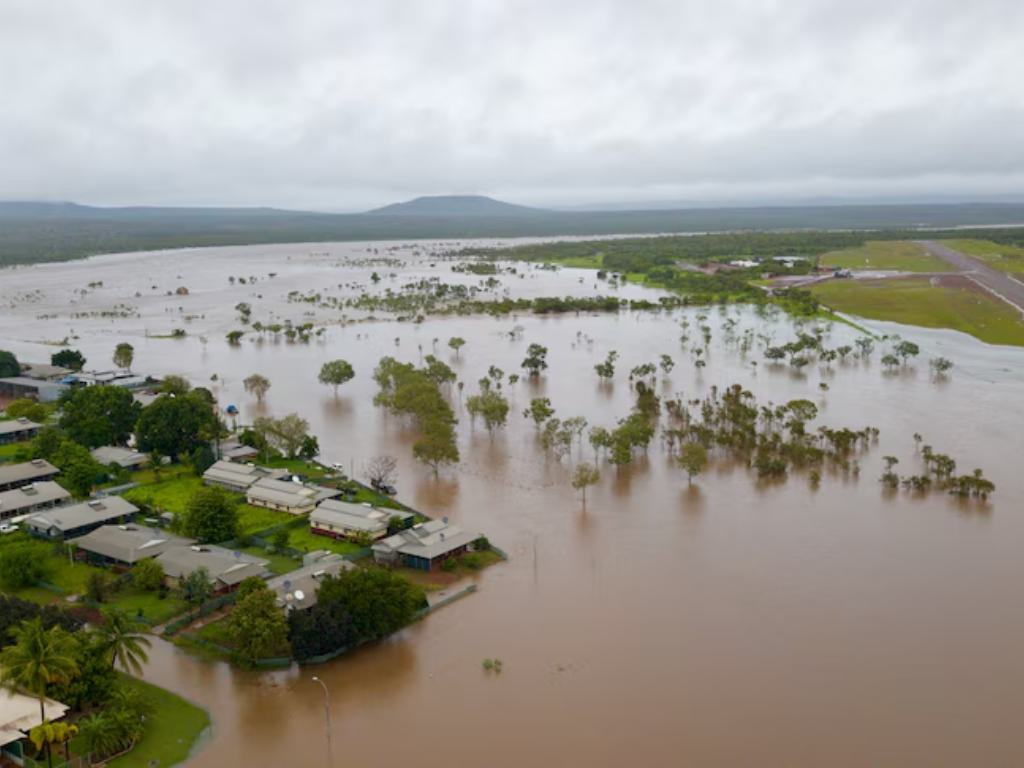}}
    \subfigure[$H_{norm} = 0.0003$,\newline Pred = SP, GT = SP] {\includegraphics[width=0.24\textwidth]{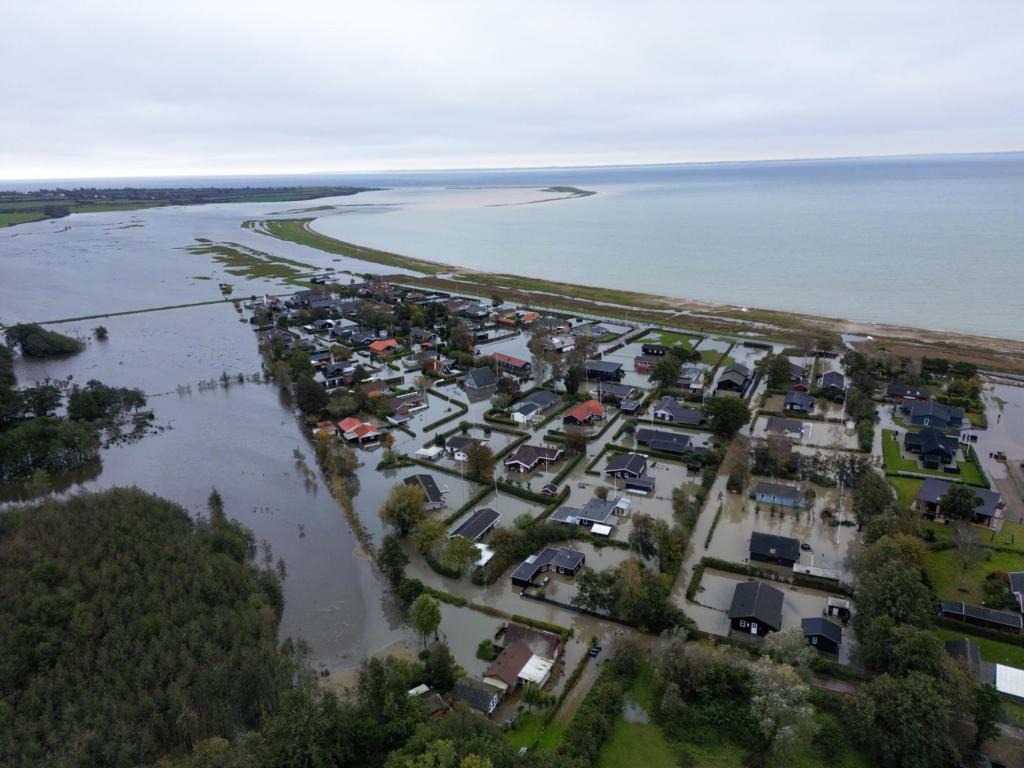}}
    \subfigure[$H_{norm} = 0.0004$,\newline Pred = SP, GT = SP] {\includegraphics[width=0.24\textwidth]{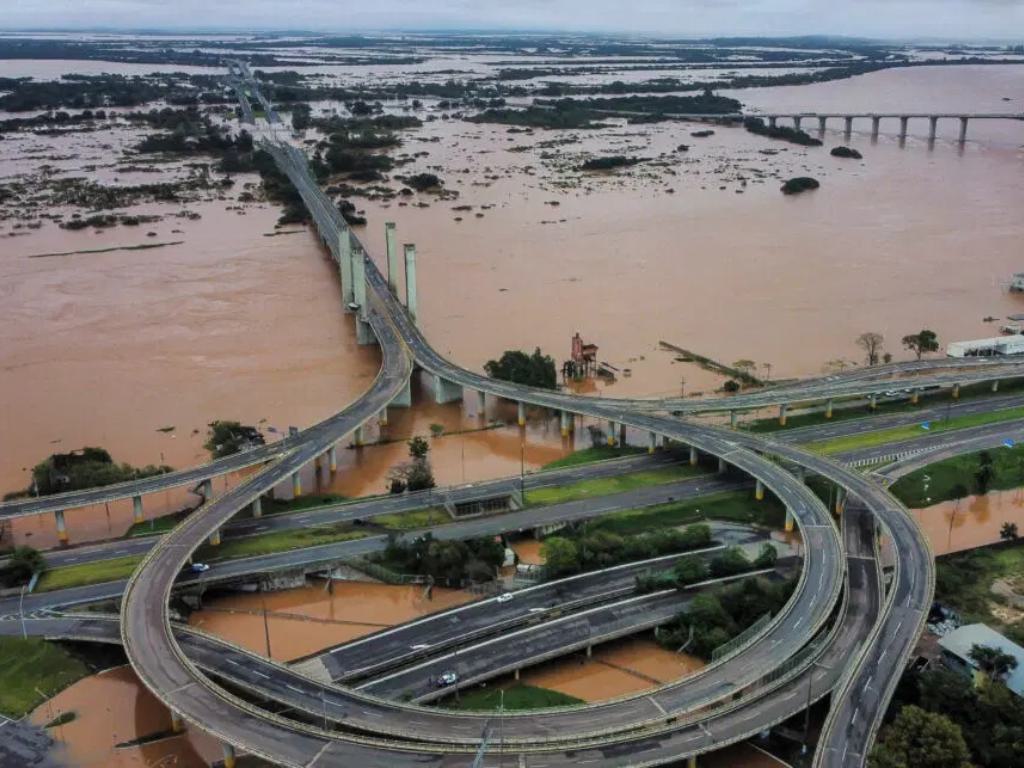}}
    \\

    \subfigure[$H_{norm} = 0.9766$,\newline Pred = SA, GT = SA] {\includegraphics[width=0.24\textwidth]{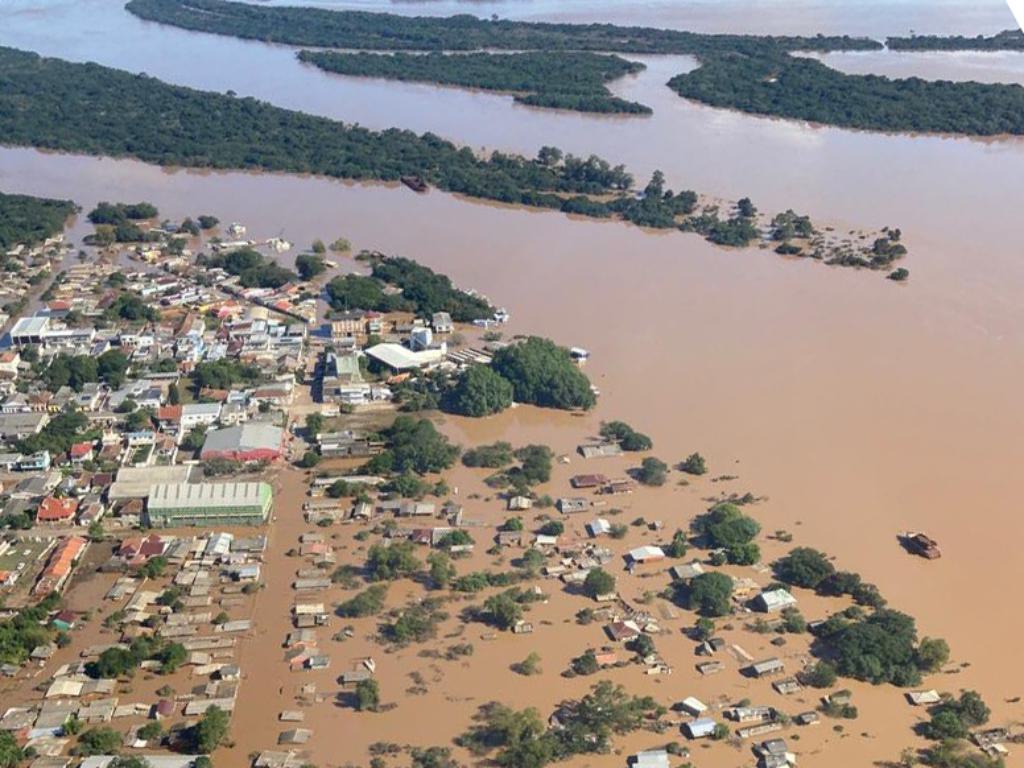}}
    \subfigure[$H_{norm} = 0.9878$,\newline Pred = SP, GT = SA] {\includegraphics[width=0.24\textwidth]{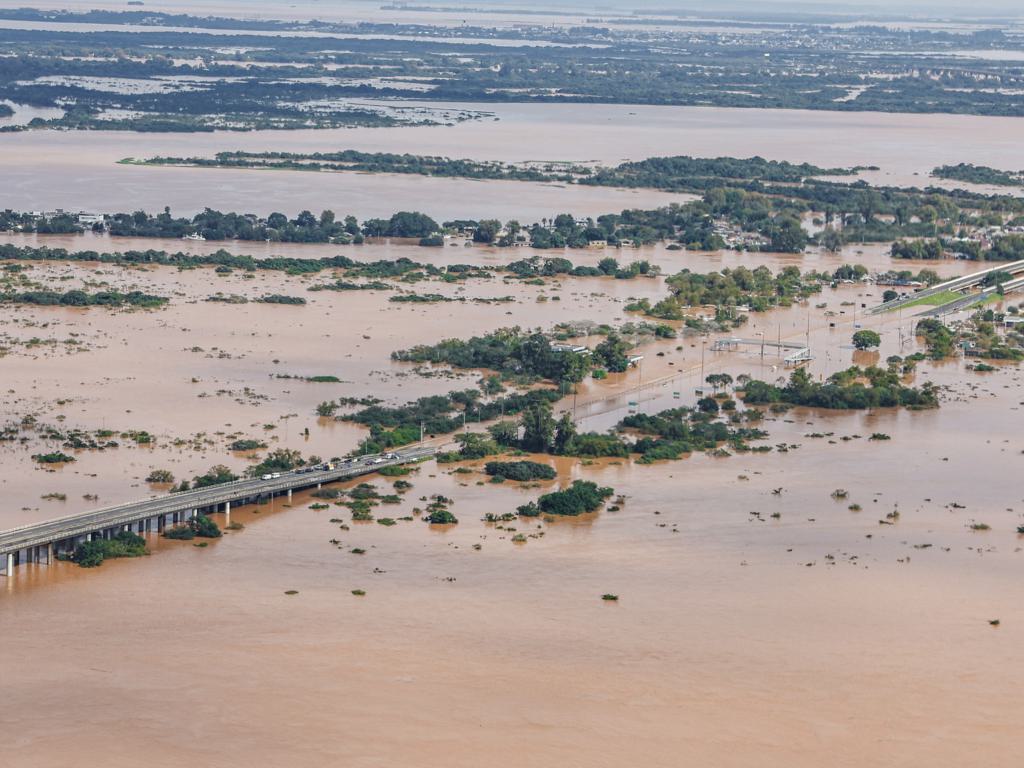}}
    \subfigure[$H_{norm} = 0.9890$,\newline Pred = SP, GT = SA] {\includegraphics[width=0.24\textwidth]{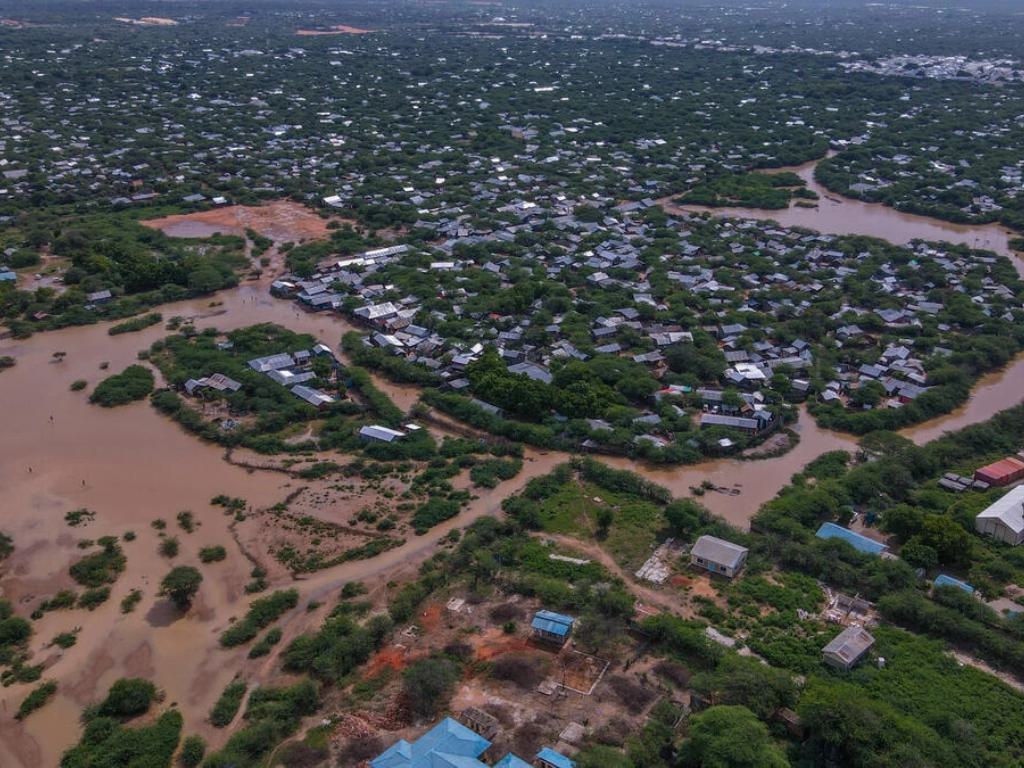}}
    \subfigure[$H_{norm} = 0.9989$,\newline Pred = SA, GT = SA] {\includegraphics[width=0.24\textwidth]{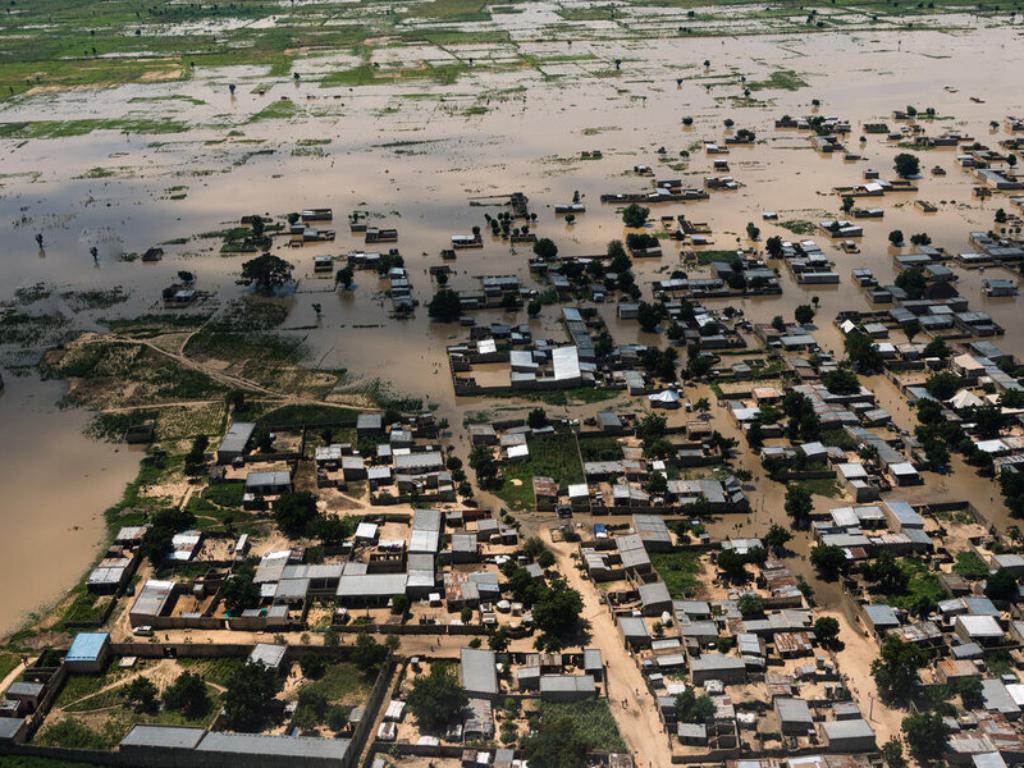}}
    \\
    
    \caption{Top 4 (first row) and bottom 4 (second row) samples with the lowest/highest normalized entropy ($H_{norm}$) for the \textit{AIFloodSense} test set of the best performing sky absence (SA) vs sky presence (SP) classifier, ResNet50 pre-trained. Pred = prediction, GT = ground truth.}
    \label{fig:2_top_bottom}
\end{figure*}

To visualize the extremes of model certainty, Figure~\ref{fig:2_top_bottom} depicts in the first row the four most confident predictions (lowest entropy), which typically feature unobstructed views or clear textures (a)--(d), alongside the four most uncertain predictions (highest entropy) in the second row. The latter group highlights complex scenarios, such as flood presence in the upper part of the image, which can be confused as sky (f), or results in confidence drop although the classification was correct (e), (h). Also, ambiguous lighting and coarse granularity of image details, due to a high-altitude capture and viewpoint, impede the extraction of features necessary for accurate classification (g).

    
\noindent \textbf{Continent Classification:} regarding the novel task of classifying the continent in which a UAV flood image was captured, the results in Table~\ref{tab:continentclf} reveal several important trends. First, pretraining plays a critical role in this task. All models, regardless of architecture, show significantly higher performance when initialized with pretrained weights. Transformer-based models, in particular, benefit the most with gains in F$_1$ score of 25.93\% and 27.83\% for ViT and Swin-T respectively. The convolutional architectures also exhibit substantial gains ranging from 15.7\% to 18.28\%. This contrast underscores the difficulty of the continent classification task given the available training data, highlighting that the inductive biases and feature representations learned from large-scale pre-training datasets are essential for achieving tractable performance in this domain.

Within the pre-trained paradigm, transformer-based architectures demonstrated superior capability compared to their convolutional counterparts. The Vision Transformer (ViT) achieved the highest overall Accuracy (54.26\%) and Macro F$_1$ score (50.29\%), closely followed by the Swin Transformer (Swin-T). CNN-based methods generally lagged behind, with ResNet18 performing best among them with an accuracy of 48.94\%. Notably, the deeper ResNet50 underperformed the shallower ResNet18 across all pre-trained metrics. This counterintuitive result suggests that for this specific task complexity and dataset size, the deeper architecture may have suffered from overfitting, or that the additional capacity of ResNet50 was not necessary to capture the requisite discriminatory features for continent identification.

Conversely, when training from scratch, the performance hierarchy shifts slightly, though overall efficacy is low. While ViT and ResNet50 achieved the joint-highest accuracy (36.17\%) in this setting, an examination of Macro F$_1$ scores reveals that ResNet18 maintained the most robust relative performance (28.44\%). The transformer models, known for lacking the inherent inductive biases of CNNs (such as translation invariance/equivariance), exhibited severe performance degradation without pre-training data to guide feature learning. Furthermore, the large discrepancies between Accuracy and Macro Precision observed in the trained from scratch results (e.g., EfficientNet-B0: 34.04\% Acc. vs. 20.31\% Macro Pr.) indicate that models trained without pre-training likely struggled significantly with class imbalance, achieving moderate accuracy by favoring majority classes while failing to correctly classify minority continents.

\begin{table}[t!]
\centering
\caption{Evaluation of the performance of the baseline methods on the task of continent classification on the \textit{AIFloodSense} test set. Results are reported as percentages (\%), with and without ImageNet pretraining. Macro F$_1$ scores were computed as the unweighted mean of class-wise F$_1$ scores.}
\begin{tabular}{|l||c|c|c|c|c|}
\hline
\multirow{2}{*}{\textbf{Model}} & 
\multirow{2}{*}{\begin{tabular}[c]{@{}c@{}}\textbf{Pre-}\\\textbf{trained}\end{tabular}} & 
\multirow{2}{*}{\textbf{Acc.}} & 
\multicolumn{3}{c|}{\textbf{Macro average}} \\ 
\cline{4-6}
 & & & \textbf{Pr.} & \textbf{Rec.} & \textbf{F$_1$} \\
\hline
\hline
EfficientNet-B0 & \checkmark & 44.68 & \textbf{54.41} & 43.16 & 38.73 \\ \hline
ResNet18        & \checkmark & 48.94 & 49.52 & 48.11 & 46.72 \\ \hline
ResNet50        & \checkmark & 43.62 & 42.79 & 44.51 & 41.13 \\ \hline
Swin-T          & \checkmark & 53.19 & 53.16 & \textbf{53.48} & 50.28 \\ \hline
ViT             & \checkmark & \textbf{54.26} & 51.90 & 53.40 & \textbf{50.29} \\ \hline
\hline
EfficientNet-B0 & \ding{55}  & 34.04 & 20.31 & 26.47 & 20.63 \\ \hline
ResNet18        & \ding{55}  & 31.91 & 34.58 & 31.05 & 28.44 \\ \hline
ResNet50        & \ding{55}  & 36.17 & 33.69 & 28.89 & 25.43 \\ \hline
Swin-T          & \ding{55}  & 34.04 & 20.60 & 28.73 & 22.45 \\ \hline
ViT             & \ding{55}  & 36.17 & 21.12 & 30.28 & 24.36 \\ \hline
\end{tabular}
\label{tab:continentclf}
\end{table}

\begin{figure}[t!]
    \centering
    \subfigure[ViT normalized confusion matrix.]{    \includegraphics[width=0.48\textwidth]{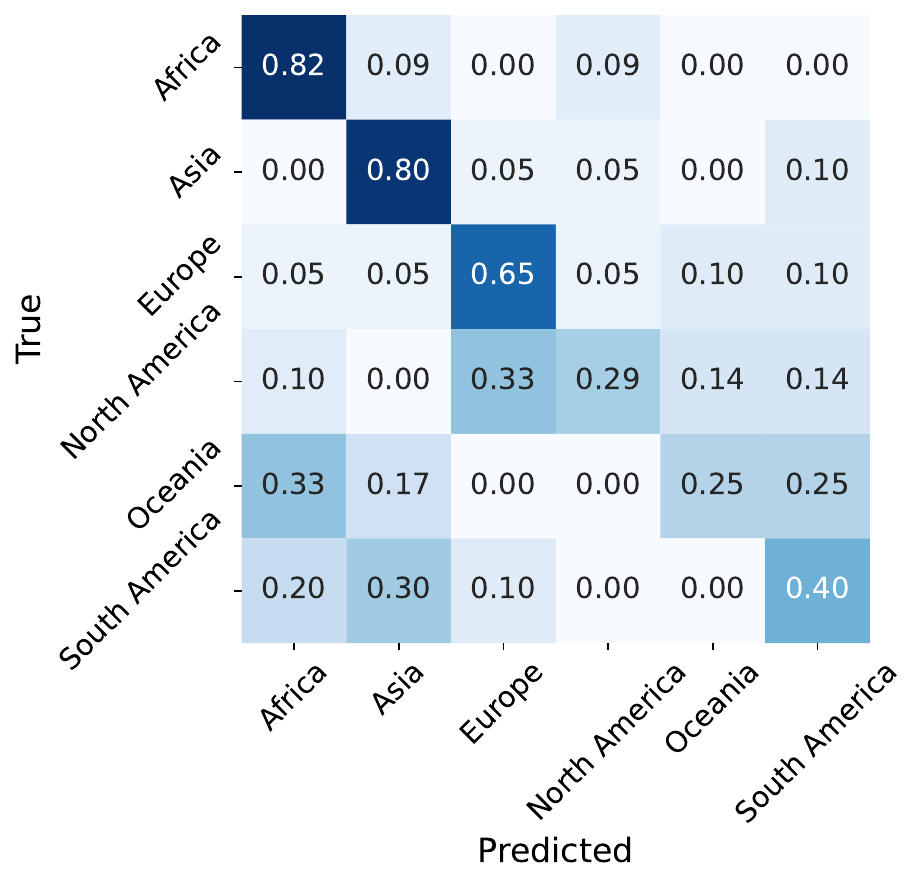}}
    \subfigure[ResNet18 normalized confusion matrix.]{    \includegraphics[width=0.48\textwidth]{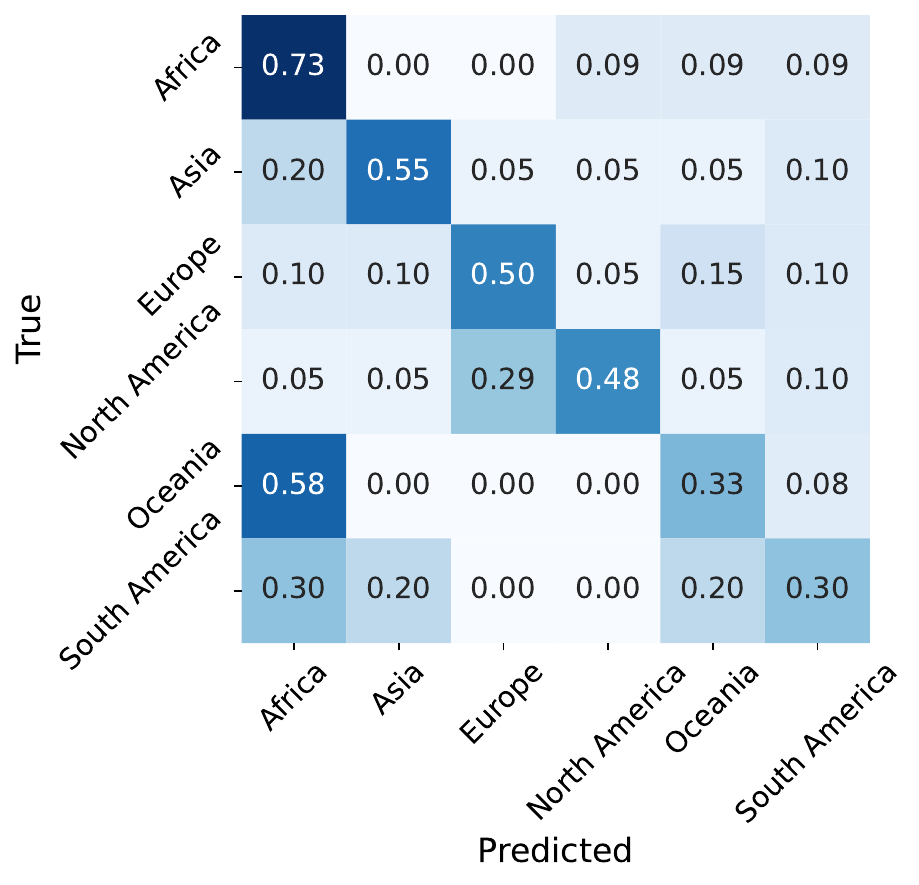}}
    \caption{Normalized confusion matrices of the best performing models, ViT and ResNet18, both pre-trained, for the continents classification task.}
    \label{fig:cmresn18vit}
\end{figure}

Furthermore, in Figure~\ref{fig:cmresn18vit} we present the normalized confusion matrices for the best performing transformer and CNN architectures, namely ViT and ResNet18, both pre-trained, regarding the task of continent classification. An examination of these reveals distinct error patterns between transformer-based and convolutional architectures. While the ViT model achieved superior detection rates for distinct classes such as Africa (82\%) and Asia (80\%), it exhibited significant confusion between North America and Europe, misclassifying 33\% of North American samples as European, potentially due to similarities in built environments or temperate vegetation. In contrast, ResNet18 demonstrated a different failure profile, struggling to distinguish South America from Africa, with 30\% of South American samples mislabeled as African. A particularly sharp divergence occurred for the Oceania class, where the ViT performance collapsed to 25\% accuracy, confusing it with Africa (33\%), South America (25\%), and Asia (17\%). Conversely, ResNet18 exhibited a severe and concentrated bias, misclassifying 58\% of Oceanian samples as African, suggesting that the CNN architecture struggles profoundly to differentiate the visual features of these two regions. Notably, this specific tendency to misclassify Oceania as Africa appears to be a systemic limitation across all evaluated CNN architectures in their pre-trained versions.

\begin{figure}[t!]
    \centering  
    \subfigure[Sorted Normalized Entropy, Confidence, and Error Locations.]{    \includegraphics[width=0.49\textwidth]{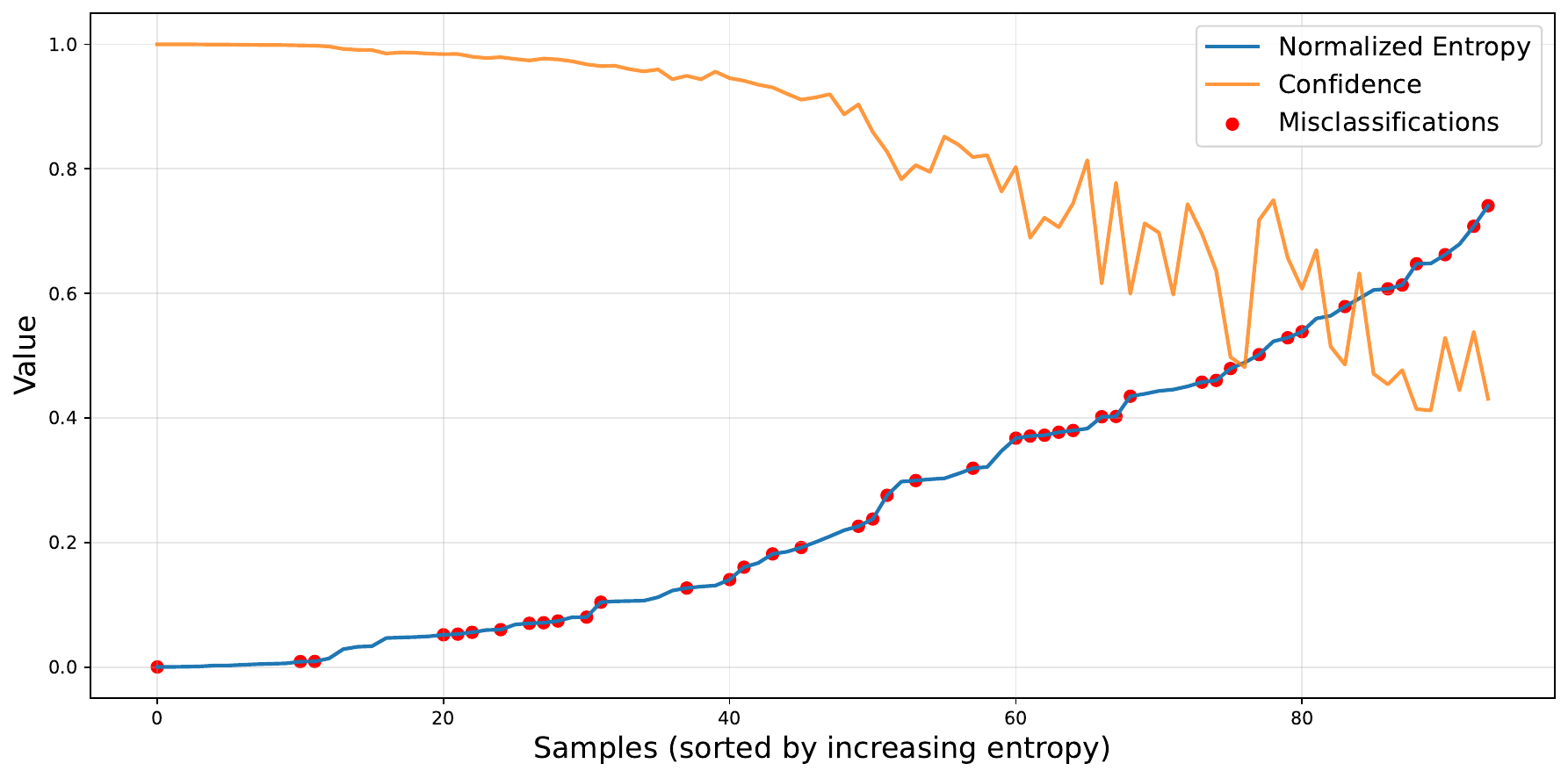}}
    \subfigure[Normalized entropy histogram.]{\includegraphics[width=0.49\textwidth]{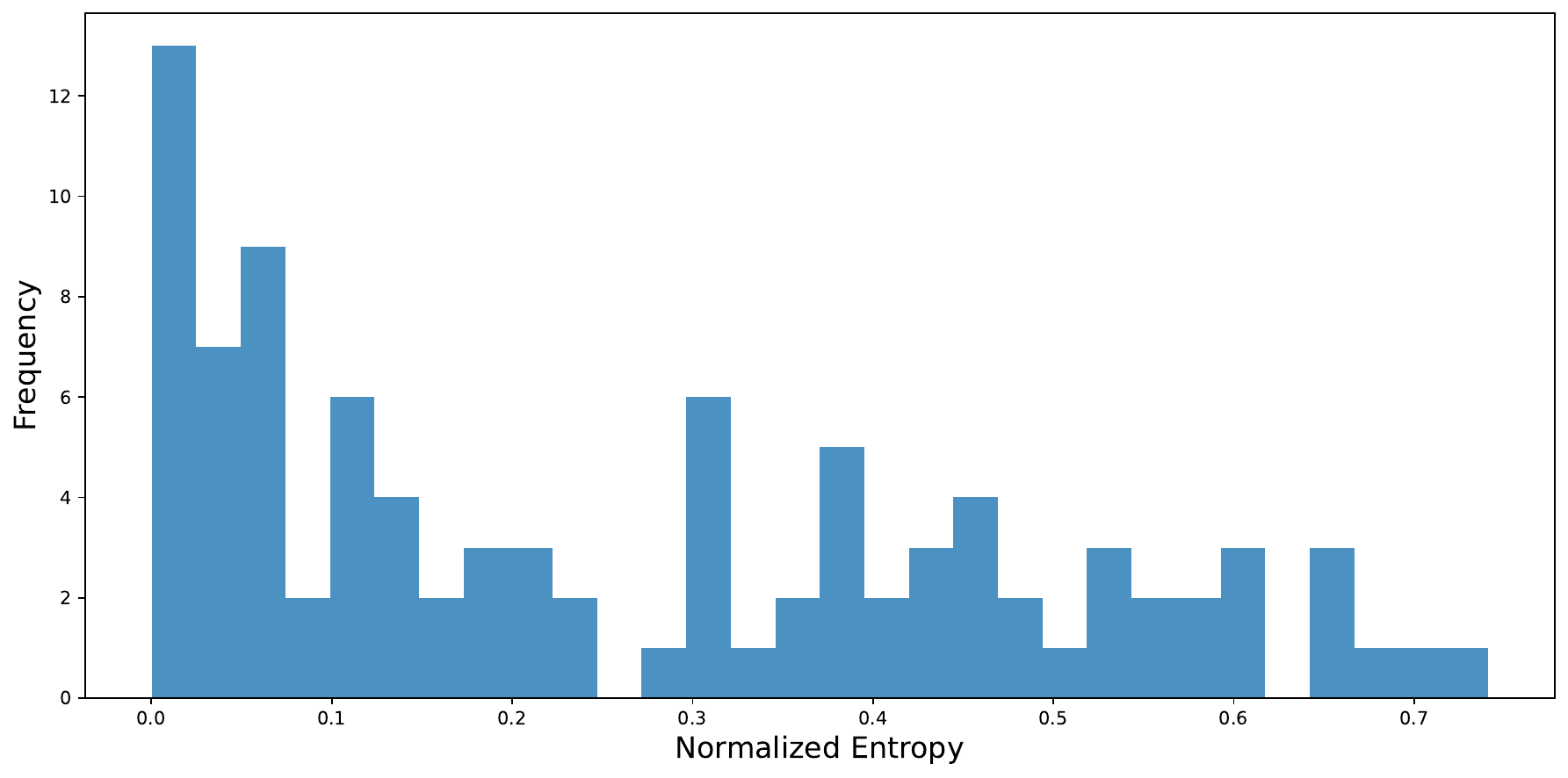}}
    \caption{Normalized entropy against prediction confidence with misclassifications and normalized entropy histogram for the \textit{AIFloodSense} test set of the best performing continent classifier, ViT pre-trained.}
    \label{fig:3_entr_conf_hist}
\end{figure}

Finally, we characterize the uncertainty profile of the multi-class continent classification task. Figure~\ref{fig:3_entr_conf_hist} (a) illustrates the sorted normalized entropy and confidence values, with misclassifications marked in red to visualize the alignment between model uncertainty and error rates in this more complex semantic space. In contrast to the previous binary classification tasks, the continent classification results in a significantly more complex uncertainty landscape, characteristic of a difficult multi-class problem with high inter-class similarity. The sorted normalized entropy curve (blue) demonstrates a weaker calibration between confidence (orange) and accuracy. Unlike the previous tasks, where errors were sequestered in the high-uncertainty tail, misclassifications (red markers) here are ubiquitously distributed across the curve. Notably, a distinct density of errors appears in the low-entropy region, indicating that the model frequently produces confident failures, assigning a high probability to incorrect classes. This suggests that shared visual features between continents (e.g., similar urban architectures in Europe and North America) often trigger strong but erroneous activations. The histogram of normalized entropy in Figure~\ref{fig:3_entr_conf_hist} (b) corroborates this struggle. Rather than a sharp peak at zero, it exhibits a broad dispersion with significant mass extending into the moderate-to-high entropy ranges. This confirms that the model operates under a regime of systemic uncertainty for a large proportion of the test set, struggling to isolate salient features for specific geographies.

\begin{figure*}[]
    \centering
    \subfigure[$H_{norm} = 0.0002$,\newline Pred = SA,GT = AS] {\includegraphics[width=0.24\textwidth]{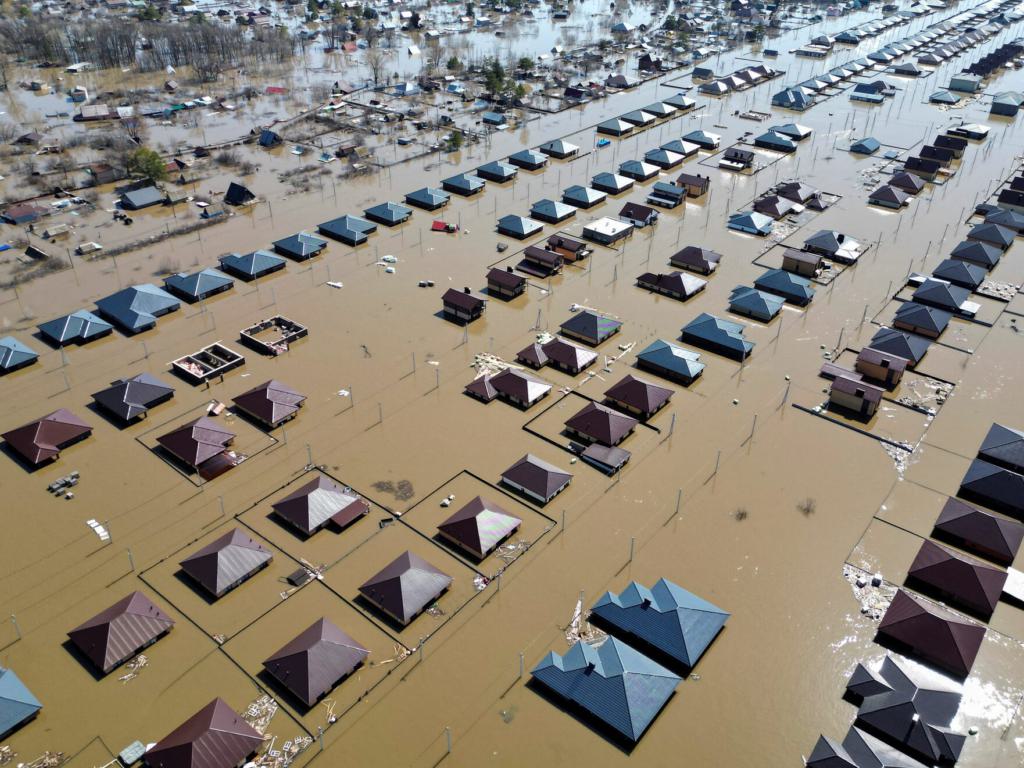}}
    \subfigure[$H_{norm} = 0.0003$,\newline Pred = AS,GT = AS] {\includegraphics[width=0.24\textwidth]{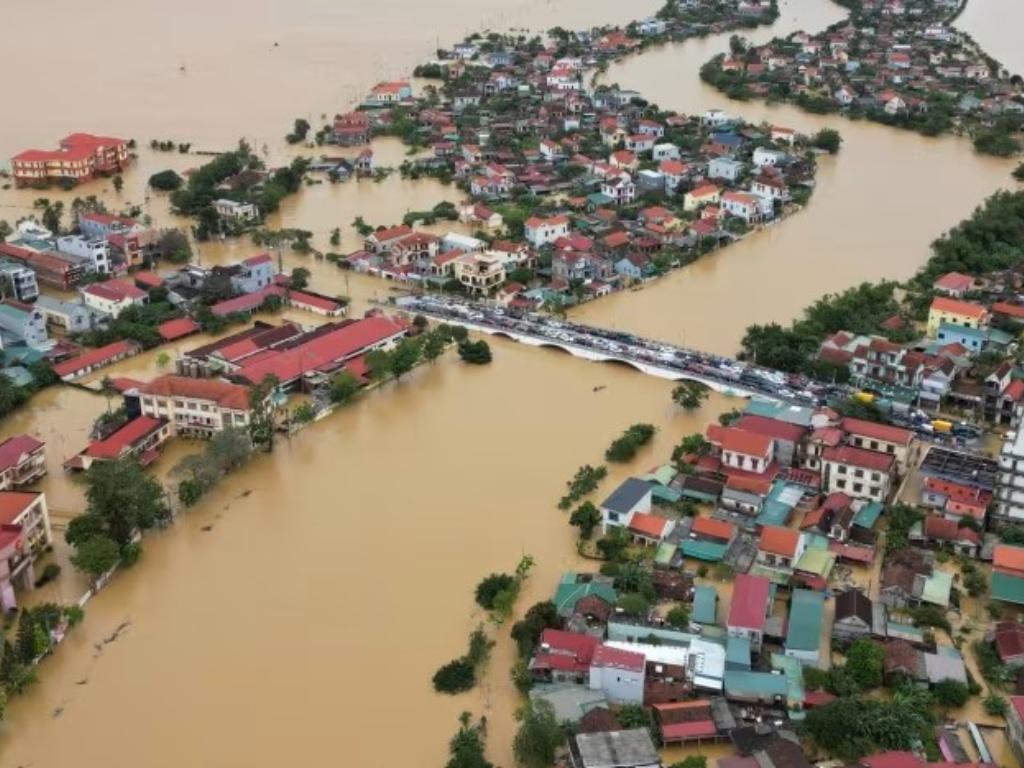}}
    \subfigure[$H_{norm} = 0.0006$,\newline Pred = AS,GT = AS] {\includegraphics[width=0.24\textwidth]{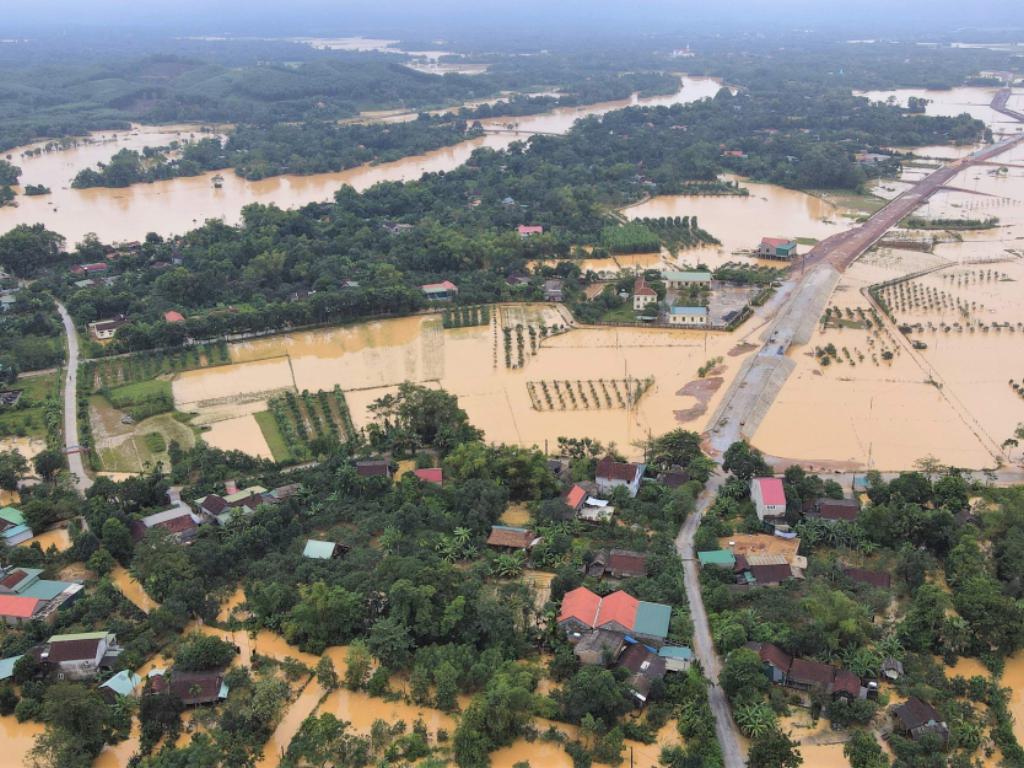}}
    \subfigure[$H_{norm} = 0.0011$,\newline Pred = AS,GT = AS] {\includegraphics[width=0.24\textwidth]{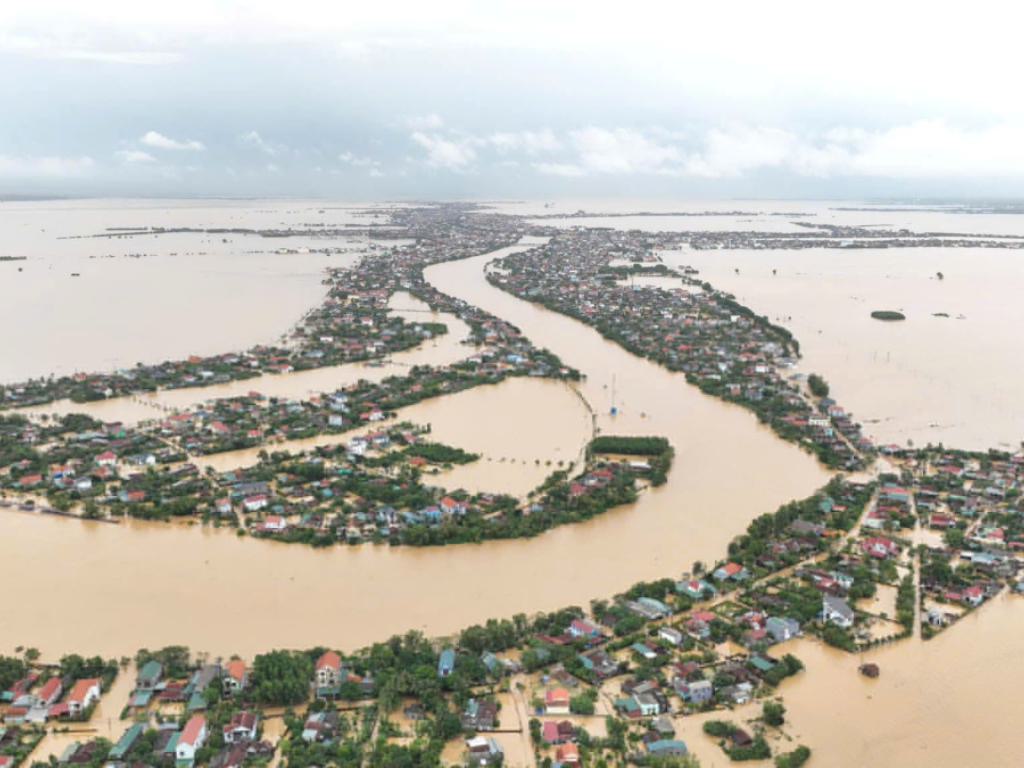}}
      \subfigure[$H_{norm} = 0.6622$,\newline Pred = AF,GT = NA] {\includegraphics[width=0.24\textwidth]{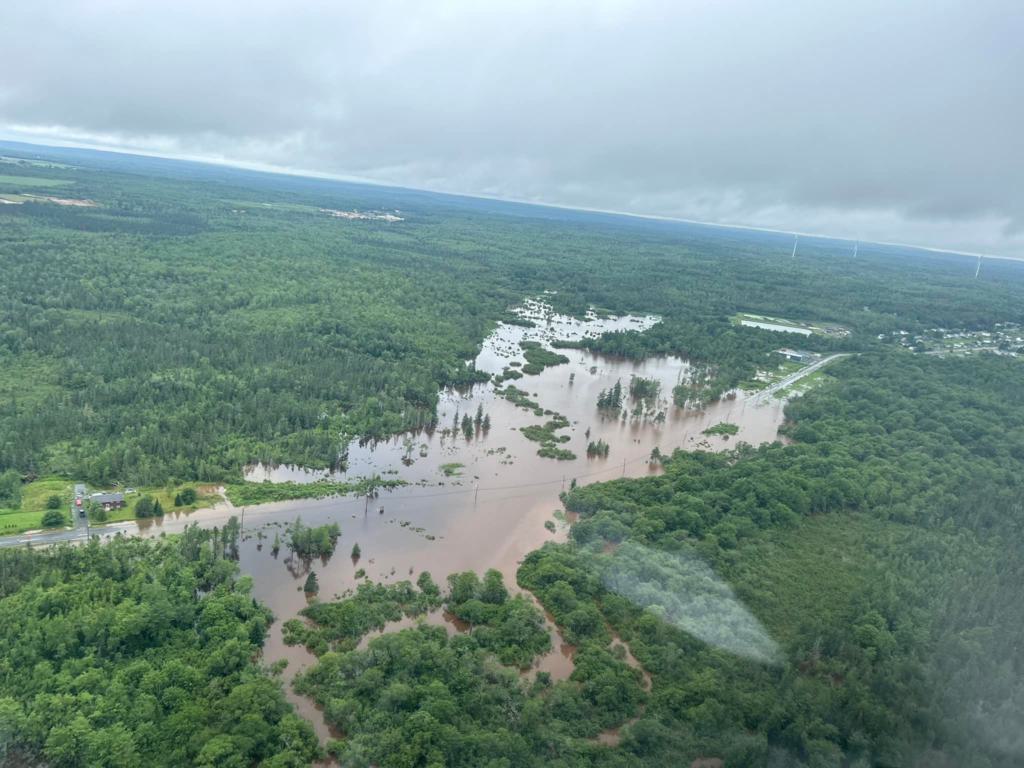}}
    \subfigure[$H_{norm} = 0.6790$,\newline Pred = EU,GT = EU] {\includegraphics[width=0.24\textwidth]{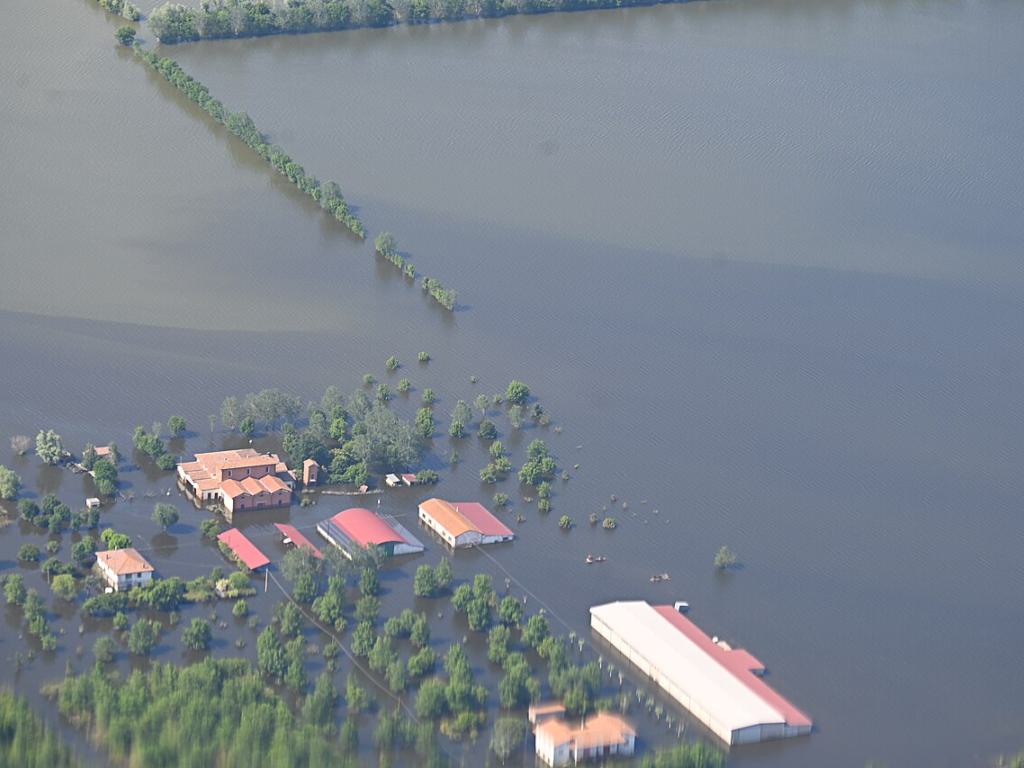}}
    \subfigure[$H_{norm} = 0.7078$,\newline Pred = OC,GT = EU] {\includegraphics[width=0.24\textwidth]{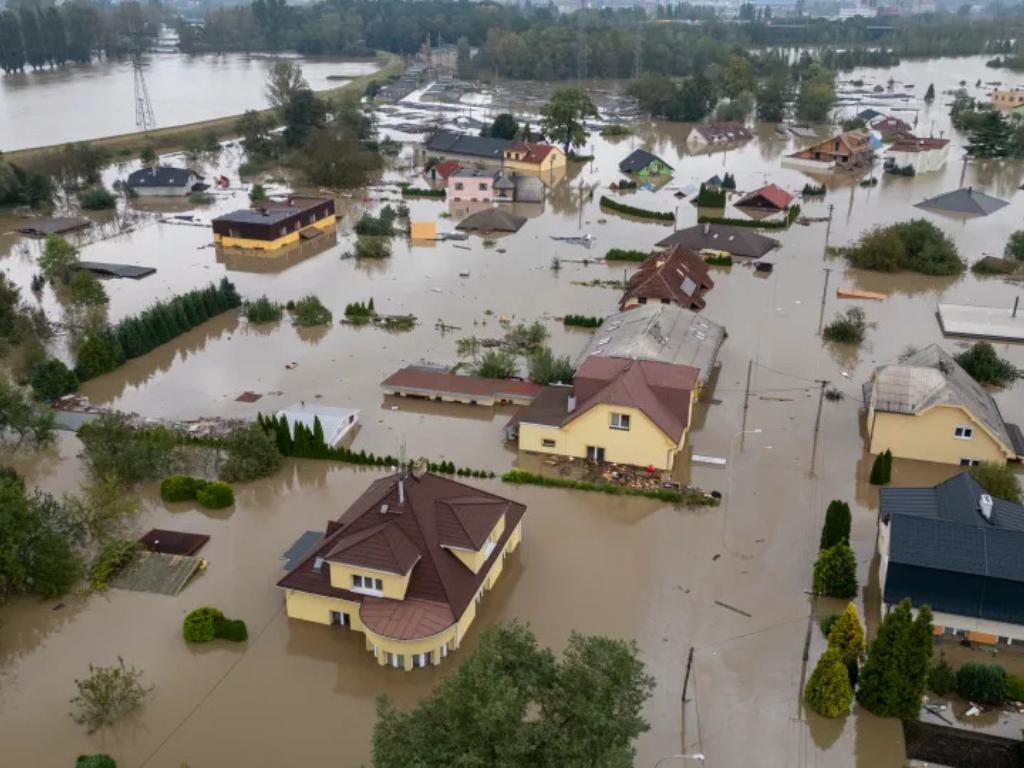}}
    \subfigure[$H_{norm} = 0.7408$,\newline Pred = OC,GT = NA] {\includegraphics[width=0.24\textwidth]{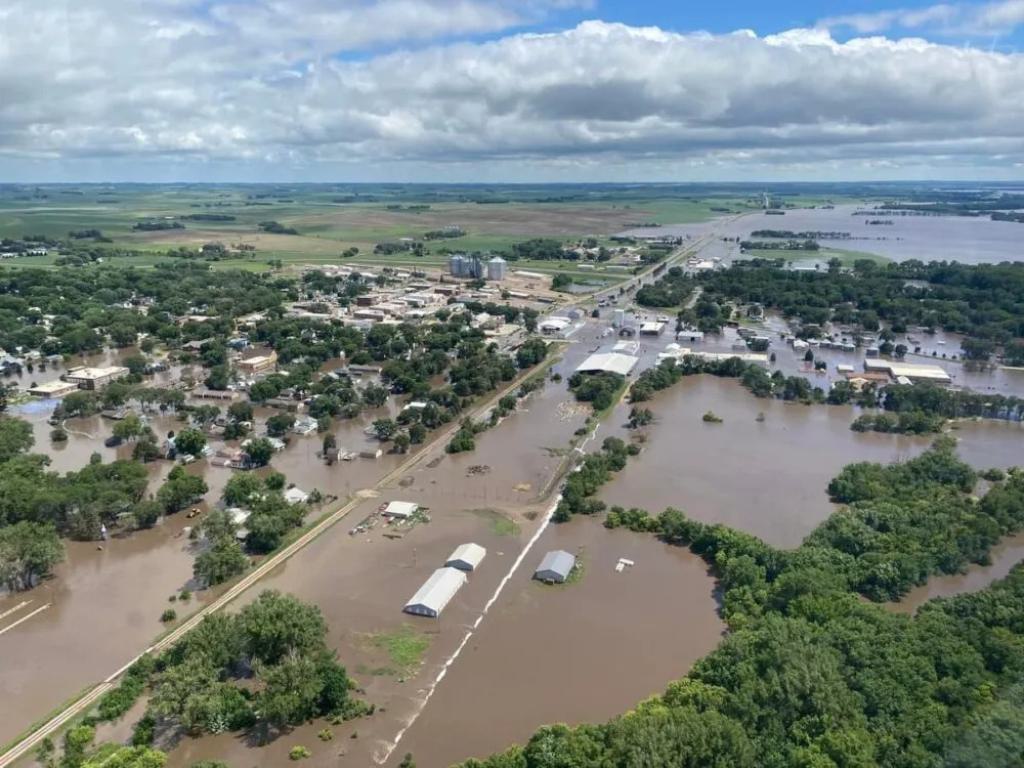}}
    \caption{Top 4 (first row) and bottom 4 (second row) samples with the lowest/highest normalized entropy ($H_{norm}$) for the \textit{AIFloodSense} test set of the best performing continent classifier, ViT pre-trained. AF = Africa, AS = Asia, EU = Europe, NA = North America, OC = Oceania, SA = South America, Pred = prediction, GT = ground truth.}
    \label{fig:3_top_bottom}
\end{figure*}

Complementing these metrics, Figure~\ref{fig:3_top_bottom} provides a qualitative assessment of the model's decision-making extremes. The top row displays the four samples with the lowest normalized entropy, representing the most confident and distinct geographical features learned by the model. Interestingly, the most confident prediction is a misclassification confusing Asia with South America likely due to similar infrastructures in the image (a). In contrast, the bottom row presents the four samples with the highest normalized entropy. These most uncertain predictions offer insight into the hard samples which characterize this task. Generic landscapes (e), (h) or similar architectural features (g) lack the discriminative visual cues required for confident continent identification. Overall, the results demonstrate the challenging nature of the continent classification task, as even the best-performing model remains well below the ceiling performance. This indicates that the visual variability across continents is not trivially captured and highlights the diversity embedded in the proposed dataset.

    
\subsection{Semantic Segmentation Task}
In this subsection, we focus on the semantic segmentation of UAV images, provided in our novel \textit{AIFloodSense} dataset,  annotated with four classes: flooded regions, buildings, sky, and background. As mentioned in Section~\ref{ssubsec:sstask} these classes are particularly important as they allow accurate assessment of flood impact on urban and rural areas, identification of affected infrastructure, and support for disaster response planning. The segmentation task can be structured in multiple ways: as a \textbf{multiclass problem}, where all four classes are predicted simultaneously, or as a \textbf{binary problem}, where a dedicated model is trained for each category. While multiclass segmentation provides a unified model, the binary approach allows each model to specialize on its target class, potentially improving per-class performance. 

We report per-class macro-averaged metrics (Precision, Recall, $F_1$) along with pixel-wise Accuracy and mean Intersection-over-Union (mIoU) for the multiclass segmentation task. Micro-averaged metrics are omitted, as they are dominated by majority classes and provide limited insight for evaluating segmentation performance. Unless otherwise noted, we use the widely adopted definition of mIoU (also called IoU macro): we first compute the Intersection-over-Union for each class across the entire test set, and then average these values over classes. This definition is consistent with common benchmarks (e.g. Cityscapes, ADE20K). For the binary segmentation task, IoU, Precision, Recall, and $F_1$ are reported for the target class only, with the background class excluded.

\noindent \textbf{Multi-class Protocol:} When framing the task as a multi-class segmentation problem, the performance of the baseline models, summarized in Table~\ref{tab:segm4classes}, leads to several key observations. First, ImageNet pretraining consistently improved performance across all architectures, in both global metrics (Accuracy, mIoU, F$_1$) and per-class IoU scores, highlighting the critical role of transfer learning. 
The absence of pretraining leads to a marked performance degradation, especially for transformer-based models. Specifically, the mean Intersection over Union (mIoU) decreases by 23.84\% for Swin-T and 19.64\% for SegFormer-B0, illustrating the strong dependence of high-capacity models on proper weight initialization.

Among convolutional architectures, U-Net achieved the highest mean Intersection-over-Union (mIoU: 77.60\%), macro F$_1$ score (86.84\%), as well as the best per-class IoUs, but slightly outperforming FCN for the Sky class (93.90\% vs 93.74\%), and Building class (59.95\% vs 59.75\%). In particular, U-Net also demonstrated superior performance when trained from scratch, surpassing the other convolutional and transformer-based models in all metrics, except for the per-class IoU for class Sky (76.83\%), where both DeepLabV3 (83.94\%) and FCN (86.71\%) excelled. Among the transformer-based models, Swin-T achieved the highest overall performance (Accuracy 87.31\%, mIoU 78.75\%, macro F$_1$ 87.72\%), exceeding both SegFormer-B0 and all convolutional architectures. Swin-T further exhibited consistently strong per-class IoU scores, particularly for challenging classes such as Flood (82.43\%) and Building (62.98\%), indicating a more balanced segmentation capability across both large and small objects.  

\begin{table*}[]
\centering
\caption{Performance comparison of baseline semantic segmentation models on the \textit{AIFloodSense} test set under the multiclass segmentation protocol. Results are reported as percentages (\%), with and without ImageNet pretraining.}
\label{tab:segm4classes}
\setlength{\tabcolsep}{3pt}
\resizebox{\linewidth}{!}{%
\begin{tabular}{|l||c||c|c|c|c|c||cccc|}
\hline
\multirow{2}{*}{\textbf{Model}} & 
\multirow{2}{*}{\begin{tabular}[c]{@{}c@{}}\textbf{Pre-}\\\textbf{trained}\end{tabular}} & 
\multirow{2}{*}{\textbf{Acc.}} & 
\multirow{2}{*}{\textbf{mIoU}} & 
\multicolumn{3}{c||}{\textbf{Macro average}} & 
\multicolumn{4}{c|}{\textbf{Per-class IoU}} \\
\cline{5-11}
 & & & & 
\textbf{Pr.} & \textbf{Rec.} & \textbf{F$_1$} & 
\textbf{Bgr} & \textbf{Flood} & \textbf{Sky} & \textbf{Build.} \\
\hline
\hline
DeepLabV3 & \checkmark & 84.69 & 75.37 & 84.87 & 85.95 & 85.40 & \multicolumn{1}{c|}{71.35} & \multicolumn{1}{c|}{78.16} & \multicolumn{1}{c|}{93.01} & \multicolumn{1}{c|}{58.96}\\
\hline
FCN & \checkmark & 85.15 & 76.10 & 85.38 & 86.39 & 85.88 & \multicolumn{1}{c|}{71.97} & \multicolumn{1}{c|}{78.94} & \multicolumn{1}{c|}{93.74} & \multicolumn{1}{c|}{59.75}\\
\hline
U-Net & \checkmark & 86.61 & 77.60 & 86.24 & 87.54 & 86.84 & \multicolumn{1}{c|}{75.23} & \multicolumn{1}{c|}{81.31} & \multicolumn{1}{c|}{93.90} & \multicolumn{1}{c|}{59.95}\\
\hline
SegFormer-B0 & \checkmark & 80.23 & 69.77 & 80.08 & 83.03 & 81.39 & \multicolumn{1}{c|}{66.30} & \multicolumn{1}{c|}{70.71} & \multicolumn{1}{c|}{90.76} & \multicolumn{1}{c|}{51.31}\\
\hline
Swin-T & \checkmark & \textbf{87.31} & \textbf{78.75} & \textbf{86.47} & \textbf{89.33} & \textbf{87.72} & \multicolumn{1}{c|}{\textbf{75.82}} & \multicolumn{1}{c|}{\textbf{82.43}} & \multicolumn{1}{c|}{\textbf{94.15}} & \multicolumn{1}{c|}{\textbf{62.98}}\\
\hline
\hline
DeepLabV3 & \ding{55} & 75.43 & 62.09 & 75.29 & 75.61 & 75.08 & \multicolumn{1}{c|}{63.02} & \multicolumn{1}{c|}{66.17} & \multicolumn{1}{c|}{83.94} & \multicolumn{1}{c|}{35.23}\\
\hline
FCN & \ding{55} & 75.25 & 62.86 & 74.79 & 77.83 & 75.55 & \multicolumn{1}{c|}{62.79} & \multicolumn{1}{c|}{66.56} & \multicolumn{1}{c|}{86.71} & \multicolumn{1}{c|}{35.38}\\
\hline
U-Net & \ding{55} & 78.48 & 64.11 & 77.75 & 78.15 & 77.17 & \multicolumn{1}{c|}{67.12} & \multicolumn{1}{c|}{71.71} & \multicolumn{1}{c|}{76.83} & \multicolumn{1}{c|}{40.77}\\
\hline
SegFormer-B0 & \ding{55} & 66.78 & 50.13 & 65.75 & 72.68 & 65.74 & \multicolumn{1}{c|}{58.79} & \multicolumn{1}{c|}{55.33} & \multicolumn{1}{c|}{57.91} & \multicolumn{1}{c|}{28.47}\\
\hline
Swin-T & \ding{55} & 71.43 & 54.91 & 67.74 & 77.46 & 70.07 & \multicolumn{1}{c|}{60.11} & \multicolumn{1}{c|}{61.59} & \multicolumn{1}{c|}{63.33} & \multicolumn{1}{c|}{34.60}\\
\hline
\end{tabular}
}
\end{table*}

In contrast, SegFormer-B0 showed the lowest performance among all evaluated models, lagging behind Swin-T and the convolutional architectures (U-Net, DeepLabV3, FCN) in both pre-trained and from-scratch settings. Its lightweight design and limited parameter capacity likely restricted its ability to capture the complex spatial patterns and multiscale contextual information necessary for accurately segmenting diverse classes such as Flood and Building. Additionally, the simpler decoder structure may have hindered effective aggregation of fine-grained details with global context, resulting in reduced per-class IoU scores and overall mIoU. These findings highlight the sensitivity of low-capacity transformer models to dataset complexity and initialization, suggesting that more robust architectures or additional pretraining may be required for reliable multiclass segmentation.

\begin{figure}[t!]
    \centering
    \subfigure[Swin-T normalized confusion matrix.]{    \includegraphics[width=0.48\textwidth]{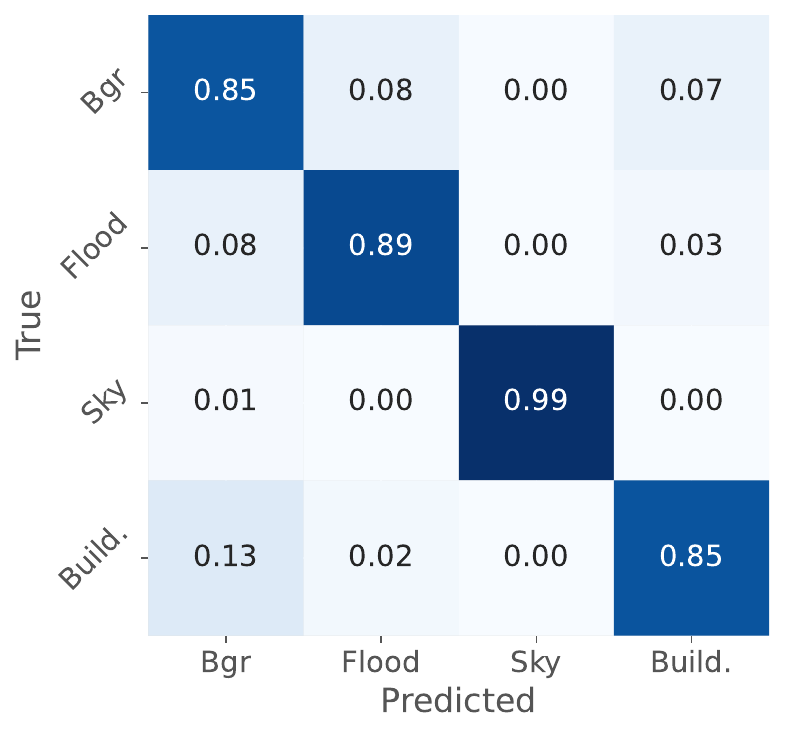}}
    \subfigure[U-Net normalized confusion matrix.]{    \includegraphics[width=0.48\textwidth]{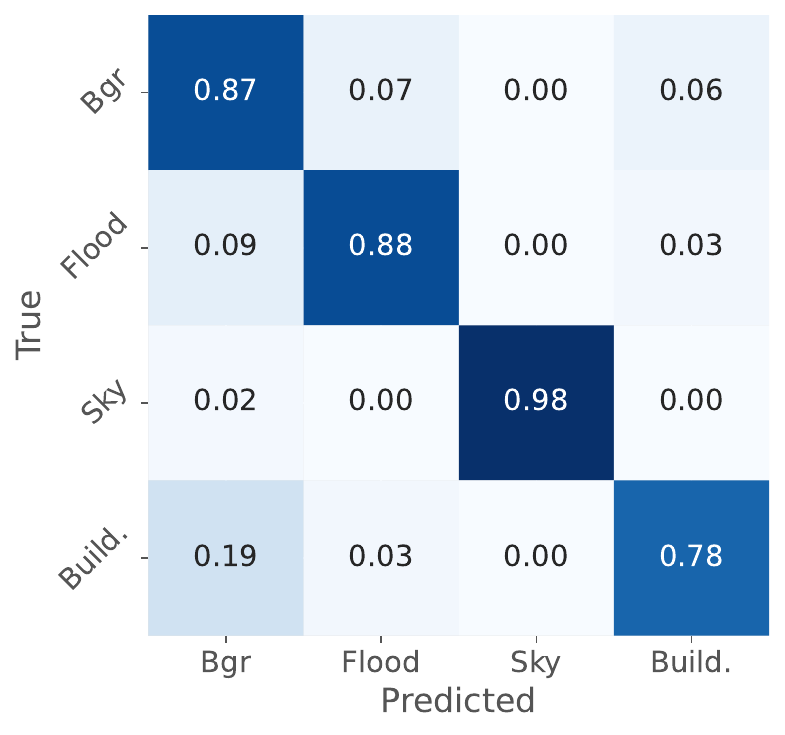}}
    \caption{Normalized confusion matrices of the best performing models, Swin-T and U-Net, both pre-trained.}
    \label{fig:cmunetswint}
\end{figure}

\begin{figure}[t!]
    \centering
    
    \subfigure[RGB distribution for classes Building, Sky, and Flood.]{    \includegraphics[width=0.48\textwidth]{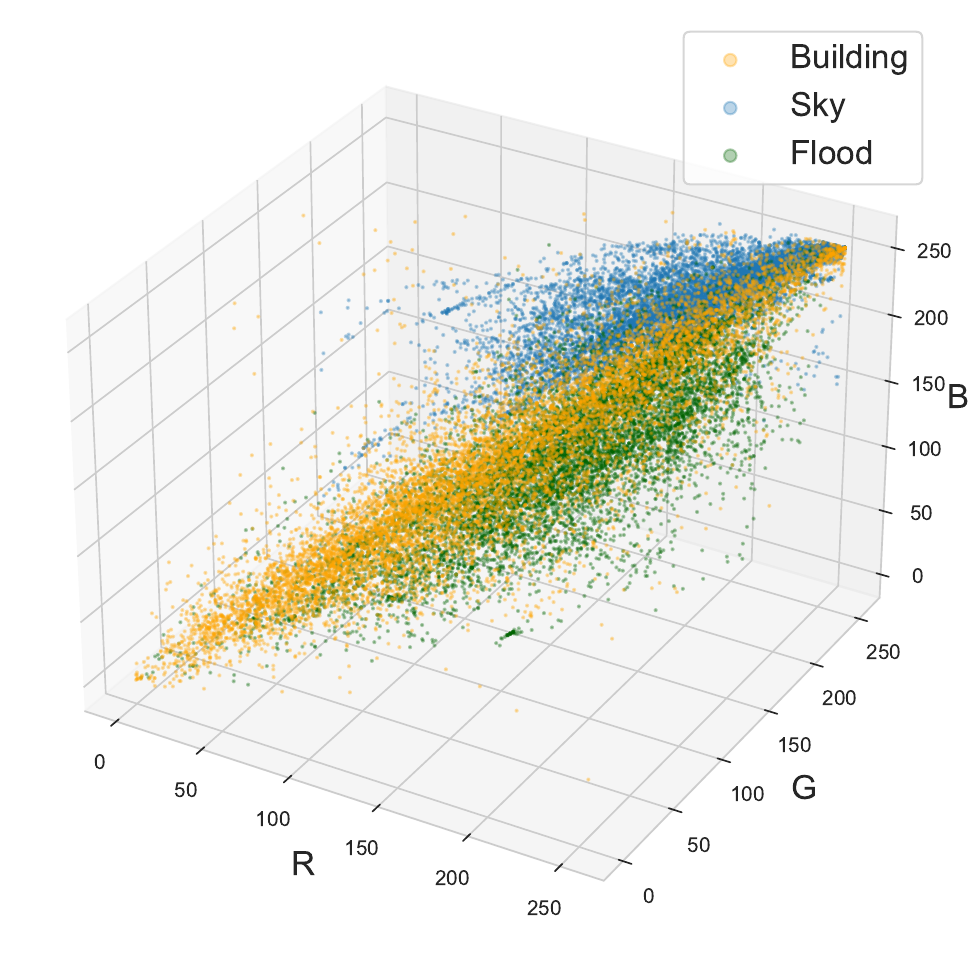}}
    \subfigure[RGB distribution for class Background.]{    \includegraphics[width=0.48\textwidth]{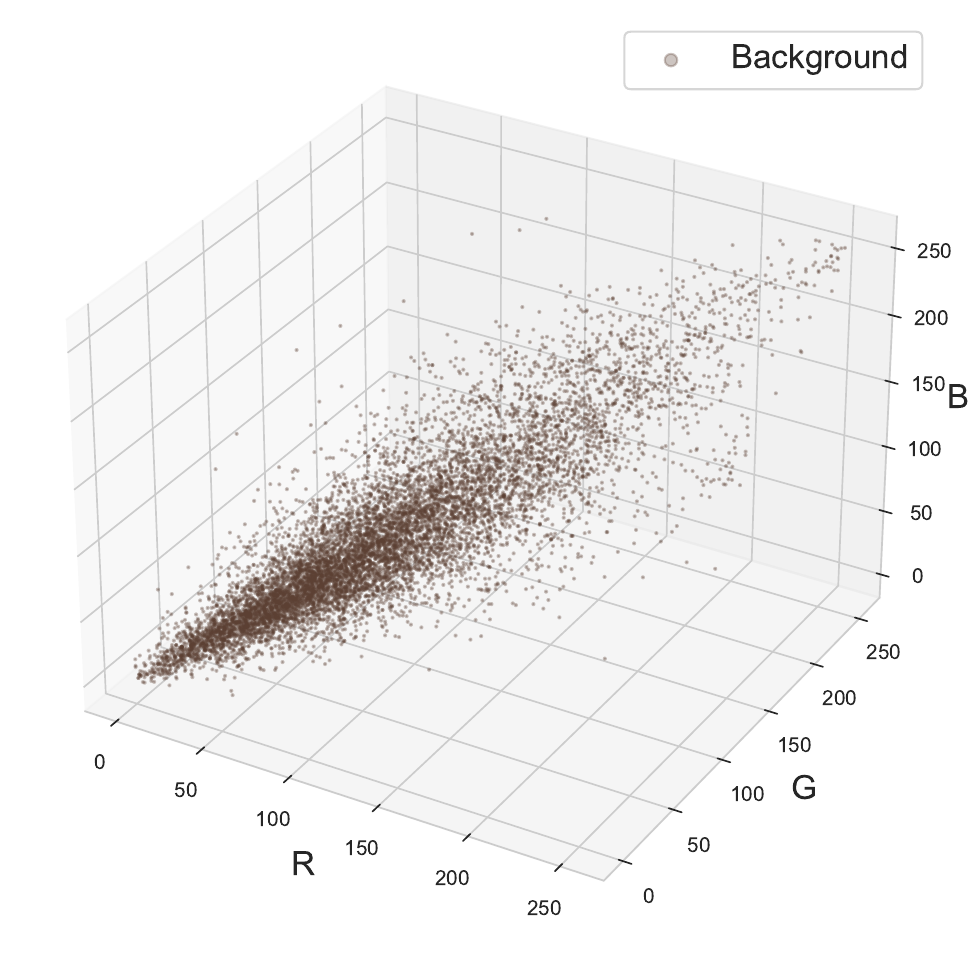}}
    
    \caption{RGB distributions per class.}
    \label{fig:rgbdistr}
\end{figure}

Nonetheless, all models consistently underperformed in the Building class, which was frequently misclassified as Flood or Background, as evidenced by the confusion matrices of the top-performing models (Swin-T and U-Net) shown in Figure~\ref{fig:cmunetswint}. A likely contributing factor is the similarity in RGB values between pixels corresponding to the \textit{Flood} and \textit{Building} classes, as illustrated in Figure~\ref{fig:rgbdistr}. This phenomenon arises because the color signature of flood water often closely resembles that of the surrounding land or built environment, particularly in urban regions where reflections, shadows, or turbid water can introduce additional ambiguity. In addition, buildings that are partially or fully surrounded by floodwater frequently exhibit boundary regions where the spectral values of water and building materials blend, resulting in color bleeding across adjacent pixels. These factors make the visual separation of \textit{Flood} and \textit{Building} classes inherently challenging when relying on RGB data alone.  It is important to note that pixels belonging to the \textit{Background} class are not included in this analysis, despite their occasional resemblance to floodwater or building color signatures. We deliberately exclude them for two reasons: (a) the \textit{Background} class does not play a direct role in the flood-related tasks under consideration, and (b) its content is heterogeneous and may correspond to any number of objects or surfaces (e.g., vegetation, sky, roads), making it unsuitable as a consistent reference class. 


The aforementioned findings further highlight the inherent challenge of accurately segmenting spatially sparse and fine-grained structures in aerial flood imagery, reinforcing the conclusion that \textit{AIFloodSense} constitutes a demanding benchmark for current state-of-the-art segmentation models.

\begin{figure*}[]
    \centering
    \vspace{-2.6cm}
    \renewcommand\tabularxcolumn[1]{>{\centering\arraybackslash}m{#1}}
    \begin{tabularx}{0.80\linewidth}{XXXX}
     \large{\textbf{P0}} & \large{\textbf{P25}} &\normalsize{\textbf{P50}} & \large{\textbf{P75}} 
    \end{tabularx}
    \\

    \subfigure[] {\includegraphics[width=0.21\textwidth]{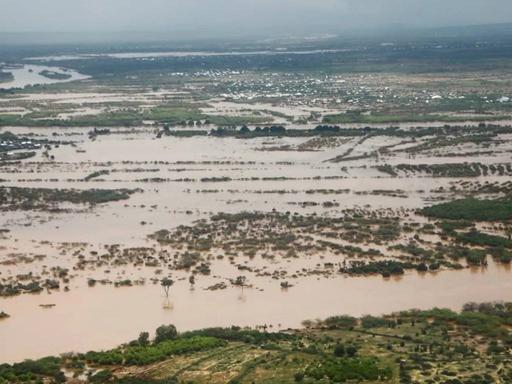}}
    \subfigure[] {\includegraphics[width=0.21\textwidth]{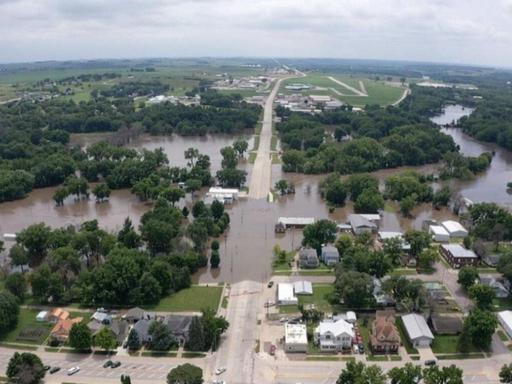}}
    \subfigure[] {\includegraphics[width=0.21\textwidth]{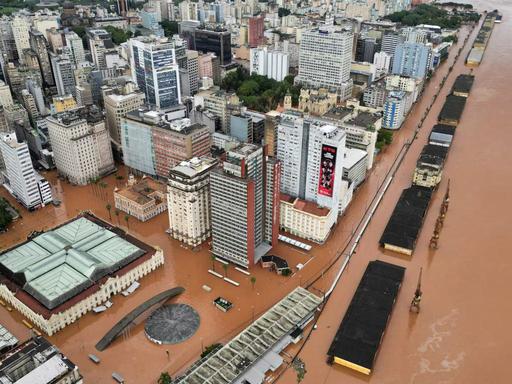}}
    \subfigure[] {\includegraphics[width=0.21\textwidth]{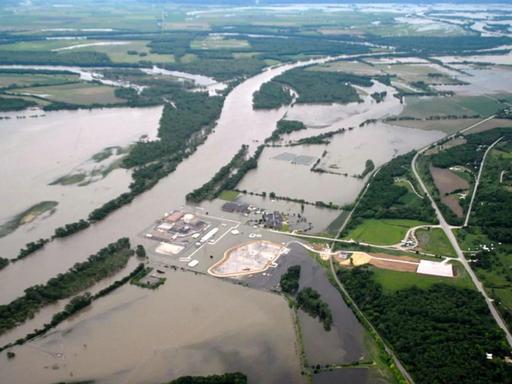}}
    \\

    \subfigure[] {\includegraphics[width=0.21\textwidth]{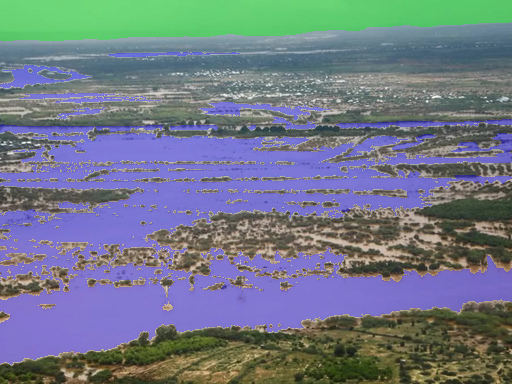}}
    \subfigure[] {\includegraphics[width=0.21\textwidth]{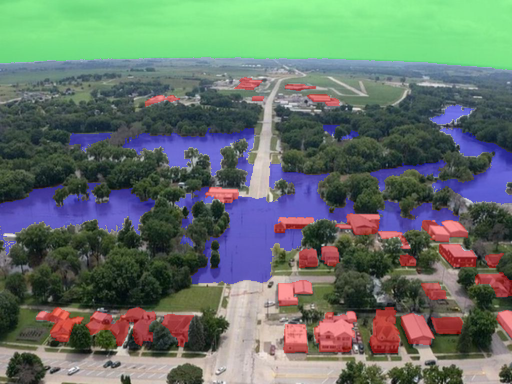}}
    \subfigure[] {\includegraphics[width=0.21\textwidth]{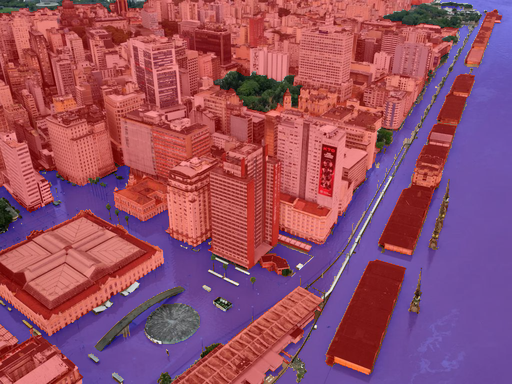}}
    \subfigure[] {\includegraphics[width=0.21\textwidth]{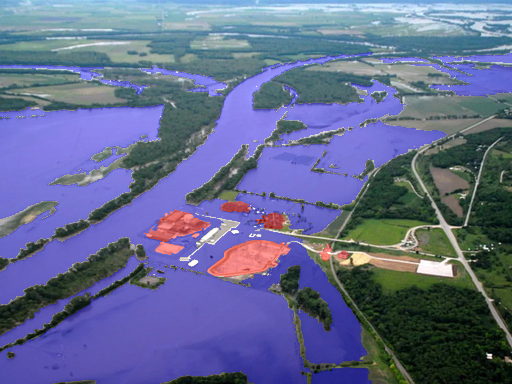}}
    \\
    
    \subfigure[\scriptsize mIoU = 86.93\%] {\includegraphics[width=0.21\textwidth]{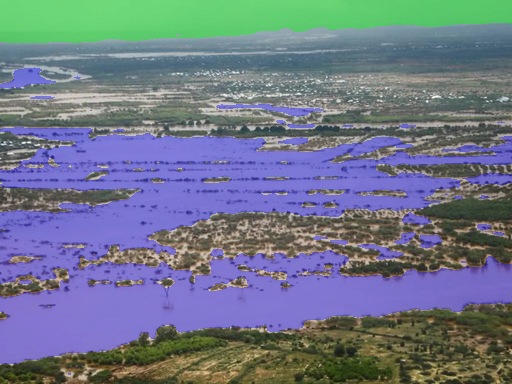}}
    \subfigure[\scriptsize mIoU = 77.69\%] {\includegraphics[width=0.21\textwidth]{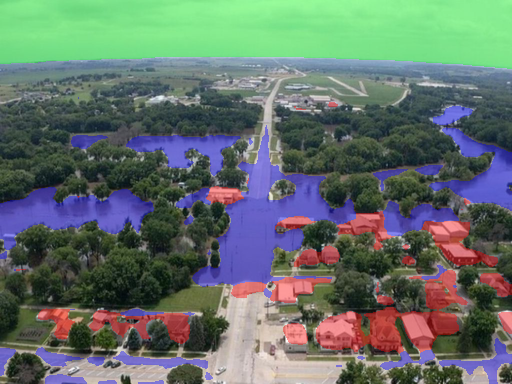}}
    \subfigure[\scriptsize mIoU = 71.37\%] {\includegraphics[width=0.21\textwidth]{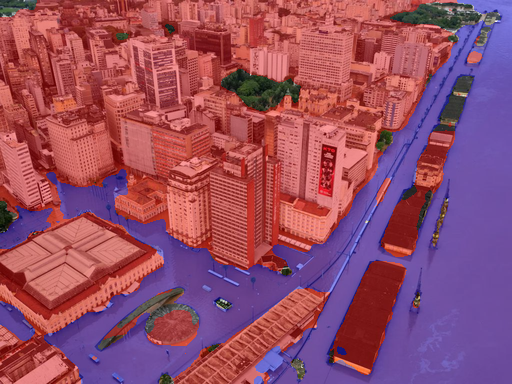}}
    \subfigure[\scriptsize mIoU = 63.32\%] {\includegraphics[width=0.21\textwidth]{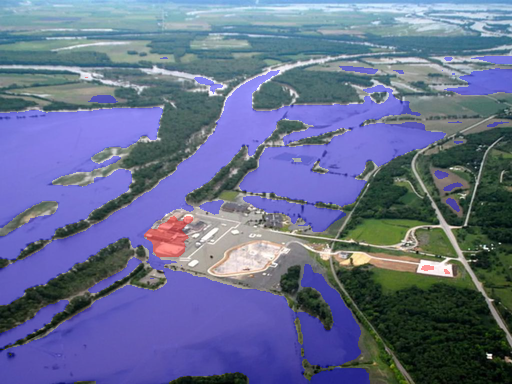}}
    \\

    \subfigure[\scriptsize mIoU = 79.93\%] {\includegraphics[width=0.21\textwidth]{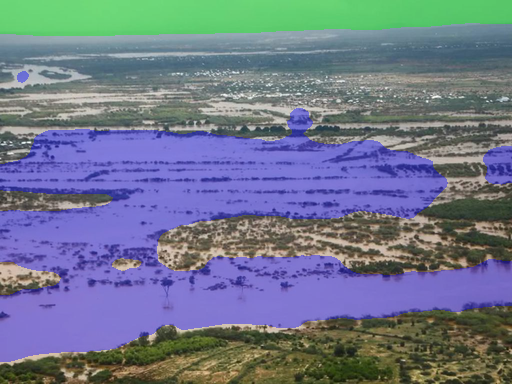}}
    \subfigure[\scriptsize mIoU = 68.48\%] {\includegraphics[width=0.21\textwidth]{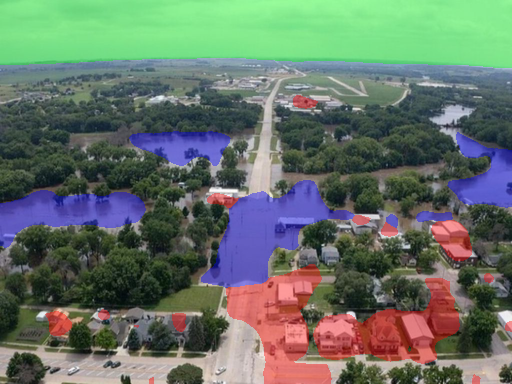}}
    \subfigure[\scriptsize mIoU = 61.96\%] {\includegraphics[width=0.21\textwidth]{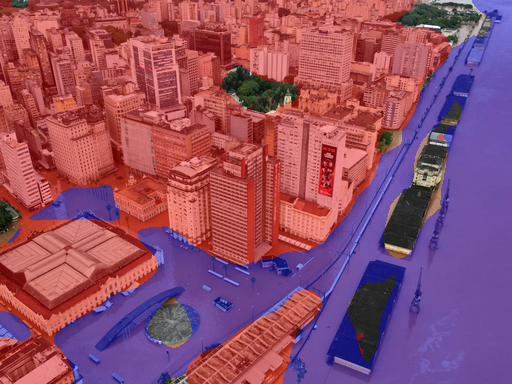}}
    \subfigure[\scriptsize mIoU = 69.58\%] {\includegraphics[width=0.21\textwidth]{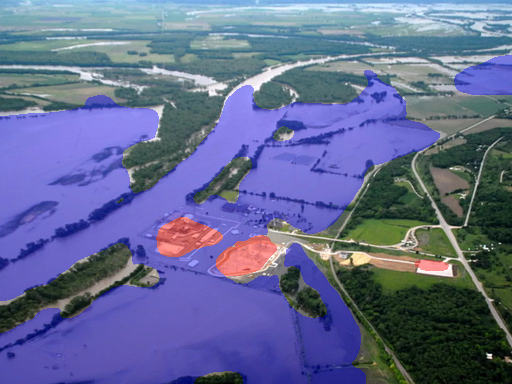}}
    \\

    \subfigure[\scriptsize mIoU = 84.72\%] {\includegraphics[width=0.21\textwidth]{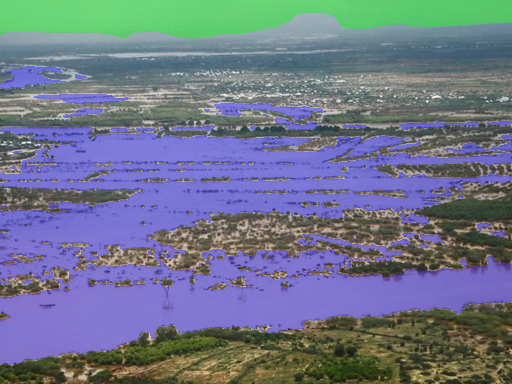}}
    \subfigure[\scriptsize mIoU = 71.84\%] {\includegraphics[width=0.21\textwidth]{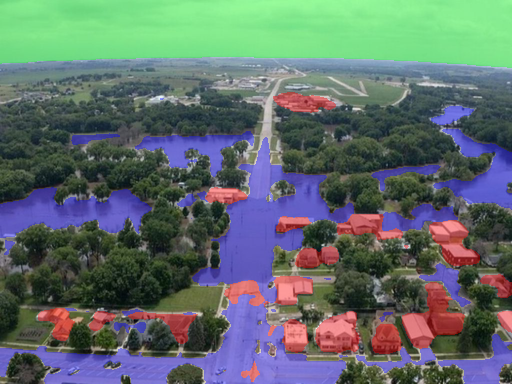}}
    \subfigure[\scriptsize mIoU = 70.71\%] {\includegraphics[width=0.21\textwidth]{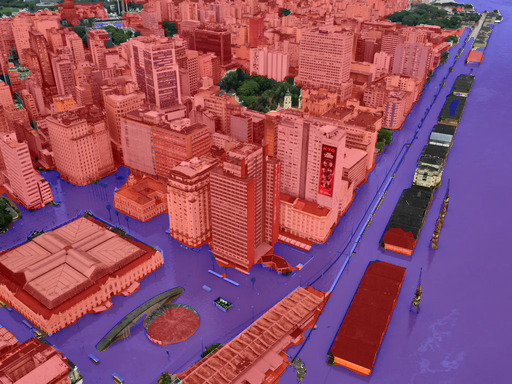}}
    \subfigure[\scriptsize mIoU = 58.28\%] {\includegraphics[width=0.21\textwidth]{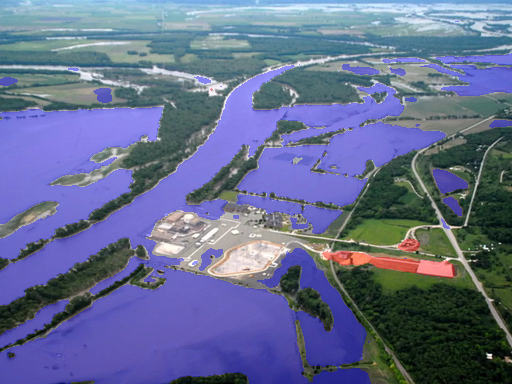}}
 \\

    \subfigure[\scriptsize mIoU = 78.58\%] {\includegraphics[width=0.21\textwidth]{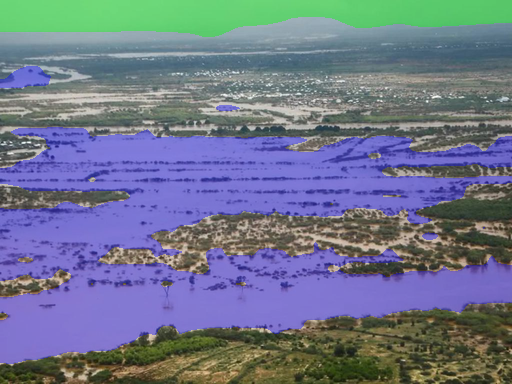}}
    \subfigure[\scriptsize mIoU = 71.54\%] {\includegraphics[width=0.21\textwidth]{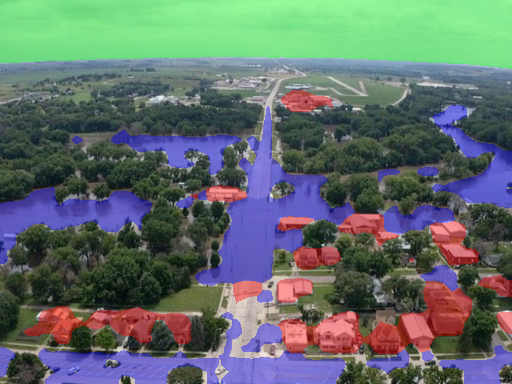}}
    \subfigure[\scriptsize mIoU = 69.95\%] {\includegraphics[width=0.21\textwidth]{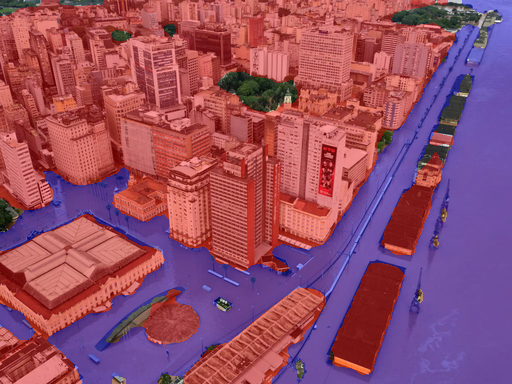}}
    \subfigure[\scriptsize mIoU = 56.52\%] {\includegraphics[width=0.21\textwidth]{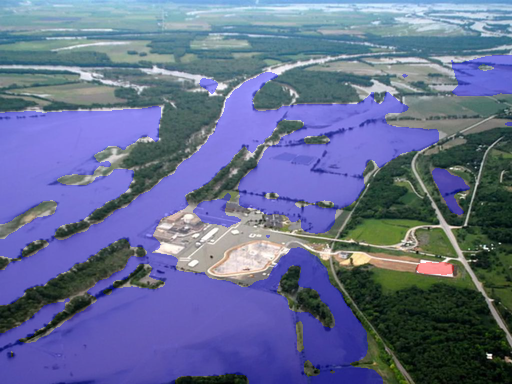}}
    \caption{Qualitative comparison of multiclass segmentation results. The rows display, from top to bottom: original images, ground truth overlays, and prediction overlays from the pre-trained Swin-T, SegFormer-B0, U-Net, and FCN models. Segmentation masks are visualized using the following color scheme: blue (flood), green (sky), and red (building). Columns are organized by sample difficulty based on the top-performing model (Swin-T), where \textit{Pxx} denotes the sample corresponding to the \textit{xx-th} percentile of the per-image mIoU distribution, sorted in descending order (best to worst).    
    }
    \label{fig:segmresmulti}
\end{figure*}

To present qualitative results, we additionally compute a per-image mIoU for the best performing model, Swin-T: for each test image, we calculate the IoU of each class present in the ground truth, average over these classes, and use this value to sort the images in a descending order. This ensures that the example images are representative across a range of model performance levels. In Figure~\ref{fig:segmresmulti} we then show the 0th (best), 25th, 50th, and 75th percentiles of the per-image mIoU distribution. For comparative reasons, we also include the other transformer model, SegFormer-B0, and the two best convolutional models, U-Net and FCN. For visualization and mask encoding, we adopt an overlaid colormap: the flood class is represented as blue, the sky class as green, and the building class as red.

Qualitative segmentation examples further illustrate the quantitative trends discussed earlier. The Swin-T model produces the most coherent and spatially consistent segmentations, effectively delineating complex boundaries and capturing fine details across all classes (Fig.~\ref{fig:segmresmulti} (i)--(k)). Although the U-Net architecture secured the second-highest quantitative ranking, a qualitative inspection revealed distinct limitations in boundary adherence. Despite its competitive metric performance, the model exhibits a tendency to over-segment (flood in (r) and building in (t)) or under-segment (sky in (q) and building in (s)) class regions, frequently extending predictions beyond or failing to reach exact object contours. This behavior suggests that, while U-Net captures global context effectively, it struggles with fine-grained localization in transition zones where visual features between classes are ambiguous. In contrast, SegFormer-B0 exhibits noticeably poorer delineation of object boundaries and misclassification across regions (m)--(o), confirming its underperformance relative to Swin-T and the convolutional models. Although case (p) greater mIoU bl bla The visual artifacts in these results are consistent with the model’s limited capacity to integrate fine spatial detail with broader contextual cues, as previously described.


\noindent \textbf{Binary Protocol:} In order to further investigate the intrinsic segmentation difficulty of each class, we reformulated the task as a binary segmentation problem. Table~\ref{tab:segm2classes} presents the results of these experiments on the test set, where the first class corresponds to the target category (Flood, Sky, or Building) and the second class aggregates all remaining classes along with the background.

\begin{table*}[]
\centering
\caption{Performance comparison of baseline semantic segmentation models on the AIFloodSense test set under the binary segmentation protocol. Results are reported as percentages (\%), with and without ImageNet pretraining.}
\label{tab:segm2classes}
\setlength{\tabcolsep}{3pt}

\resizebox{\linewidth}{!}{%
\begin{tabular}{|l||c||cccc||cccc||cccc|}
\hline
\multirow{2}{*}{\textbf{Model}} & \multirow{2}{*}{\begin{tabular}[c]{@{}c@{}}\textbf{Pre-}\\\textbf{trained}\end{tabular}} & \multicolumn{4}{c||}{\textbf{Flood}} & \multicolumn{4}{c||}{\textbf{Sky}} & \multicolumn{4}{c|}{\textbf{Build.}} \\ \cline{3-14} 
 &  & \multicolumn{1}{c|}{\textbf{IoU}} & \multicolumn{1}{c|}{\textbf{Pr.}} & \multicolumn{1}{c|}{\textbf{Rec.}} & \textbf{F$_1$} & \multicolumn{1}{c|}{\textbf{IoU}} & \multicolumn{1}{c|}{\textbf{Pr.}} & \multicolumn{1}{c|}{\textbf{Rec.}} & \textbf{F$_1$} & \multicolumn{1}{c|}{\textbf{IoU}} & \multicolumn{1}{c|}{\textbf{Pr.}} & \multicolumn{1}{c|}{\textbf{Rec.}} & \textbf{F$_1$} \\ \hline
 \hline
DeepLabV3 & \checkmark & \multicolumn{1}{c|}{80.42} & \multicolumn{1}{c|}{88.37} & \multicolumn{1}{c|}{89.94} & 89.15 & \multicolumn{1}{c|}{93.52} & \multicolumn{1}{c|}{94.75} & \multicolumn{1}{c|}{98.63} & 96.65 & \multicolumn{1}{c|}{60.84} & \multicolumn{1}{c|}{73.96} & \multicolumn{1}{c|}{77.43} & 75.66 \\ \hline
FCN & \checkmark & \multicolumn{1}{c|}{80.17} & \multicolumn{1}{c|}{88.85} & \multicolumn{1}{c|}{89.13} & 88.99 & \multicolumn{1}{c|}{91.27} & \multicolumn{1}{c|}{93.11} & \multicolumn{1}{c|}{97.88} & 95.44 & \multicolumn{1}{c|}{62.47} & \multicolumn{1}{c|}{75.05} & \multicolumn{1}{c|}{78.85} & 76.90 \\ \hline
U-Net & \checkmark & \multicolumn{1}{c|}{84.14} & \multicolumn{1}{c|}{\textbf{92.61}} & \multicolumn{1}{c|}{90.19} & 91.39 & \multicolumn{1}{c|}{\textbf{93.72}} & \multicolumn{1}{c|}{\textbf{95.26}} & \multicolumn{1}{c|}{98.31} & \textbf{96.76} & \multicolumn{1}{c|}{63.90} & \multicolumn{1}{c|}{\textbf{78.51}} & \multicolumn{1}{c|}{77.45} & 77.98 \\ \hline
SegFormer-B0 & \checkmark & \multicolumn{1}{c|}{80.01} & \multicolumn{1}{c|}{89.63} & \multicolumn{1}{c|}{88.17} & 88.89 & \multicolumn{1}{c|}{93.30} & \multicolumn{1}{c|}{95.08} & \multicolumn{1}{c|}{98.03} & 96.54 & \multicolumn{1}{c|}{57.78} & \multicolumn{1}{c|}{69.70} & \multicolumn{1}{c|}{77.16} & 73.24 \\ \hline
Swin-T & \checkmark & \multicolumn{1}{c|}{\textbf{84.29}} & \multicolumn{1}{c|}{91.59} & \multicolumn{1}{c|}{\textbf{91.37}} & \textbf{91.48} & \multicolumn{1}{c|}{93.71} & \multicolumn{1}{c|}{94.71} & \multicolumn{1}{c|}{\textbf{98.89}} & 96.75 & \multicolumn{1}{c|}{\textbf{65.46}} & \multicolumn{1}{c|}{74.87} & \multicolumn{1}{c|}{\textbf{83.90}} & \textbf{79.13} \\ \hline
\hline
DeepLabV3 & \ding{55} & \multicolumn{1}{c|}{70.21} & \multicolumn{1}{c|}{87.62} & \multicolumn{1}{c|}{77.94} & 82.50 & \multicolumn{1}{c|}{83.59} & \multicolumn{1}{c|}{90.08} & \multicolumn{1}{c|}{92.06} & 91.06 & \multicolumn{1}{c|}{37.65} & \multicolumn{1}{c|}{43.29} & \multicolumn{1}{c|}{74.31} & 54.71 \\ \hline
FCN & \ding{55} & \multicolumn{1}{c|}{70.47} & \multicolumn{1}{c|}{88.61} & \multicolumn{1}{c|}{77.49} & 82.68 & \multicolumn{1}{c|}{84.22} & \multicolumn{1}{c|}{92.38} & \multicolumn{1}{c|}{90.51} & 91.43 & \multicolumn{1}{c|}{41.82} & \multicolumn{1}{c|}{50.69} & \multicolumn{1}{c|}{70.50} & 58.98 \\ \hline
U-Net & \ding{55} & \multicolumn{1}{c|}{75.79} & \multicolumn{1}{c|}{90.38} & \multicolumn{1}{c|}{82.44} & 86.23 & \multicolumn{1}{c|}{83.34} & \multicolumn{1}{c|}{93.53} & \multicolumn{1}{c|}{88.44} & 90.91 & \multicolumn{1}{c|}{44.78} & \multicolumn{1}{c|}{52.18} & \multicolumn{1}{c|}{75.95} & 61.86 \\ \hline
SegFormer-B0 & \ding{55} & \multicolumn{1}{c|}{68.32} & \multicolumn{1}{c|}{86.52} & \multicolumn{1}{c|}{76.46} & 81.18 & \multicolumn{1}{c|}{68.65} & \multicolumn{1}{c|}{86.30} & \multicolumn{1}{c|}{77.04} & 81.41 & \multicolumn{1}{c|}{34.79} & \multicolumn{1}{c|}{39.62} & \multicolumn{1}{c|}{74.07} & 51.63 \\ \hline
Swin-T & \ding{55} & \multicolumn{1}{c|}{69.55} & \multicolumn{1}{c|}{83.04} & \multicolumn{1}{c|}{81.07} & 82.04 & \multicolumn{1}{c|}{74.29} & \multicolumn{1}{c|}{80.59} & \multicolumn{1}{c|}{90.48} & 85.25 & \multicolumn{1}{c|}{33.84} & \multicolumn{1}{c|}{38.00} & \multicolumn{1}{c|}{75.56} & 50.57 \\ \hline
\end{tabular}
}
\end{table*}

Consistent with the observations in the multi-class segmentation setting, models initialized with pre-trained weights consistently outperform those trained from scratch across all evaluation metrics, emphasizing the advantages of transfer learning for this task. Pre-training provides particularly notable improvements for all classes, with IoU and F$_1$ gains of 8.35--14.74\% and 5.16--9.44\% for Flood, 7.05--24.65\% and 4.01--15.13\% for Sky, and 19.12--31.62\% and 16.12--28.56\% for Building, respectively. Especially the transformer models benefit from ImageNet pre-training, and among classes, the Building class exhibits a performance boost.

Regarding class-specific performance, Sky segmentation is the least challenging, with IoU scores exceeding 91\% for all pre-trained models and reaching up to 93.71\% for the Swin-T, and 93.72\% for the U-Net. Both models in their pre-trained setting resulted in almost identical F$_1$ scores, 96.75\% and 96.76\% respectively, where the first excelled in Precision and the latter in Recall. This superior performance can be attributed to the distinct color signature of sky pixels, which facilitates their separation from other classes (see Figure~\ref{fig:cmunetswint}). Flood segmentation presents an intermediate level of difficulty, with Swin-T achieving the highest IoU (84.29\%) and F$_1$ score (91.48\%) among pre-trained models, followed this time by the U-Net (84.14\% and 91.39\% respectively). Building segmentation is the most challenging, reflected in lower IoU values (down to 33.84\% when trained from scratch, and 57.78\% in the respective pre-trained versions). While Swin-T demonstrates the best performance in IoU, Recall, and F$_1$ in the Flood and Building classes, U-Net achieves the highest IoU, Precision, and F$_1$ score in the Sky class. For all classes, U-Net showed the best Precision, whereas Swin-T excelled in Recall. Overall, these results suggest that architectures initialized with pre-trained weights offer the most consistent and robust performance in binary segmentation tasks, particularly for classes characterized by high visual variability and structural complexity.

A comparison between the binary and multiclass segmentation IoUs (Table~\ref{tab:segm4classes} Per-class IoU columns) reveals that binary models generally outperform their multiclass counterparts, except for the Sky class, where the multitask setting can yield slightly higher scores. However, the differences remain minor regarding the best performing models, 1.86\% and 2.48\%, respectively, for the Flood and Building classes with Swin-T, and -0.44\% for the Sky class with U-Net, indicating that the added complexity of learning multiple classes simultaneously does not substantially hinder performance. This suggests that the evaluated architectures maintain strong feature generalization and class separation capabilities, even when trained jointly across visually diverse categories. Moreover, the limited performance degradation in the multiclass setting highlights the robustness and transferability of representations learned by modern encoder-decoder architectures, particularly those leveraging pre-trained weights. These findings reinforce the potential of such models for scalable flood analysis pipelines, where multiple scene attributes must be inferred concurrently from the same visual input.

\begin{figure*}[]
    \centering
    \vspace{-2.6cm}
    \renewcommand\tabularxcolumn[1]{>{\centering\arraybackslash}m{#1}}
    \begin{tabularx}{0.80\linewidth}{XXXX}
     \large{\textbf{P0}} & \large{\textbf{P25}} & \large{\textbf{P50}} & \large{\textbf{P75}} 
    \end{tabularx}\\

    \subfigure[] {\includegraphics[width=0.21\textwidth]{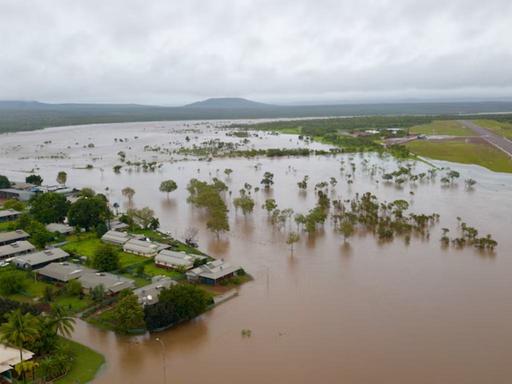}}
    \subfigure[] {\includegraphics[width=0.21\textwidth]{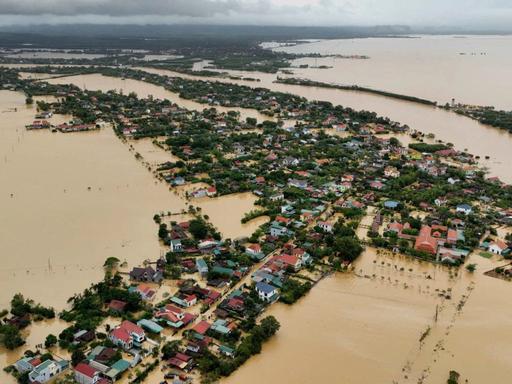}}
    \subfigure[] {\includegraphics[width=0.21\textwidth]{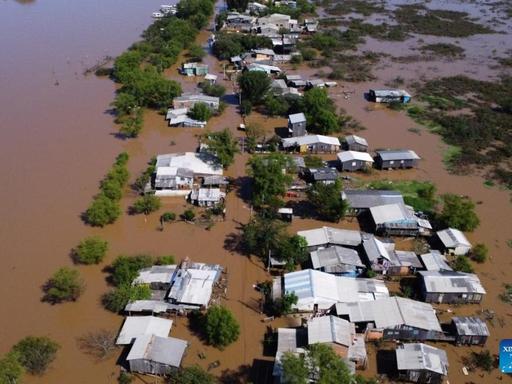}}
    \subfigure[] {\includegraphics[width=0.21\textwidth]{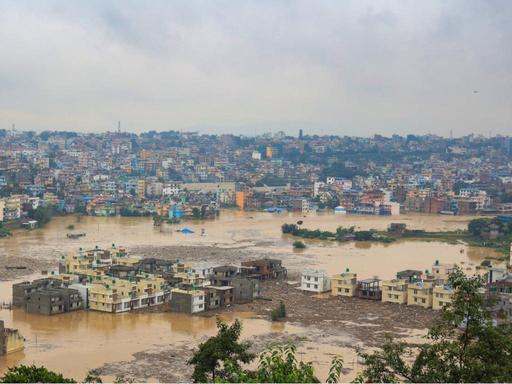}}
    \\

    \subfigure[] {\includegraphics[width=0.21\textwidth]{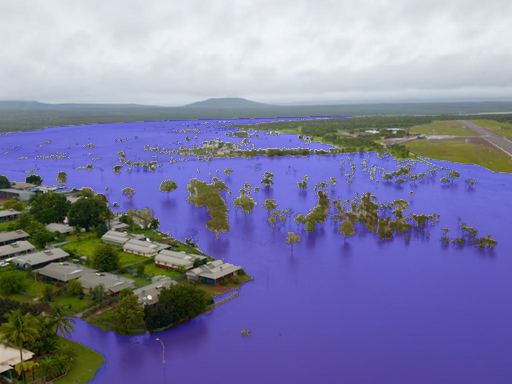}}
    \subfigure[] {\includegraphics[width=0.21\textwidth]{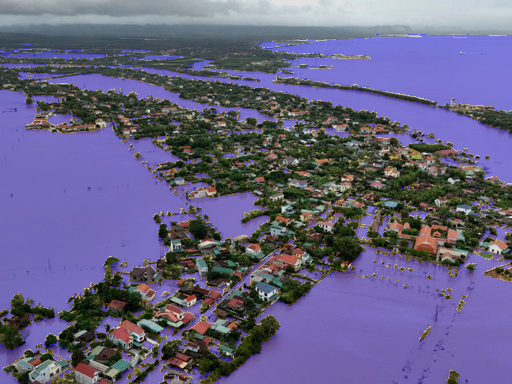}}
    \subfigure[] {\includegraphics[width=0.21\textwidth]{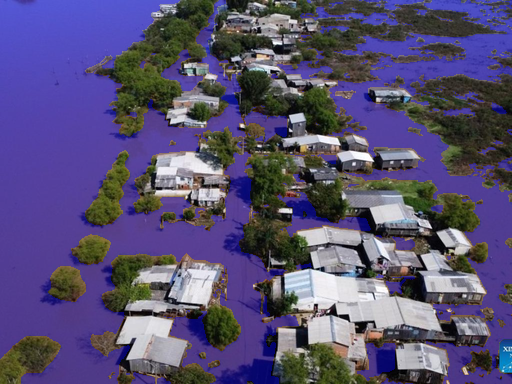}}
    \subfigure[] {\includegraphics[width=0.21\textwidth]{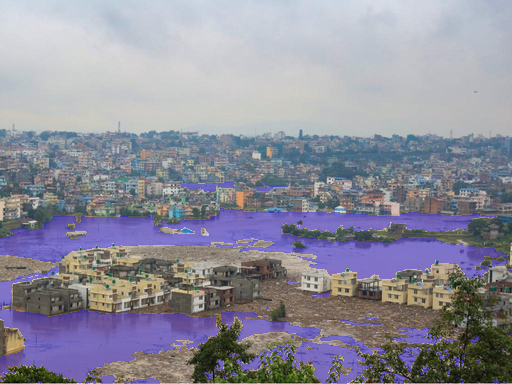}}
    \\
    
    \subfigure[IoU = 93.82\%]{\includegraphics[width=0.21\textwidth]{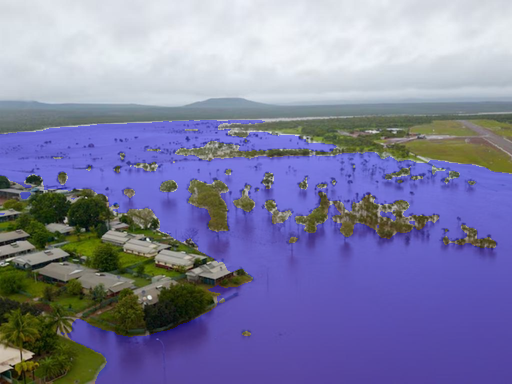}}
    \subfigure[IoU = 88.82\%] {\includegraphics[width=0.21\textwidth]{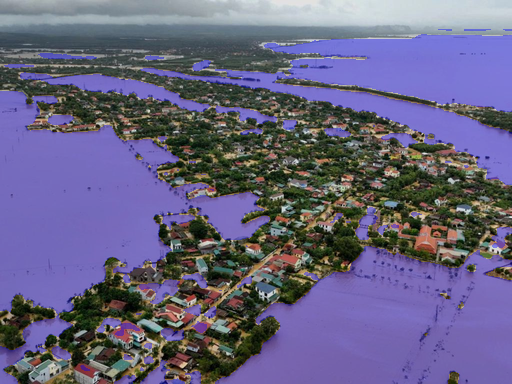}}
    \subfigure[IoU = 83.86\%] {\includegraphics[width=0.21\textwidth]{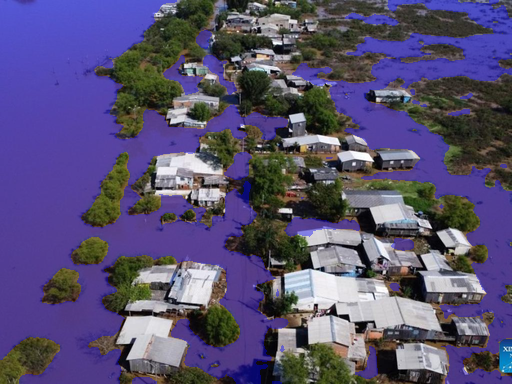}}
    \subfigure[IoU = 78.33\%] {\includegraphics[width=0.21\textwidth]{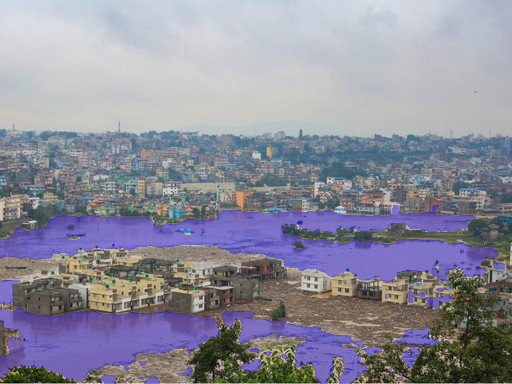}}
    \\

    \subfigure[IoU = 90.68\%]{\includegraphics[width=0.21\textwidth]{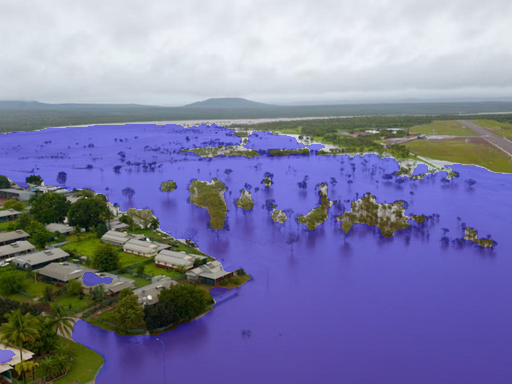}}
    \subfigure[IoU = 84.27\%] {\includegraphics[width=0.21\textwidth]{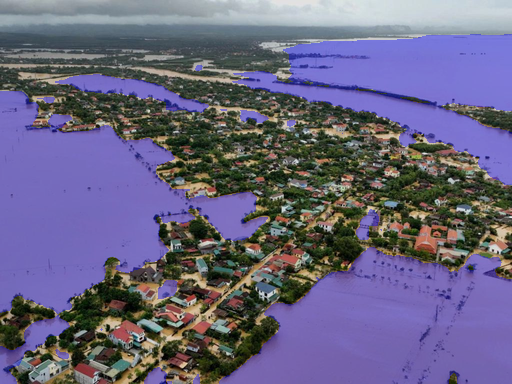}}
    \subfigure[IoU = 78.44\%] {\includegraphics[width=0.21\textwidth]{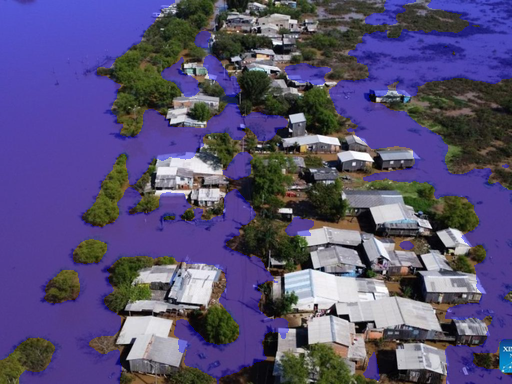}}
    \subfigure[IoU = 72.64\%] {\includegraphics[width=0.21\textwidth]{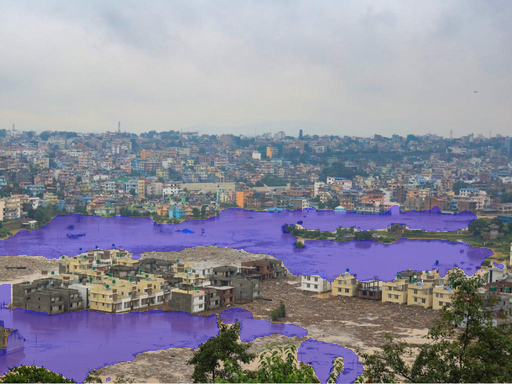}}
    \\

    \subfigure[IoU = 95.16\%]{\includegraphics[width=0.21\textwidth]{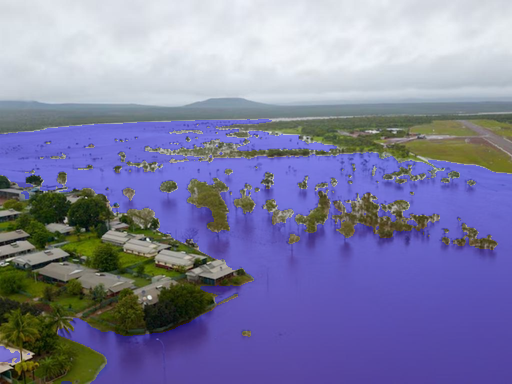}}
    \subfigure[IoU = 90.91\%] {\includegraphics[width=0.21\textwidth]{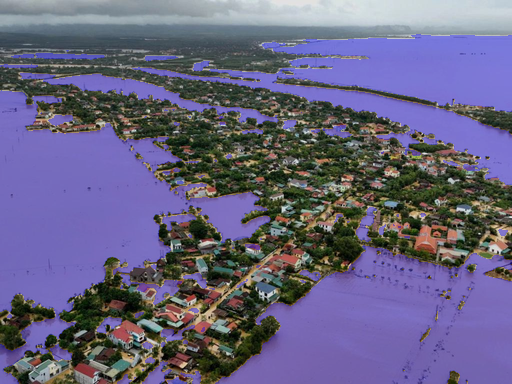}}
    \subfigure[IoU = 82.76\%] {\includegraphics[width=0.21\textwidth]{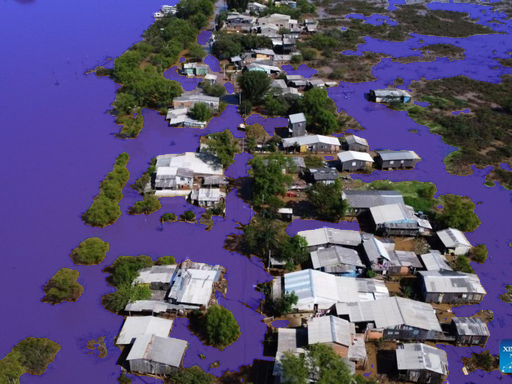}}
    \subfigure[IoU = 75.22\%] {\includegraphics[width=0.21\textwidth]{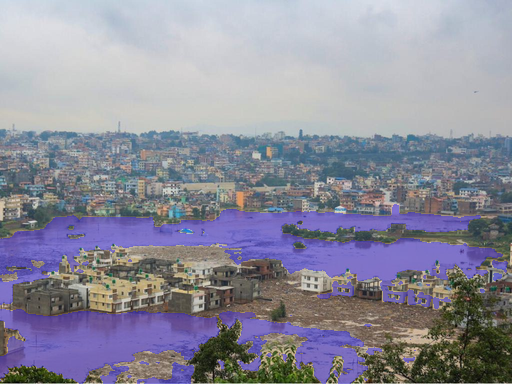}}
    \\

    \subfigure[IoU = 91.59\%]{\includegraphics[width=0.21\textwidth]{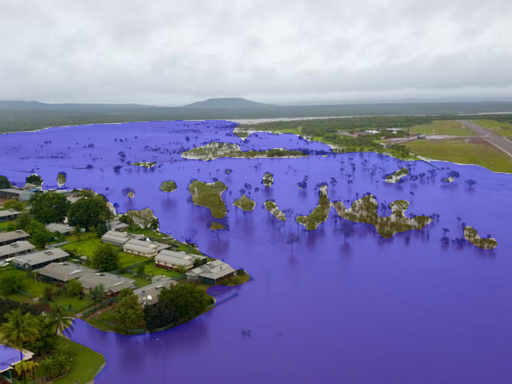}}
    \subfigure[IoU = 84.92\%] {\includegraphics[width=0.21\textwidth]{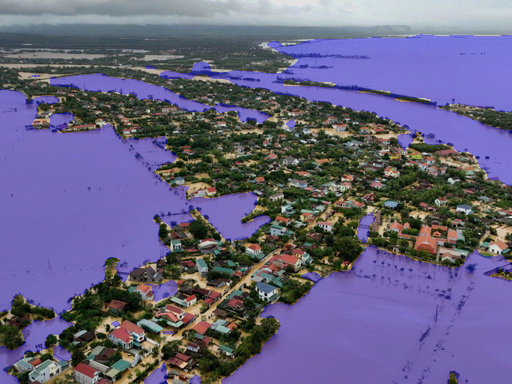}}
    \subfigure[IoU = 80.82\%] {\includegraphics[width=0.21\textwidth]{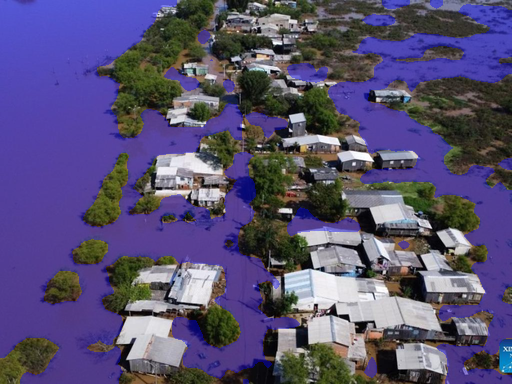}}
    \subfigure[IoU = 72.11\%] {\includegraphics[width=0.21\textwidth]{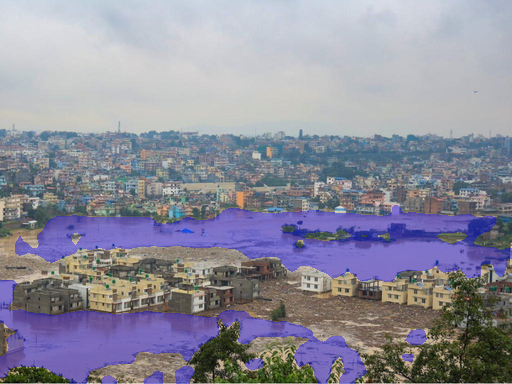}}
    
    \caption{Qualitative comparison of binary segmentation results for class flood. The rows display, from top to bottom: original images, ground truth overlays, and prediction overlays from the pre-trained Swin-T, SegFormer-B0, U-Net, and DeepLabV3 models. Segmentation masks are visualized using blue color for flood. Columns are organized by sample difficulty based on the overall top-performing model (Swin-T), where \textit{Pxx} denotes the sample corresponding to the \textit{xx-th} percentile of the per-image mIoU distribution, sorted in descending order (best to worst).
    }
    \label{fig:segmresbinflood}
\end{figure*}

\begin{figure*}[]
    \centering
    \vspace{-2.6cm}
    \renewcommand\tabularxcolumn[1]{>{\centering\arraybackslash}m{#1}}
    \begin{tabularx}{0.60\linewidth}{XXXX}
     \large{\textbf{Best}} & \large{\textbf{Median}} & \large{\textbf{Worst}} 
    \end{tabularx}\\

    \subfigure[] {\includegraphics[width=0.21\textwidth]{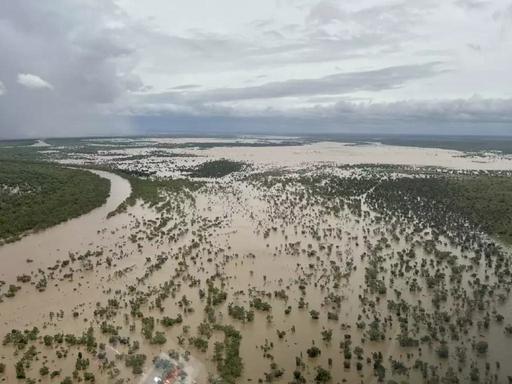}}
    \subfigure[] {\includegraphics[width=0.21\textwidth]{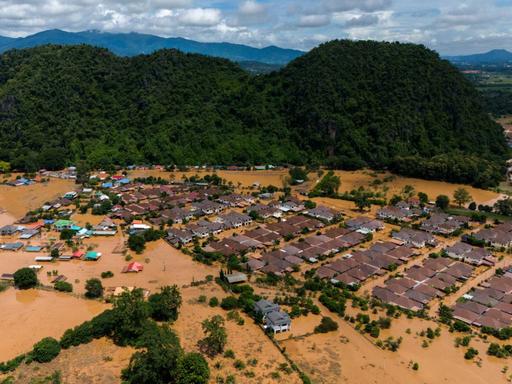}}
    \subfigure[] {\includegraphics[width=0.21\textwidth]{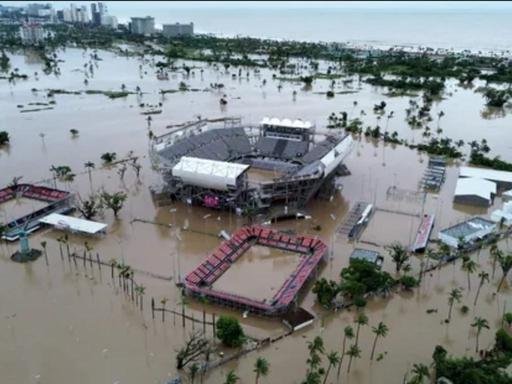}}
    \\

    \subfigure[] {\includegraphics[width=0.21\textwidth]{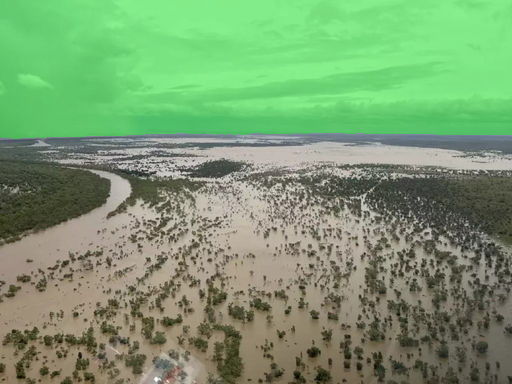}}
    \subfigure[] {\includegraphics[width=0.21\textwidth]{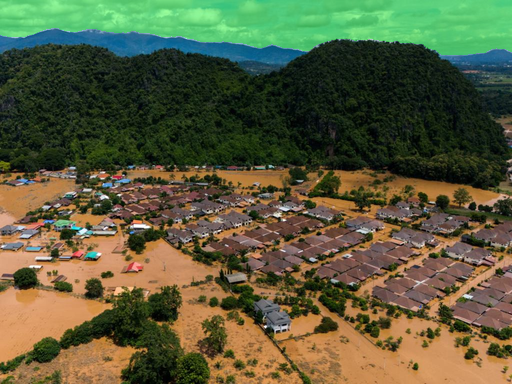}}
    \subfigure[] {\includegraphics[width=0.21\textwidth]{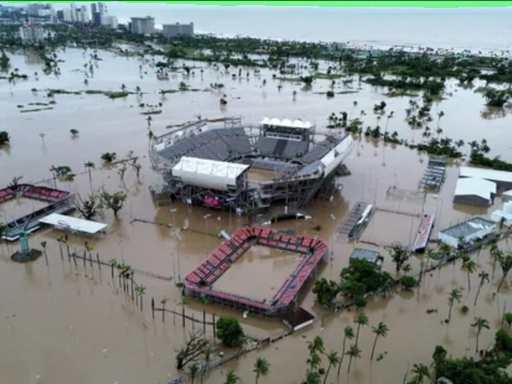}}
    \\
    
    \subfigure[IoU = 99.44\%]{\includegraphics[width=0.21\textwidth]{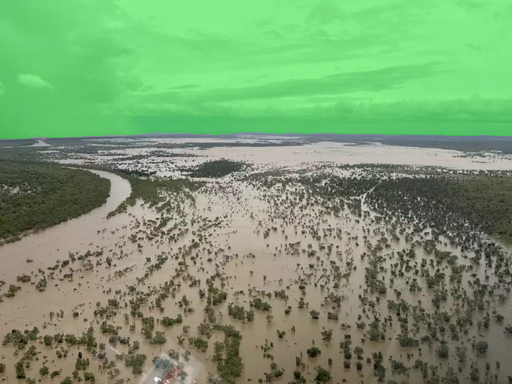}}
    \subfigure[IoU = 94.81\%] {\includegraphics[width=0.21\textwidth]{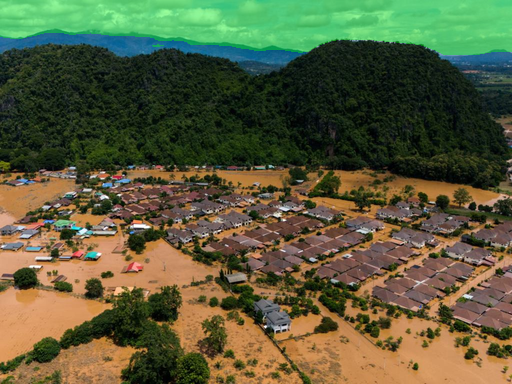}}
    \subfigure[IoU = 20.94\%] {\includegraphics[width=0.21\textwidth]{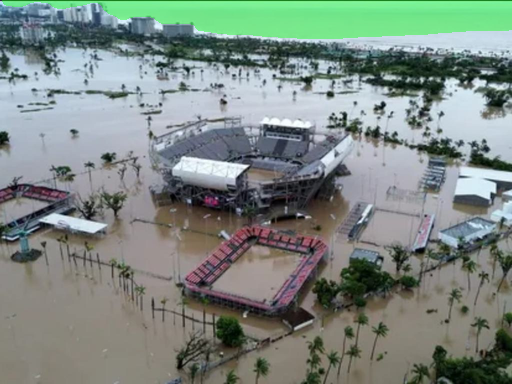}}
    \\

    \subfigure[IoU = 99.10\%]{\includegraphics[width=0.21\textwidth]{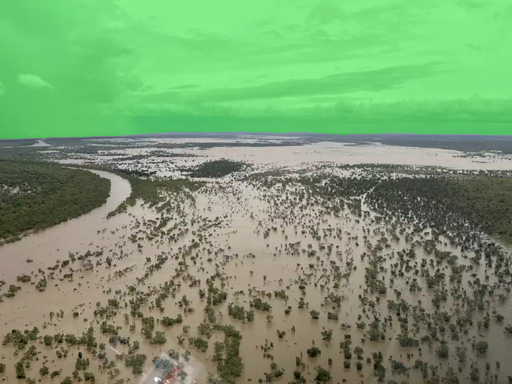}}
    \subfigure[IoU = 95.15\%] {\includegraphics[width=0.21\textwidth]{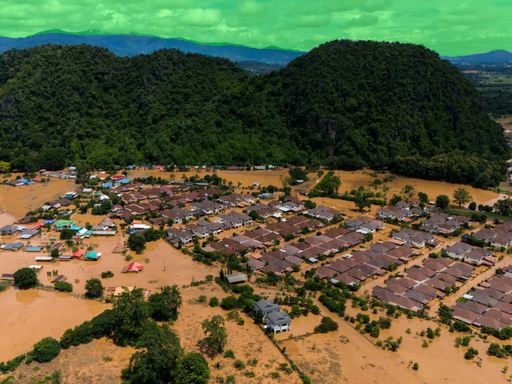}}
    \subfigure[IoU = 15.98\%] {\includegraphics[width=0.21\textwidth]{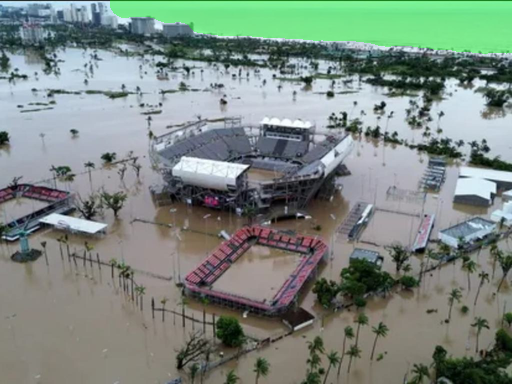}}
    \\

    \subfigure[IoU = 99.30\%]{\includegraphics[width=0.21\textwidth]{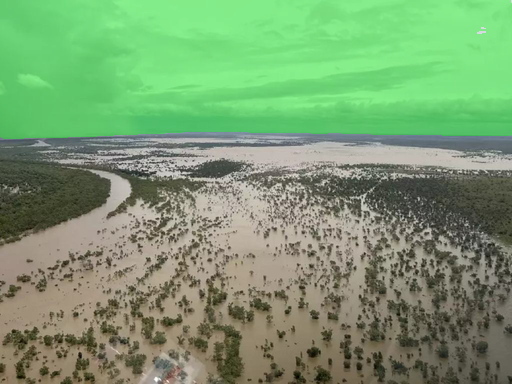}}
    \subfigure[IoU = 98.21\%] {\includegraphics[width=0.21\textwidth]{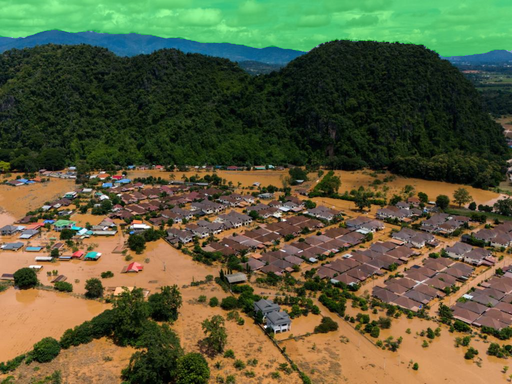}}
    \subfigure[IoU = 14.22\%] {\includegraphics[width=0.21\textwidth]{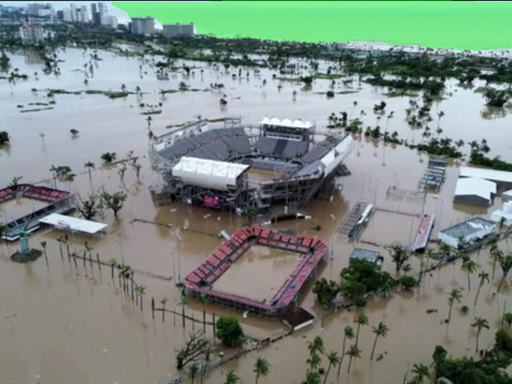}}
    \\

    \subfigure[IoU = 99.39\%]{\includegraphics[width=0.21\textwidth]{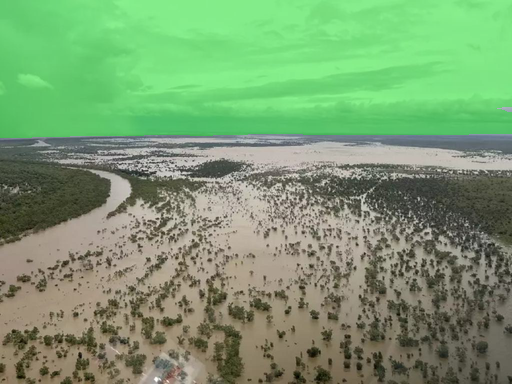}}
    \subfigure[IoU = 96.53\%] {\includegraphics[width=0.21\textwidth]{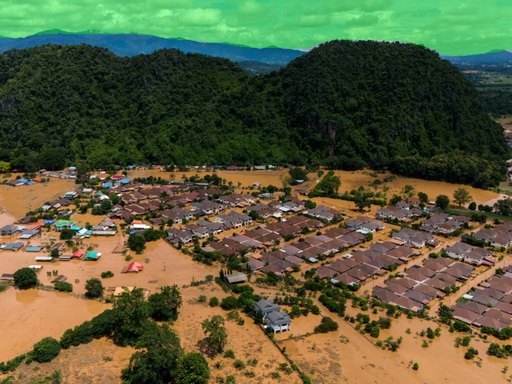}}
    \subfigure[IoU = 16.01\%] {\includegraphics[width=0.21\textwidth]{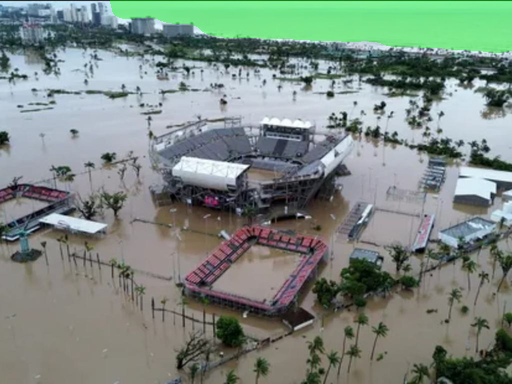}}
    
    \caption{Qualitative comparison of binary segmentation results for class sky. Rows display, from top to bottom: original images, ground truth overlays, and prediction overlays from the pre-trained Swin-T, SegFormer-B0, U-Net, and DeepLabV3 models. Segmentation masks are visualized using green color for sky. Columns correspond to best, median, and worst segmentation results according to the descending sorted per-image IoU values of the top-performing overall model, Swin-T.
    }
    \label{fig:segmresbinsky}
\end{figure*}

\begin{figure*}[]
    \centering
    \vspace{-2.6cm}
    \renewcommand\tabularxcolumn[1]{>{\centering\arraybackslash}m{#1}}
    \begin{tabularx}{0.60\linewidth}{XXXX}
     \large{\textbf{Best}} & \large{\textbf{Median}} & \large{\textbf{Worst}} 
    \end{tabularx}\\

    \subfigure[] {\includegraphics[width=0.21\textwidth]{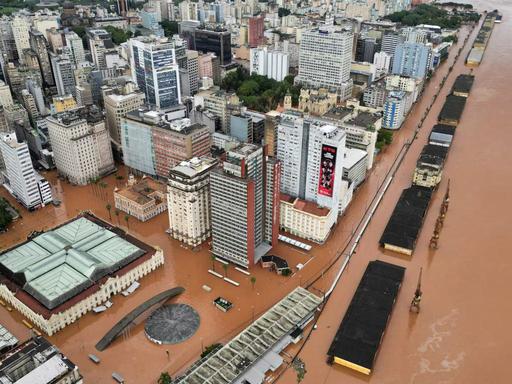}}
    \subfigure[] {\includegraphics[width=0.21\textwidth]{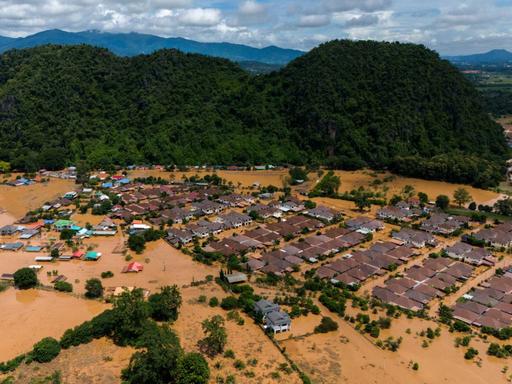}}
    \subfigure[] {\includegraphics[width=0.21\textwidth]{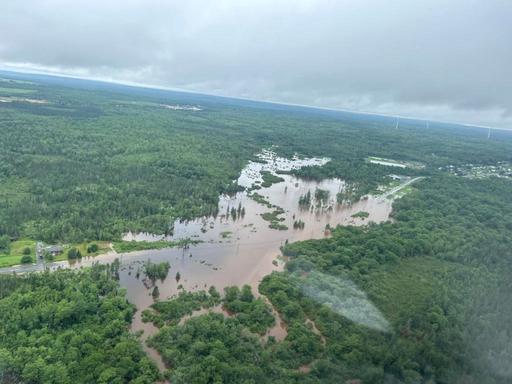}}
    \\

    \subfigure[] {\includegraphics[width=0.21\textwidth]{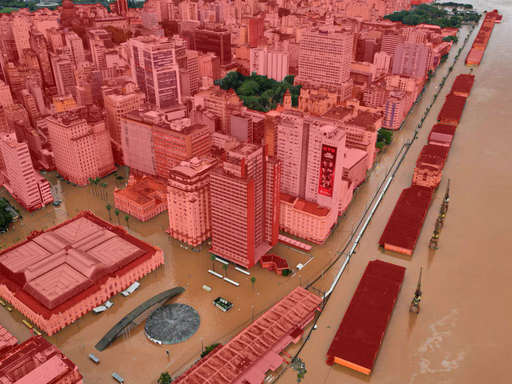}}
    \subfigure[] {\includegraphics[width=0.21\textwidth]{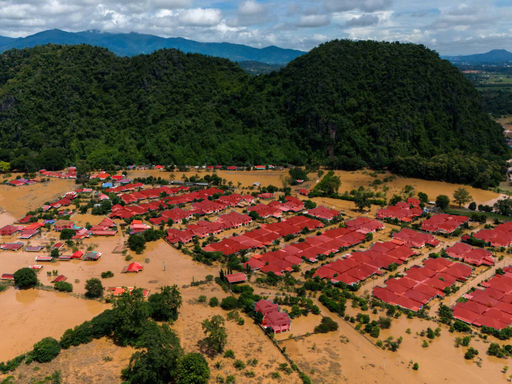}}
    \subfigure[] {\includegraphics[width=0.21\textwidth]{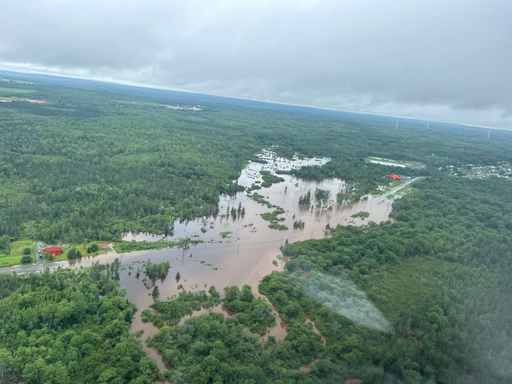}}
    \\
    
    \subfigure[IoU = 89.87\%]{\includegraphics[width=0.21\textwidth]{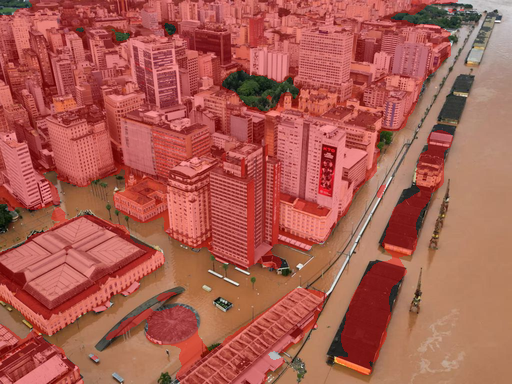}}
    \subfigure[IoU = 62.71\%] {\includegraphics[width=0.21\textwidth]{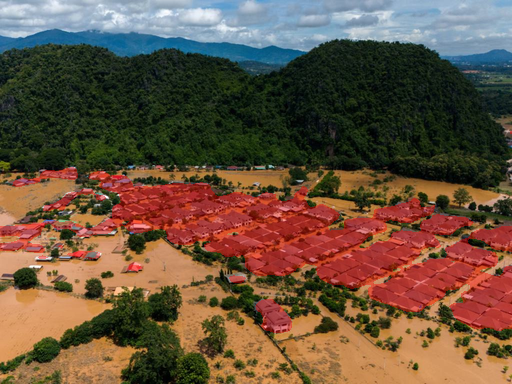}}
    \subfigure[IoU = 0.00\%] {\includegraphics[width=0.21\textwidth]{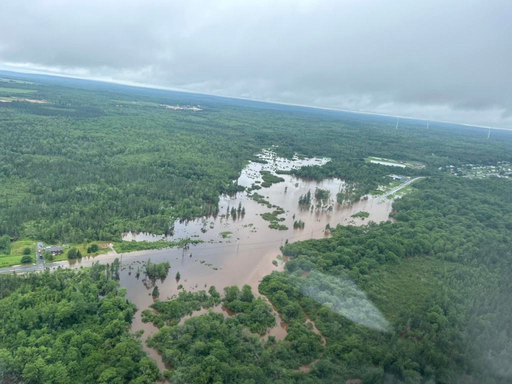}}
    \\

    \subfigure[IoU = 88.86\%] {\includegraphics[width=0.21\textwidth]{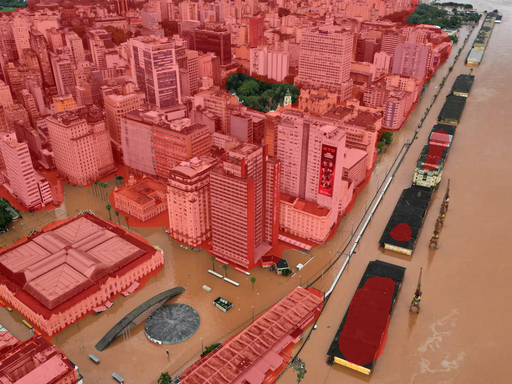}}
    \subfigure[IoU = 53.31\%] {\includegraphics[width=0.21\textwidth]{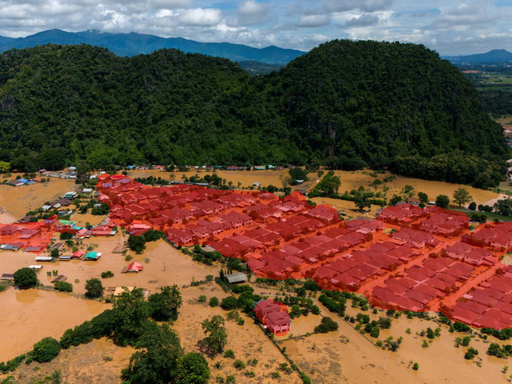}}
    \subfigure[IoU = 0.00\%] {\includegraphics[width=0.21\textwidth]{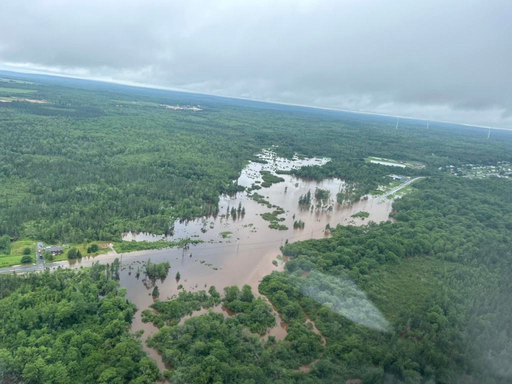}}
    \\

    \subfigure[IoU = 90.80\%] {\includegraphics[width=0.21\textwidth]{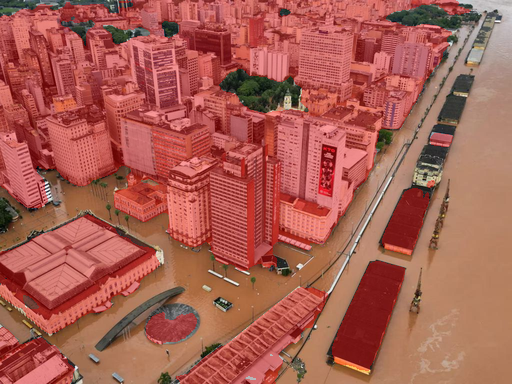}}
    \subfigure[IoU = 61.70\%] {\includegraphics[width=0.21\textwidth]{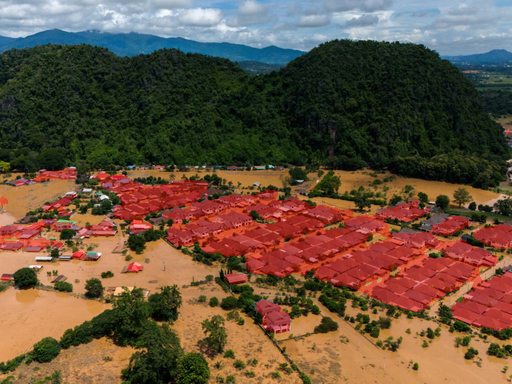}}
    \subfigure[IoU = 34.39\%] {\includegraphics[width=0.21\textwidth]{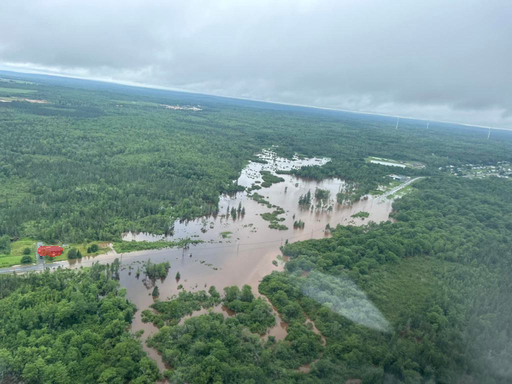}}
    \\

    \subfigure[IoU = 92.63\%] {\includegraphics[width=0.21\textwidth]{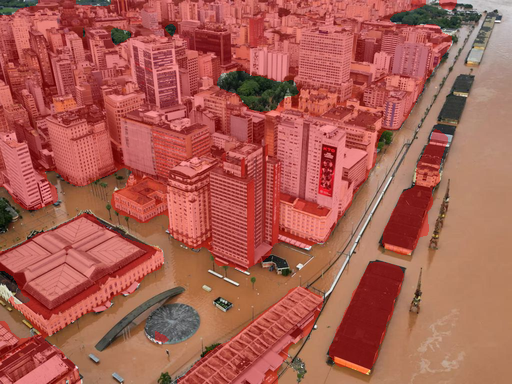}}
    \subfigure[IoU = 56.91\%] {\includegraphics[width=0.21\textwidth]{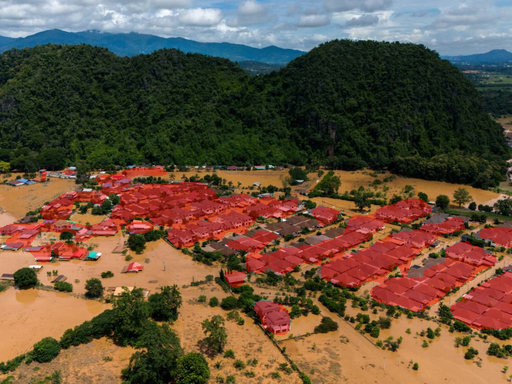}}
    \subfigure[IoU = 0.00\%] {\includegraphics[width=0.21\textwidth]{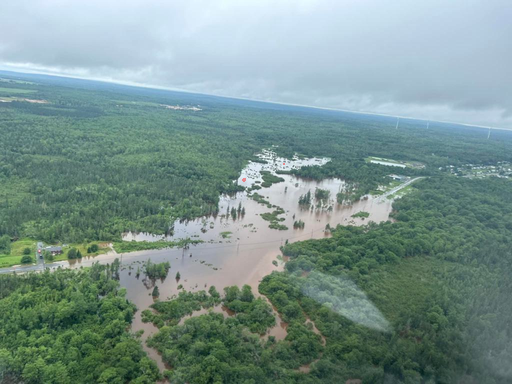}}
    
    \caption{Qualitative comparison of binary segmentation results for class building. The rows display, from top to bottom: original images, ground truth overlays, and prediction overlays from the pre-trained Swin-T, SegFormer-B0, U-Net, and FCN models. Segmentation masks are visualized using red color for building. Columns correspond to best, median, and worst segmentation results according to the descending sorted per-image IoU values of the top-performing overall model, Swin-T.
    }
    \label{fig:segmresbinbuild}
\end{figure*}

Qualitative segmentation examples are presented in Figures~\ref{fig:segmresbinflood}, \ref{fig:segmresbinsky}, and \ref{fig:segmresbinbuild} for the classes Flood, Sky, and Building, respectively. Again per-image IoU is computed and segmentations are shown according to the descending sorted per-image IoUs of the Swin-T pre-trained model, which performed better in the Flood and Building class. For comparative reasons, the respective segmentations of the other pre-trained models are also shown, specifically the other transformer model, SegFormer-B0, and the two best performing convolutional models, U-Net and DeepLabV3 for classes Flood and Sky, and U-Net and FCN for class Building. For visualization and mask encoding, we adopt an overlaid colormap: the flood class is represented as blue, the sky class as green, and the building class as red.

Figure~\ref{fig:segmresbinflood} provides a qualitative assessment of flood segmentation performance in the model spectrum, illustrating the best (0th percentile), 25th, median (50th), and 75th percentile outcomes based on the per-image mIoU distribution. Among the evaluated architectures, Swin-T (i)--(l) consistently yields the most precise flood delineation. However, the U-Net model demonstrates competitive efficacy, matching Swin-T's performance and, in specific instances (e.g., (q) vs. (i) and (r) vs. (j)), exhibiting superior segmentation accuracy. Conversely, SegFormer-B0 (m)--(p) displays the most significant limitations in segmentation quality, characterized by reduced boundary adherence and consistently lower IoU scores, though it occasionally surpasses DeepLabV3 in select challenging scenarios (e.g., (p) vs. (x)).

For the Sky and Building classes, qualitative examples were selected to ensure meaningful visual interpretation, particularly when target regions are absent in the ground truth. Out of 94 test images, only 40 contain sky pixels and 83 contain building pixels. Thus, images without these classes yield IoU = 0 for all models. Including such zero-IoU samples in percentile-based selection (e.g., P0-P75) would bias the visual comparison toward trivial non-sky or non-building scenes. To avoid this, qualitative evaluation was restricted to samples with non-zero IoU values, ensuring that the comparisons reflect actual segmentation performance rather than class absence. The non-zero IoU samples for Swin-T were ranked in descending order, and three representative cases, the best, median, and worst, were chosen to illustrate the variability in segmentation quality across models. These results are presented in Figures~\ref{fig:segmresbinsky} and \ref{fig:segmresbinbuild}, respectively.

As shown in Figure~\ref{fig:segmresbinsky}, all models achieved near-perfect segmentations in certain cases (g), (j), (m), and (p), attaining IoU scores exceeding 99\%. The median Swin-T example reached an IoU of 94.81\% (h), although in this particular case, the remaining models slightly outperformed it (k), (n), and (q). Nevertheless, segmentation quality deteriorated in more challenging scenes, particularly where the sky bordered regions with similar color and texture, such as sea or flooded areas (i), (l), (o), and (r). Interestingly, even the weakest overall model, SegFormer-B0, was able to segment the sky class reasonably well, likely due to the class's distinct color and spatial separation from other classes.

In contrast, the Building class proved to be the most challenging. Buildings appear in diverse shapes, colors, and configurations, often multicolored with differing wall and roof tones, and are captured from variable camera angles, including oblique and top-down perspectives. Moreover, partial submersion due to flooding further complicates boundary delineation. In scenes containing densely packed building clusters, the models generally performed well, as seen in Figure~\ref{fig:segmresbinbuild} (g), (j), (m), and (p). However, when visible spacing exists between adjacent structures (e), the models frequently interpreted them as a single connected region, leading to loss of structural detail (h), (k), (n), and (q). The most challenging scenarios involved sparsely distributed small buildings (f), which were often missed entirely (i), (l), and (r), or only partially segmented (o).

\subsection{VQA Task Assessment}

The results in Table~\ref{tab:vqa_results} highlight clear performance trends across models and question types, reflecting both the complexity of UAV flood imagery in the dataset and the diverse reasoning demands of each question. For all models, we report the mean accuracy computed over the full set of question--answer pairs in the test split. 
In addition to the overall score, we also provide per-question accuracy to capture performance differences across the various semantic categories present in the dataset. 
This is particularly important in domain-specific VQA settings, where models may excel in certain types of reasoning (e.g., yes/no questions) but struggle with others (e.g., spatial or attribute-based queries).

For questions involving numerical estimation or explicit counting, we supplement accuracy with the Root Mean Squared Error (RMSE). 
While accuracy indicates whether the predicted count matches the ground truth exactly, RMSE captures the magnitude of deviations, offering a more informative error measure for near-miss predictions (e.g., predicting ``3'' instead of the correct ``4''). 
This dual metric reporting provides a more complete and robust assessment of model performance on counting-related tasks, where partial correctness is meaningful and often operationally relevant.

\begin{table*}[h]
\centering
\caption{Quantitative results on VQA benchmark. 
For counting questions we report both Accuracy and RMSE (ACC(\%)/RMSE). 
For non-counting questions we report Accuracy only. Accuracy is reported in percent (\%).}
\label{tab:vqa_results}
\resizebox{\textwidth}{!}{
\begin{tabular}{|l||c|c|c|c|c|}
\hline
\textbf{Method} &
\begin{tabular}[c]{@{}c@{}}\textbf{How many}\\\textbf{buildings?}\end{tabular} &
\begin{tabular}[c]{@{}c@{}}\textbf{What}\\\textbf{environment}\\\textbf{is flooded?}\end{tabular} &
\begin{tabular}[c]{@{}c@{}}\textbf{Is there}\\\textbf{sky?}\end{tabular} &
\begin{tabular}[c]{@{}c@{}}\textbf{Is the}\\\textbf{image}\\\textbf{flooded?}\end{tabular} &
\begin{tabular}[c]{@{}c@{}}\textbf{How many}\\\textbf{buildings}\\\textbf{are flooded?}\end{tabular} \\
\hline
\hline
ViLT (frozen) &
16.28 / 23.89 &
60.47 &
69.77 &
9.30 &
18.60 / 21.94 \\
\hline
ViLT (finetuned) &
27.91 / 22.01 &
53.49 &
93.02 &
60.47 &
32.56 / 20.81 \\
\hline
\hline
BLIP-2 (frozen) &
~1.20 / 24.62 &
55.81 &
30.23 &
55.81 &
30.23 / 22.54 \\
\hline
BLIP-2 (LoRA) &
32.56 / 15.61 &
79.07 &
86.05 &
100.0 &
34.88 / 14.60 \\
\hline
\hline
Gemini 2.5 Flash &
18.60 / 9.15 &
69.77 &
74.42 &
90.70 &
18.60 / ~7.64 \\
\hline
\hline
LLaMA 3.2 Vision &
~4.65 / 10.34 &
74.42 &
86.05 &
100.0 &
~6.98 / 10.94 \\
\hline
\end{tabular}
}
\end{table*}

\noindent\textbf{Overall performance and question difficulty:}
Binary questions, such as "\textit{Is the image flooded?}, are consistently the easiest to address across all models. This is expected, as flood presence typically manifests through strong, scene-level cues such as large homogeneous water surfaces, altered color distributions, or reflections, which remain visually distinct even under low-quality imaging conditions. However, this is not the case for the binary question "\textit{What environment is flooded?}" which appears to present moderate difficulty. Distinguishing between rural and urban environments requires recognizing fine-grained structural and textural attributes, such as the presence of buildings, vegetation density, and road layouts. However, in flood conditions, these cues are often partially submerged, occluded, or visually distorted. The water layer can obscure building footprints or mimic uniform textures, reducing the discriminative power of scene-level features, especially for lighter-weight models (e.g. ViLT).

In contrast, the counting questions ("\textit{How many buildings?}" and "\textit{How many buildings are flooded?}") are the most challenging across all methods. This is primarily due to structural and visual ambiguity introduced by flooding which manifest as follows: (a) \textit{Water reflections and debris} can partially obscure building edges, making instance boundaries difficult to detect, (b) \textit{Significant color and texture overlap} between water, land, and rooftops leads to ambiguous building-non-building transitions, (c) \textit{Perspective distortion and variable UAV altitudes} change the apparent size and shape of buildings, complicating object enumeration, and (d) \textit{Flooded buildings lose their characteristic appearance}, making it hard even for humans to estimate the number of submerged structures. These factors collectively make exact counting difficult and explain why accuracy remains low even for fine-tuned models (ViLT, BLIP2).

\noindent\textbf{Counting-question performance discrepancy across the two metrics (Acc, RMSE):}
The discrepancy between accuracy and RMSE between models underscores how challenging precise enumeration is. BLIP-2 (LoRA) yields the best results among trained models, with the highest accuracy and lowest RMSE values for both counting questions. This indicates that fine-tuning helps the model learn more stable visual cues to approximate building counts, even in the presence of occlusions and flood-induced distortions.

Foundation models such as Gemini 2.5 Flash and LLaMA 3.2 Vision exhibit a notable pattern: while their accuracies on counting tasks are low, their RMSE values are competitive, with Gemini 2.5 Flash achieving the lowest RMSE overall. This suggests that these models can often approximate the correct magnitude (e.g., predicting 5 instead of 6) but lack the precise object grounding needed for exact counts.

Finally, ViLT, despite fine-tuning, struggles with both metrics, highlighting the limitations of lighter, early-fusion architectures when detailed object-level reasoning is required.

\section{Discussion on Future Directions}
\label{sec:future}

As highlighted in this work, while existing UAV- and satellite-based flood analysis datasets have made significant strides, the vast majority suffer from geographic myopia or insufficient resolution for fine-grained analysis. AIFloodSense bridges these gaps by providing the first multi-continent  aerial imagery dataset with rich semantic annotations for Sky, Buildings, and Flood regions. These unique characteristics open several new avenues for research.

First, the dataset serves as a rigorous testbed for Domain Generalization (DG) and Unsupervised Domain Adaptation (UDA). The explicit continent labels allow researchers to investigate how models transfer across diverse terrains, for instance, validating if synthetic data strategies or domain adaptation techniques can bridge the gap between simulated environments and the real-world diversity found in AIFloodSense. This additionally positions the dataset as a benchmark for synthetic-to-real studies and generative data augmentation pipelines~\cite{simantiris2025closing}, \cite{simantiris2024unsupervised}, enabling analysis of how diffusion models or domain randomization strategies improve transferability across global flood contexts. Such experiments are critical for developing robust, “deploy-anywhere” models.

Furthermore, auxiliary masks extend analysis capabilities: building masks facilitate precise damage quantification of inundated infrastructure, while sky masks help resolve horizon ambiguities and reduce false positives caused by haze or atmospheric reflections. These structured annotations also make the dataset valuable for studying explainability and human-AI collaboration, for example, by evaluating methods that generate visual rationales for flooded regions or highlight image regions supporting VQA answers. 

Expanding the concept of explainability, the support of AIFloodSense for VQA, transforms flood analysis into an interactive decision-support paradigm, allowing responders to query UAV feeds naturally and bridging the gap between computer vision and actionable human intelligence. This multimodal capability opens additional directions that involve the training and evaluation of models in Earth observation-aware vision language~\cite{Kuckreja2024}, \cite{Soni2025}, enabling research on cross-modal reasoning, resolution of visual uncertainty through language, and the development of tools based on interpretable foundation-models for disaster assessment. 

Looking ahead, we aim to deepen the analytical capabilities of AIFloodSense by moving beyond purely optical imagery. We plan to integrate multispectral sensors, enabling the fusion of RGB data with near-infrared and thermal bands to better discriminate water from vegetation and wet soil. Furthermore, we intend to expand the dataset into a truly multimodal framework by incorporating heterogeneous data sources, such as real-time text from news feeds, meteorological station reports, and geomorphological data (e.g., ground elevation and topographic maps). This rich integration of visual, textual, and spatial data will facilitate the development of more resilient models capable of context-aware flood detection and comprehensive damage assessment, even in complex or visually ambiguous scenarios.

\section{Conclusion}
\label{sec:conclusion}

In this work, we introduced AIFloodSense, a novel and comprehensive dataset of high-resolution UAV imagery captured between 2022 and 2024, covering 230 flood events across 64 countries and all 6 inhabited continents. This collection is designed to advance the state of computer vision for disaster management by addressing the limitations of existing benchmarks, which are often constrained by geographic scope, temporal relevance, or annotation granularity.

The dataset supports a multi-faceted analysis framework by offering annotations for three complementary tasks: (i) Image Classification, including novel sub-tasks for continent and environmental recognition; (ii) Semantic Segmentation, featuring pixel-level masks for Flood, Sky, and Buildings; and (iii) Visual Question Answering (VQA), which introduces natural language reasoning to flood assessment. By uniquely integrating these tasks, AIFloodSense bridges the critical gap between pixel-level perception and high-level reasoning.

We also established a benchmark for future research, by conducting extensive experiments using state-of-the-art CNNs and Transformer based architectures. Our results demonstrate that while modern architectures like Swin-Transformer can achieve promising sky/flood segmentation performance, the diversity of the dataset poses a significant challenge, particularly for fine-grained tasks such as building counting and cross-continent generalization. The established baselines highlighted the critical importance of transfer learning and the need for more robust context-aware models capable of handling the complex visual heterogeneity of real-world flood scenarios, as well as the added value that the proposed dataset offers to the research community.  

Conclusively, AIFloodSense contributes a robust, open-access resource to the community, aiming to accelerate the development of domain-generalized AI tools that are essential for effective climate resilience and global disaster response.

\bibliographystyle{elsarticle-num}
\bibliography{paper}

\end{document}